\documentclass{article}
\usepackage{arxiv}

\usepackage[utf8]{inputenc} 
\usepackage[T1]{fontenc}    
\usepackage{hyperref}       
\usepackage{url}           
\usepackage{booktabs}       
\usepackage{amsfonts}      
\usepackage{nicefrac}      
\usepackage{microtype}     
\usepackage{lipsum}		

\usepackage{graphicx}
\usepackage[numbers]{natbib}
\usepackage{doi}
\usepackage{subcaption}
\usepackage[dvipsnames]{xcolor}

\usepackage{microtype}
\usepackage{graphicx}
\usepackage{booktabs}
\usepackage{hyperref}

\usepackage{amsmath}
\usepackage{amssymb}
\usepackage{mathtools}
\usepackage{amsthm}
\usepackage{stmaryrd}
\usepackage{amssymb}
\newcommand{\xdashrightarrow}[1]{\dashrightarrow}

\usepackage[capitalize]{cleveref}

\theoremstyle{plain}

\theoremstyle{definition}

\theoremstyle{remark}

\usepackage{url}
\graphicspath{ {./images} }
\usepackage{subcaption}
\usepackage{tikz}
\usepackage{float}
\usepackage[ruled, vlined]{algorithm2e}
\usepackage{multirow,tabularx}
\usepackage{xspace}
\usepackage{alltt}
\usepackage{verbatim}
\usepackage{nicefrac}
\usepackage{enumitem}
\setlist[itemize]{
  noitemsep,
  topsep=0pt,
  leftmargin=*,
}

\usepackage{amsmath,amsfonts,bm}

\def\eqref#1{equation~\ref{#1}}

\def\1{\bm{1}}
\newcommand{\train}{\mathcal{D}}

\def\eps{{\epsilon}}

\DeclareMathAlphabet{\mathsfit}{\encodingdefault}{\sfdefault}{m}{sl}
\SetMathAlphabet{\mathsfit}{bold}{\encodingdefault}{\sfdefault}{bx}{n}

\newcommand{\R}{\mathbb{R}}

\DeclareMathOperator*{\argmin}{arg\,min}

\newcommand{\iset}[1]{\llbracket #1 \rrbracket}

\usepackage{bold-extra}
\newcommand{\clam}{\texttt{ClAM}\xspace}
\newcommand{\dclam}{\texttt{\textbf{DCAM}}\xspace}

\def\encoder{\mathbf{e}}
\def\decoder{\mathbf{d}}
\def\brho{{\bm{\rho}}}

\def\loss{{\mathcal{L}}}

\newcommand{\down}{\textcolor{ForestGreen}{$\blacktriangledown$}}

\newcommand{\txsps}[1]{${}^{\text{#1}}$}

\renewcommand{\paragraph}[1]{\noindent {\bf #1}}

\usepackage{multicol}
\usepackage{tcolorbox}

\title{Deep Clustering with Associative Memories}

\date{} 				

\author{%
  Bishwajit Saha \\
  IBM Research, RPI \\
  \texttt{Bishwajit.Saha3@ibm.com}
  \And
  Dmitry Krotov \\
  IBM Research \\
  \texttt{krotov@ibm.com}
  \And
  Mohammed J. Zaki \\
  RPI \\
  \texttt{zaki@cs.rpi.edu}
  \And
  Parikshit Ram \\
  IBM Research \\
  \texttt{Parikshit.Ram@ibm.com}
}

\renewcommand{\shorttitle}{Deep Clustering with Associative Memories}

\begin{document}
\maketitle

\begin{abstract}
Deep clustering -- joint representation learning and latent space clustering -- is a well studied problem especially in computer vision and text processing under the deep learning framework. While the representation learning is generally differentiable, clustering is an inherently discrete optimization task, requiring various approximations and regularizations to fit in a standard differentiable pipeline. This leads to a somewhat disjointed representation learning and clustering. 
In this work, we propose a novel loss function utilizing energy-based dynamics via Associative Memories to formulate a new deep clustering method, \dclam, which ties together the representation learning and clustering aspects more intricately in a single objective. Our experiments showcase the advantage of \dclam, producing improved clustering quality for various architecture choices (convolutional, residual or fully-connected) and data modalities (images or text). 
\end{abstract}


\section{Introduction}
{\color{black}
The goal of clustering is to find coherent groups in a dataset.  It is an important unsupervised learning task, and given the generality of the task, many different methods have been proposed for effective clustering~\citep{xu2015comprehensive, zaki2020data}. 
At a technical level, clustering  
critically relies on a notion of (pairwise) distance (or similarity) to distinguish pairs of data samples as being ``similar'' or ``different'', and the insights from clustering can be unintuitive or misleading without such a meaningful distance.}
When dealing with numerical data $S \subset \R^d$ with $d$ dimensions, metrics such as Euclidean distance are commonly used.  Nevertheless, even with numerical data and an appropriate notion of distance, increasing data dimensionality (that is, increasing $d$) makes clustering computationally hard as well as conceptually difficult since the separation between similar pairs and dissimilar ones can start to vanish~\citep{verleysen2005curse, steinbach2004challenges, assent2012clustering}.

In various domains, both these problems manifest -- first, the raw representation of samples can be extremely high dimensional (consider the number of pixels in an image, or the number of words in a vocabulary for a bag-of-words representation of documents); second, while we have an {\em ambient} representation, standard notions of vector distances (such as Euclidean) do not necessarily make sense -- for example, Euclidean distance based on pixels can be large between an image and a slightly shifted version of it, which can be problematic if the content of an image is translation or rotation invariant.

{\color{black} One effective approach to handle these challenges is through {\em deep clustering}~\citep{zhou2025comprehensive}, where the goal is to both learn a low dimensional {\em latent} space where standard distance metrics are meaningful, and to cluster or group the points at the same time.}
For the latent representations to be faithful to the original samples, deep clustering ensures that there is no significant information loss in the latent space, leading to the common use of autoencoders (AEs)~\citep{rumelhart1985learning, baldi2012autoencoders, bank2023autoencoders} that learn latent representations (via an encoder) which can be used to reconstruct the original samples (via a decoder). The goal of deep clustering is to discover the cluster structure in the latent space while ensuring low reconstruction loss. This is a well studied problem, especially in image datasets~\citep{caron2018deep, chang2017deep}.
While an autoencoder is usually differentiable, standard clustering schemes (such as $k$-means~\citep{macqueen1967classification} or agglomerative~\citep{johnson1967hierarchical}) are inherently discrete methods, since {\em hard} clustering (where each sample is only assigned to a single cluster) is a discrete optimization problem. To incorporate it in a differentiable deep learning pipeline, clustering is often ``softened'' by allowing samples to be partially assigned to multiple clusters, although various ``regularizations''  push the soft assignments to match hard assignments approximately~\citep{xie2016unsupervised, guo2017improved}.
{\color{black} The recent \clam~\citep{saha2023end} algorithm
handles the dichotomy between hard assignments and differentiability via the use of associative memories, yielding an end-to-end differentiable clustering approach.
Nevertheless, \clam works only in the ambient $d$-dimensional data space, and is not designed to learn effective lower dimensional latent representations, which poses challenges when clustering high-dimensional data.}

\begin{figure}[!ht]
\begin{center}
\begin{subfigure}{0.23\columnwidth}
\centering
    \includegraphics[width=1\textwidth]{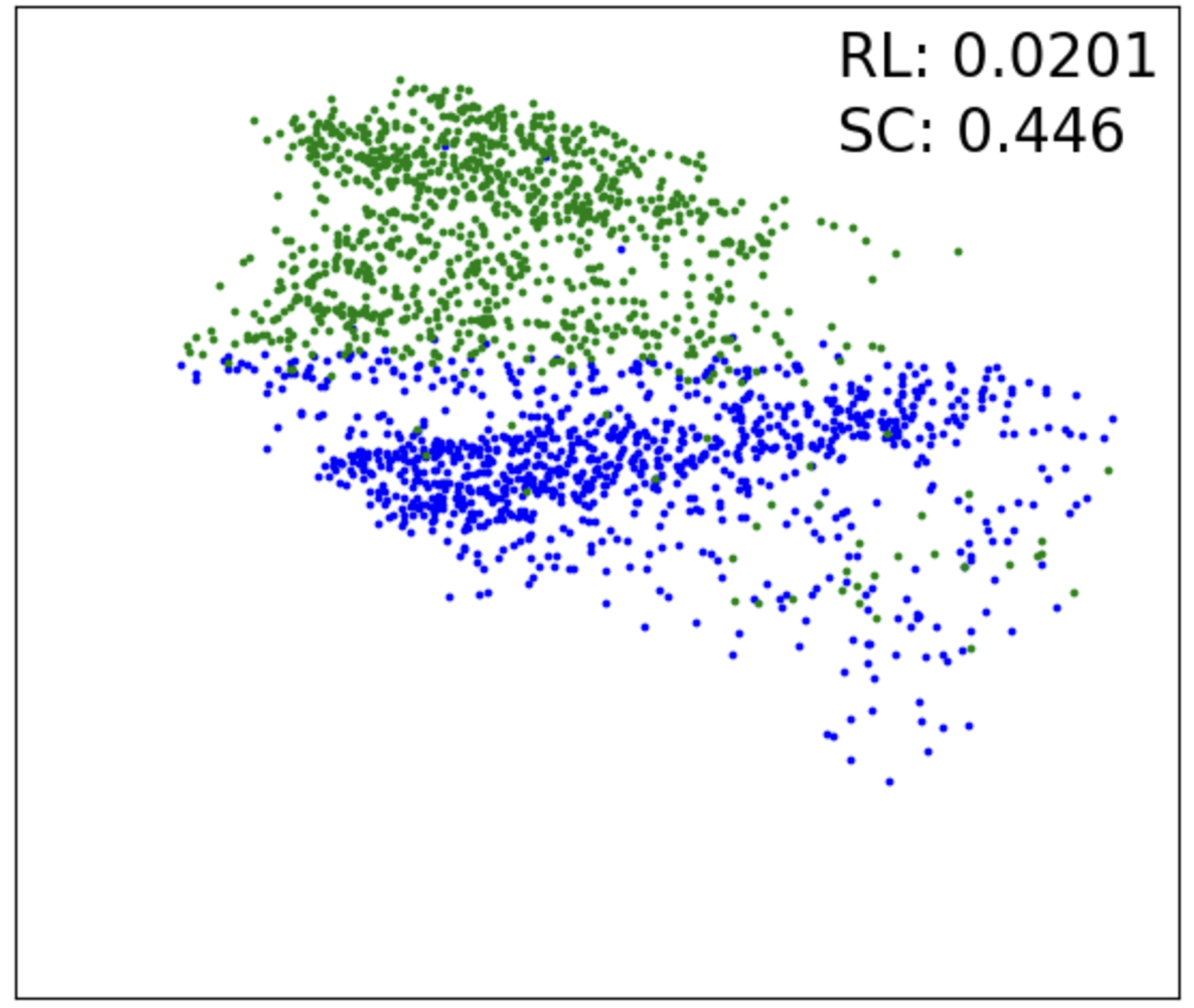}
  \caption{Pretrained}
  \label{fig:hp:pretrained}
\end{subfigure}
~
\begin{subfigure}{0.23\columnwidth}
\centering
  \includegraphics[width=1\textwidth]{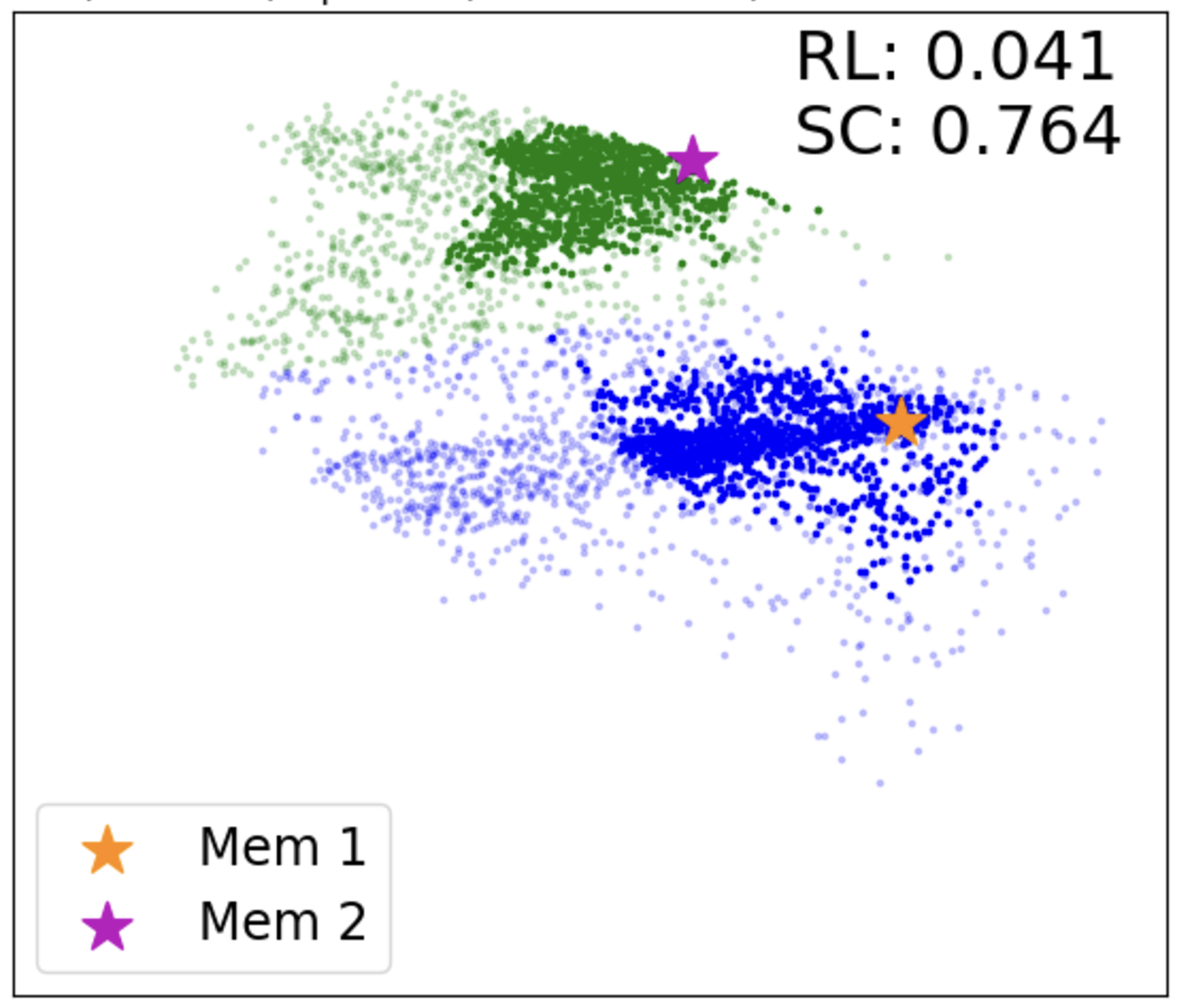}
  \caption{e:1, t:7}
  \label{fig:hp:epoch_1_step_7}
\end{subfigure}
~
\begin{subfigure}{0.23\columnwidth}
\centering
    \includegraphics[width=1\textwidth]{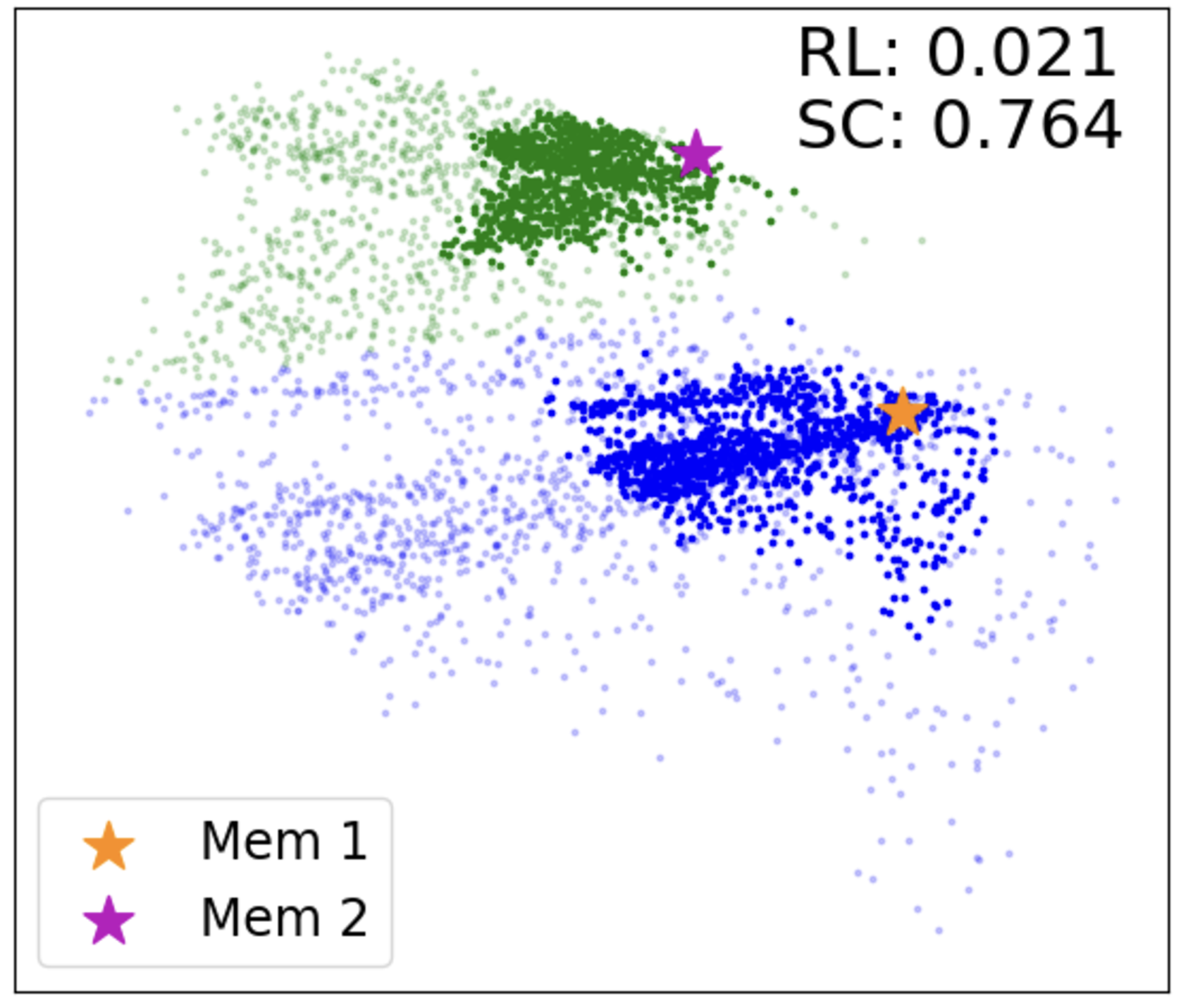}
  \caption{e:200, t:8}
  \label{fig:hp:epoch_200_step_8}
\end{subfigure}
~
\begin{subfigure}{0.23\columnwidth}
\centering
  \includegraphics[width=1\textwidth]{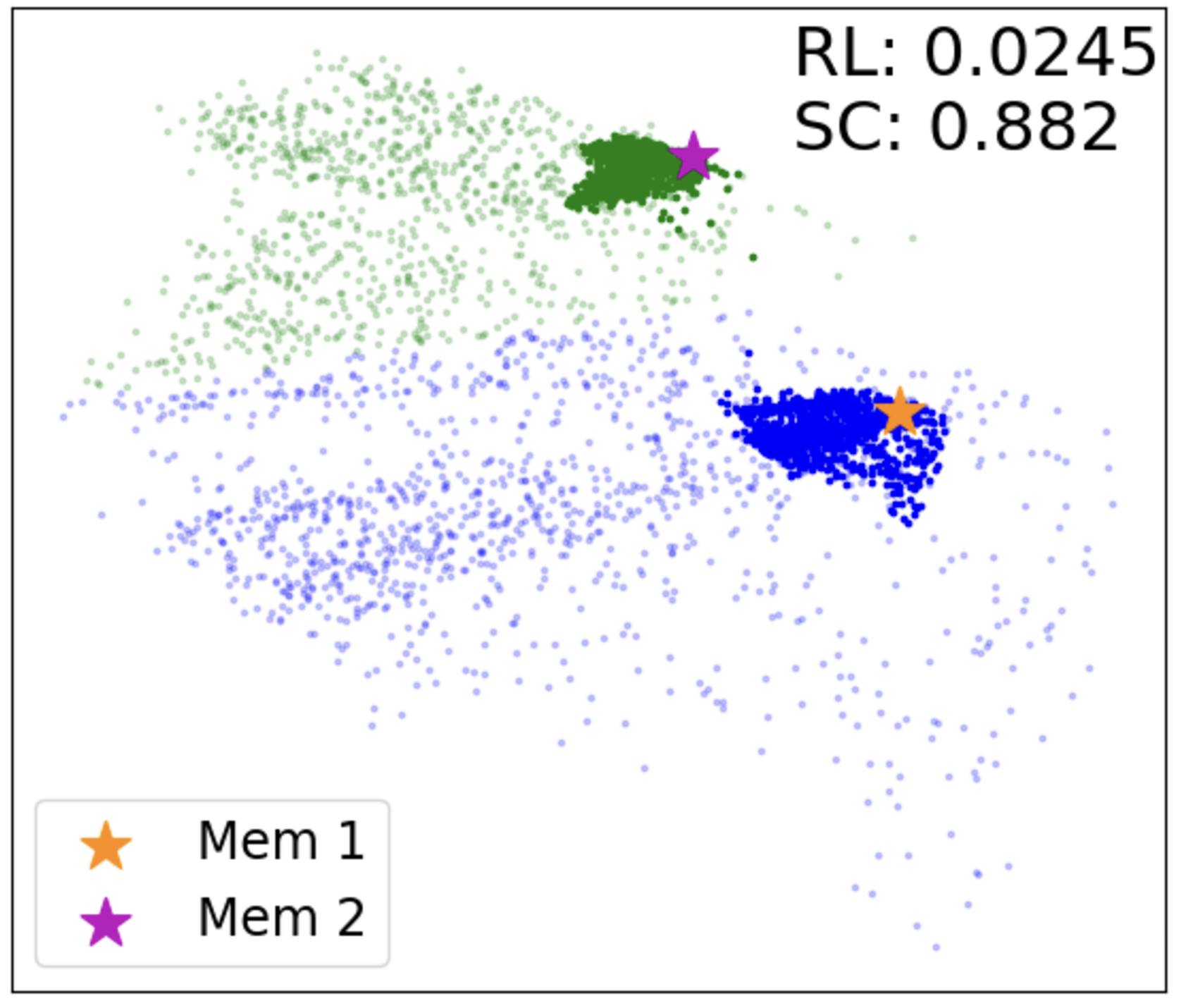}
  \caption{e:800, t:15}
  \label{fig:hp:epoch_800_step_15}
\end{subfigure}
\end{center} 
\caption{\textbf{Clustering with \dclam}. Reconstruction loss (RL; refers to the loss $\bar \loss$ in \cref{eq:dc-am}) and clustering quality (SC; refers to Silhouette Coefficient) for different epochs (e) and number of AM steps (t) for Fashion-MNIST with two clusters, where \cref{fig:hp:pretrained} represents the pretrained latent representations. The light (faded) colors indicate the encoded points before applying the attractor dynamics, whereas dark colors indicate the points after applying that step. The colored stars represent the learned prototypes. \dclam discovers more compact and clustering-friendly latent representations that simultaneously have higher clustering and reconstruction quality. 
}
\label{fig:2d-plots-fm}
\end{figure}
{\color{black} As noted above, deep clustering tackles the joint objective of learning a good latent representation where the points also cluster well. Whereas minimizing reconstruction loss is a prerequisite for deep representation learning, one option for clustering in latent space is to first {\em pretrain} an autoencoder to minimize the reconstruction loss, and then to freeze this latent space. Next, one can apply some clustering scheme to group the points in that (frozen) space. Many AE based existing deep learning methods adopt this scheme by either freezing both the encoder and decoder, or freezing only the decoder~\citep{xie2016unsupervised, guo2017deep, guo2021deep, chazan2019deep, huang2023deep}.}

We propose a new approach called \dclam that uses Associative Memories (AM) as an inductive bias over the latent space that
i) defines an (learnable) energy function over the latent space, and ii) utilizes the AM attractor dynamics to pull similar latent representations closer together. This leads to latent representations that are inherently well clustered without any explicit clustering
objective, and the use of AM makes the whole process (i.e., learning the encoder, decoder, and cluster prototypes) end-to-end differentiable.
Our key insight and contribution is that we seamlessly combine the clustering and reconstruction loss objectives into one expression that tackles the task of {\bf clustering-guided latent representations}, whereas previous deep clustering methods considered these separately. 
\cref{fig:2d-plots-fm} shows an illustration of how \dclam evolves over the epochs 
while successively finding better clusters.
Specifically, our work makes the following contributions:

\begin{itemize}
\item We propose \dclam, which uses associative memories to formulate a novel joint loss function that simultaneously learns effective representations and clusters in the latent space.
\item We conduct a thorough evaluation on image and text datasets, demonstrating that \dclam significantly improves the clustering quality over both traditional (in ambient space) and deep clustering (in latent space) baselines. 
\item We show that \dclam retains superior representation quality as measured by the reconstruction loss; it is also agnostic to the encoder/decoder architecture choice.
\end{itemize}
\section{Related Work} \label{sec:litreview}
Clustering is a long-studied and well-reviewed problem in computer science, with various formulations and several applications~\citep{kaufman2009finding, zaki2020data}. 
Given the success of deep learning, deep clustering has also attracted attention
over the past decade~\citep{ren2022deep, aljalbout2018clustering, zhou2025comprehensive}.
Inspired by t-SNE~\citep{van2008visualizing}, \citet{xie2016unsupervised} introduced DEC, enhancing clustering and feature representation by minimizing the Kullback-Leibler Divergence (KLD) to an auxiliary target distribution. However, a drawback is abandoning the decoder layer after pre-training, impacting the embedded space and clustering performance. \citet{guo2017improved} showed that keeping the decoder layer improves clustering (IDEC), and \citet{guo2017deep} proposed DCEC using convolutional autoencoders (CAE). 

\citet{yang2017dcn} jointly optimized an autoencoder and a latent-space $k$-means objective.
\citet{chazan2019deep} proposed DAMIC, a mixture of autoencoders for clustering, determined by minimizing the reconstruction loss without needing a regularization term. However, they leverage multiple AEs in their model, while we focus on schemes using a single AE.
\citet{huang2023deep} introduced an innovative embedded autoencoder architecture by incorporating it into both the encoding and decoding units of the outer autoencoder. \citet{guo2021deep} proposed DEKM which works on the embedding space (after pretraining) and transforms it to a new cluster-friendly space using an orthonormal transformation matrix.  
However, discarding the decoder after pretraining for both of these methods may lead to the distortion of the embedded space, consequently hurting clustering performance. In addressing the automatic inference of the number of clusters in a dataset, \citet{ronen2022deepdpm} introduced DeepDPM. They proposed a novel loss inspired by EM in the Bayesian Gaussian Mixture Model framework, facilitating a new amortized inference in mixture models. It is worth noting that DeepDPM diverges from the typical encoder-decoder architecture, opting instead for a multilayer perceptron model. 

While many deep clustering methods utilize KLD as a clustering objective, it falls short in preserving the global data structure (e.g., only within-cluster distances are prioritized, leaving uncertainties regarding between-cluster similarities), leading \citet{golzari2023edcwrn} (EDCWRN) to advocate for cross-entropy over KLD. They incorporate feature weighting to emphasize essential features for clustering and employ a neighborhood technique to encourage similar representations for samples within the same cluster. Addressing another challenge with KLD regarding the presence of hard, misclassified samples, \citet{cai2022unsupervised} introduced focal loss to enhance label assignment in deep clustering methods and improved the representation learning module with a contractive penalty term, capturing more discriminative representations. However, it could lead to unintentional bias in the optimization focus between the representation learning and clustering modules. \citet{dang2021nearest} introduce a novel deep clustering framework  (NNM) based on a two-level nearest neighbors matching approach. Distinguishing itself from prior methods~\citep{van2020scan}, NNM incorporates matching at both local and global levels, resulting in a notable enhancement in clustering performance. It also leverages SimCLR~\citep{chen2020simple} to pretrain a representation learning model using the state-of-the-art contrastive learning loss. 
Our \dclam approach can flexibly incorporate various autoencoder architectures by leveraging the capabilities of associative memories, and can benefit from various architectural and pretraining advancements.

Associative Memories store multidimensional vectors as fixed point attractor states in a recurrent dynamical system \cite{krotov2025modern}. AMs form associations between the initial state and a final state (memory), creating disjoint basins of attractions which are crucial for clustering. A prominent example of AM is the classical Hopfield Network~\citep{hopfield1982neural}. It possesses limited memory capacity, approximately storing only $\approx 0.14d$ arbitrary memories 
in a $d$ dimensional data domain~\citep{mceliece1987capacity, amit1985storing}. 
Subsequently, \citet{krotov2016dense,krotov2018dense} proposed Dense Associative Memory (Dense AM) or Modern Hopfield Networks introducing rapidly growing non-linearities (activation functions) into the system. This innovation allows for a denser arrangement of memories and achieves super-linear (in $d$) memory capacity~\citep{demircigil2017model, lucibello2024exponential}. With softmax activation, Dense AMs are closely related to the attention mechanism used in transformers~\citep{ramsauer2021hopfield,krotov2021large, hoover2024energy} and diffusion models \cite{hoover2023memory,ambrogioni2024search,pham2025memorization}. 

Recently, \citet{saha2023end} introduced \clam, an end-to-end differentiable clustering approach, utilizing Dense AMs for clustering. 
However, there are other fundamental difference between how \dclam uses AMs versus \clam, which performs clustering only in the ambient space utilizing AMs but there is no representation learning involved. In contrast, \dclam focuses on clustering in latent space, utilizing AMs to find good clusters and yet retain good reconstruction (which reflects the quality of the latent representations). In \clam, AM is utilized to act as a differentiable argmin solver for the $k$-means objective. In contrast, in \dclam, which involves representation learning, AM energy dynamics explicitly creates basins of attraction in the latent space, and moves/pushes the latent representations of the points into these basins, thereby explicitly inducing a clustered data distribution in the latent space. While the encoder is moving points into basins of attraction, the \dclam loss tries to minimize the information loss in the latent representations by having the decoder reconstruct these relocated latent representations. Finally, we show empirically that when \clam is directly applied in the latent space learned by a pretrained autoencoder, it does not yield competitive clustering performance. Our novel \dclam approach, which is inherently a deep clustering method that jointly clusters and learns effective latent representations, yields much better performance.

{\em To our knowledge, the coupling of deep clustering with energy dynamics in latent space as done in \dclam for cluster-guided latent space learning has not been considered in the literature before.}
\dclam continuously refines both the encoder and decoder networks and at the same time integrate the AM learning dynamics to cluster the points into $k$ groups. This bears semblance to vector-quantized variational AEs~\citep{vq-vae}, where the task is to learn a discrete vector code for each point. However, this assignment is non-differentiable, requiring gradient approximation, and there is no clustering objective considered. 
Also related is the task of deep metric learning~\citep{kaya2019deep}, where the task is to learn a distance function between samples in latent space. However, this requires the use of labeled data for full or weak supervision. 
\section{Deep Clustering}
Let $S \subset \R^d$ denote the input data in the ambient space, with an instance $x \in S$, and $\iset{n}$ a $n$-length index set $\{1, \ldots, n \}$.
Deep clustering is an unsupervised task, where we have to learn (usually lower dimensional) representations such that (i) no (critical) information is lost in the latent lower dimensional representations, and (ii) the data in the latent space forms well-separated clusters. 
To ensure that no information is lost in the latent space, we learn an encoder $\encoder: \R^d \to \R^m$ ($m < d$) that maps the input $x \in \R^d$ to a latent space (that is, $\encoder(x) \in \R^m$), along with a decoder $\decoder: \R^m \to \R^d$ that maps the latent representation back to the original ambient space. Encoder $\encoder$ and decoder $\decoder$ together give us an autoencoder, and the loss of information is often measured as the {\em reconstruction loss}:
\begin{equation*}
\loss_r(\encoder, \decoder) = \sum_{x \in S} \ell_r (x, \encoder, \decoder) = \sum_{x \in S} \|x - \decoder(\encoder(x)) \|^2 
\end{equation*}

This loss term does not account for the cluster structure in the latent space. For that purpose, we consider $k$ cluster centers $\brho = \{\rho_1, \ldots, \rho_k\} \subset \R^m$ in the latent space, so that the corresponding {\em clustering loss} is given by:
\begin{equation*}
\loss_c(\encoder, \brho) = \sum_{x \in S} \ell_c(x, \encoder, \brho) = \sum_{x \in S} \min_{i \in \iset{k}} \|\encoder(x) - \rho_i\|^2
\end{equation*}
which measures how close a sample is to its closest cluster center in the latent space with the $\min_{i\in\iset{k}}$ performed on a per-sample basis to denote the discrete assignment. 
A small value of $\loss_c(\encoder, \brho)$ implies that all points in the latent space are close to their respective cluster centers. 

Unsupervised deep clustering is often considered in the following form
~\citep{guo2017improved, guo2017deep, cai2022unsupervised}
\begin{equation}\label{eq:dc-uc}
\min_{\encoder, \decoder, \brho} \loss_r(\encoder, \decoder) + \gamma \loss_c(\encoder, \brho) 
\end{equation}
where $\gamma \geq 0$ is a hyperparameter that balances the clustering loss $\loss_c$ and the reconstruction loss $\loss_r$.
This $\gamma$ plays a critical role in balancing the two terms in \cref{eq:dc-uc}.

While $\gamma$ can be tuned via hyperparameter optimization, there is an inherent challenge in the above objective --- the terms $\loss_c$ and $\loss_r$ are not inherently comparable. The per-sample clustering loss $\ell_c(x, \encoder, \brho)$ is a loss computed between entities in the latent space $\R^m$, while per-sample reconstruction loss $\ell_r(x, \encoder, \decoder)$ is a loss between items in the ambient space $\R^d$. 
Thus, the scale of these two terms can be very different, making it hard to select a good value for $\gamma$.
Usual implementations of deep clustering~\citep{guo2017improved, guo2017deep, golzari2023edcwrn} adopt the following strategy: (i)~First, an autoencoder (that is, $\encoder$ and $\decoder$) is ``pretrained'' with the data to achieve low reconstruction error (that is, low $\loss_r$ by setting $\gamma = 0$ in \cref{eq:dc-uc}),
and (ii)~second, $\gamma$ is set to a positive value in \cref{eq:dc-uc}, and the clustering loss $\loss_c$ is minimized by learning the cluster centers $\brho$, and ``fine-tuning'' the encoder $\encoder$, while the reconstruction loss $\loss_r$ stays low by changing the decoder $\decoder$ accordingly {\em if the balancing hyperparameter $\gamma$ is set appropriately, which can be a challenge}.

\section{\dclam: Deep Clustering with AM Dynamics}

Unlike existing deep clustering methods that minimize the objective in \cref{eq:dc-uc}, which involves two separate components, namely the reconstruction loss in ambient space and clustering loss in latent space, in our novel \dclam formulation, we propose an AM-based approach that seamlessly combines these two aspects in a joint loss objective. Our approach not only updates the encoder and decoder, but it also learns effective prototypes for clustering points in latent space.

\begin{figure}[htbp]
\centering
\includegraphics[width=4in,height=4.2in]{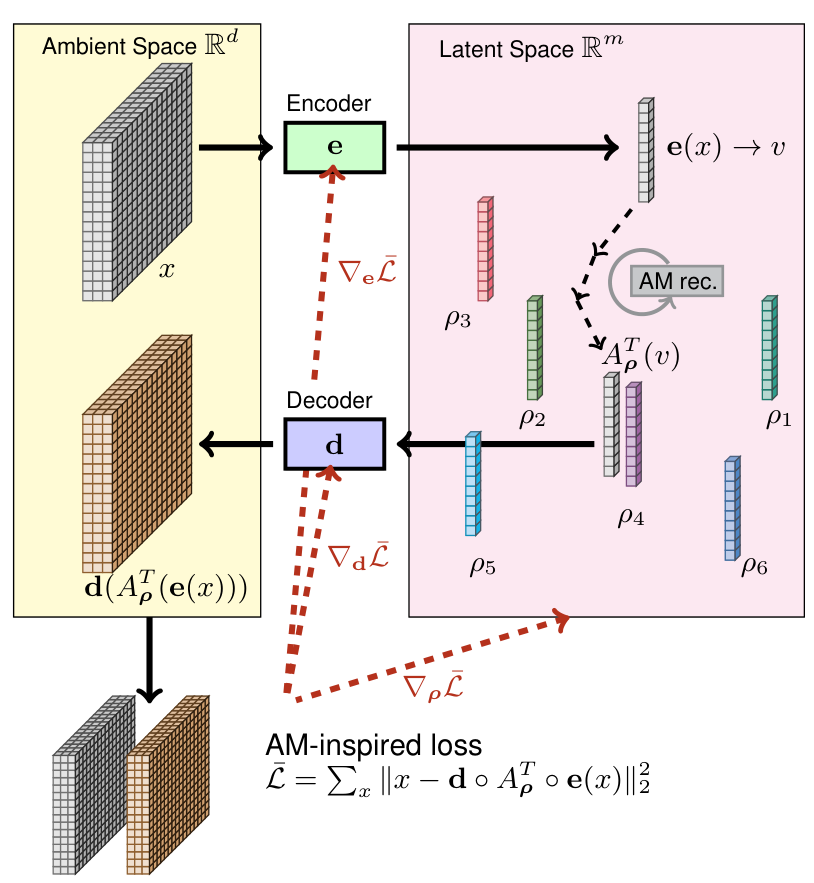}
\caption{\dclam: AM-enabled deep clustering. The solid arrows $\boldsymbol{\xrightarrow{\ \ \ \ \ \ }}$ denote the forward-pass  to compute the single loss term in \cref{eq:dc-am}. The \textcolor{BrickRed}{dashed arrows $\boldsymbol{\xdashrightarrow{\ \ \ \ \ \ \ }}$} denote the backward-pass showing the single loss driving all updates.}
\label{fig:amidc}
\end{figure}
\subsection{Novel Loss Function}
\cref{fig:amidc} shows the overall \dclam pipline.
Given input data $S \subset \R^d$ in the ambient space, 
we first map input points $x \in S$ to the encoded point $v$ in the latent space, using the encoder model $\encoder: \R^d \to \R^m$ ($m < d$), that is, $v = \encoder(x)$. Next, assume that we are given
$k$ memories or cluster prototypes (or centers) in latent space $\brho = \{\rho_1, \rho_2, \ldots, \rho_k\}$, with $\rho_i \in \R^d$, we employ AM dynamics to update the representation of latent points $v$, using $T$ recursive steps, which we denote via the attractor dynamics operator $A_\brho^T(v)$, so that
$v' = A_\brho^T(v)$. We discuss the details of the AM dynamics operator in \cref{sec:am_op} below, but it essentially tries to move the encoded point $v$ closer to a prototype $\rho_i$.
Finally, given the updated latent representation $v'$, we employ the decoder model 
$\decoder: \R^m \to \R^d$ to map it back to the original ambient space, i.e., $x' = \decoder(v')$.
A key feature of \dclam is that the whole pipeline is differentiable, and all the components, namely the encoder parameters, the latent cluster prototypes $\brho$, and the decoder parameters are learnable.
Our novel joint loss function that combines both the clustering and reconstruction aspects into a single expression is defined as:
\begin{equation}\label{eq:dc-am}
\small
\begin{split}
\min_{\encoder, \decoder, \brho} \bar \loss(\encoder, \decoder, \brho)
&
= \sum_{x \in S} 
\bar\ell(x, \encoder, \decoder, \brho)\\
&
= \sum_{x \in S}
\left \| x - \decoder\left( A_\brho^T \left( \encoder\left( x \right) \right) \right) \right \|^2
\end{split}
\end{equation}
Here AM becomes the intricate part of the encoder that transforms the embedding space (obtained by the encoder) into a clustering-friendly new space to find clusters (as opposed to the existing deep clustering schemes that use different additional loss functions, e.g., clustering loss in \cref{eq:dc-uc} and/or regularizations to get a similar effect).
This AM enabled {\em novel deep clustering loss} $\bar \loss$ is a single term that elegantly combines all the parameters in the deep learning pipeline -- for the encoder $\encoder$, the cluster centers $\brho$ and the decoder $\decoder$.

To see the relationship between the novel \dclam loss in \cref{eq:dc-am} compared to the traditional deep clustering loss in \cref{eq:dc-uc}, let us
assume that the decoder $\decoder: \R^m \to \R^d$ is $C_\decoder$-Lipschitz continuous. Then, considering the per-sample loss $\bar\ell$ in \cref{eq:dc-am}, and applying the triangle inequality and the arithmetic and geometric means inequality, we can show that
\begin{align*}
& \bar \ell(x, \encoder, \decoder, \brho) 
=  \| x - \decoder( A_\brho^T( \encoder( x ) ) ) \|^2 \nonumber \\
&\quad \leq 2 \left( \| x - \decoder( \encoder( x ) ) \|^2 + \| \decoder( \encoder( x ) ) - \decoder( A_\brho^T( \encoder( x ) ) ) \|^2 \right) \nonumber \\
&\quad  \leq 2 \left(  \| x - \decoder( \encoder( x ) ) \|^2 + C_\decoder^2 \| \encoder( x ) - A_\brho^T( \encoder( x ) ) \|^2 \right) \nonumber \\
&\quad  = 2 \ell_r(x, \encoder, \decoder) + 2 C_\decoder^2 \ell_c(x, \encoder, \brho),
\end{align*}
where the last inequality uses the Lipschitz continuity, and the last equality comes from the definition of the clustering loss in the latent space with the AM dynamics operator.
Summing the above inequalities
over $x \in S$ gives us
\begin{equation*}
\loss_r \; \leq \; \bar \loss  \; \leq \; \gamma_1 \loss_r + \gamma_2 \loss_c
\end{equation*}
where the upper-bound of $\bar \loss$ is (a scaled version of) the standard deep clustering objective with a weighted combination of the reconstruction loss $\loss_r$ and clustering loss $\loss_c$ in \cref{eq:dc-uc}.

\subsection{Energy Dynamics in Latent Space}
\label{sec:am_op}

We now give details of how the attractor dynamics in latent space.
Given the $k$ cluster prototypes $\brho = \{\rho_1, \ldots, \rho_k\}, \rho_i \in \R^d$, and an encoded latent point $v\in \R^d$, the energy function for $v$ is defined as (Saha et al, 2023):
\begin{equation*}
\small
  E(v) = -\frac{1}{2 \beta}\log
  \left(\sum\nolimits_{i \in\iset{k}} \exp ( -\beta \; \| \rho_i - v \|^2 ) \right)
\end{equation*}
where the scalar $\beta > 0$ denotes inverse temperature, so that
as $\beta$ increases the $\exp(\cdot)$ function emphasizes the leading term, suppressing the others. {\em The attractor dynamics are then driven by gradient descent on the energy landscape}. That this, with $v^0 = v$, the updated representation of $v^t$ is given as 
\begin{equation*}
\small
v^{t+1} = A_\brho(v^t) = v^t - \tau \; \nabla_v E
\end{equation*}
where $\tau > 0$ is the step size that determines how quickly the latent point moves on the energy landscape, and %
\begin{equation*}
\small
\nabla_v E = \frac{\partial E(v)}{\partial v}= \sum_{i \in \iset{k}} (\rho_i - v) 
\; \textsf{softmax}(- \beta \| \rho_i - v \|^2 )
\end{equation*}
with $\textsf{softmax}(- \beta \| \rho_i - v \|^2 )
=
\frac{
\exp( -\beta \|\rho_i - v \|_2^2 )
}{
\sum_{j \in [k]} \exp( -\beta \|\rho_j - v\|_2^2 )
}$.

Thus, the AM operator $A_\brho(v^t)$ denotes the new latent vector obtained by updating $v^t$. Further, we use the notation $A_\brho^T(v)$ to denote the dynamics for $T$ steps, i.e., $v' = A_\brho^T(v) = A_\brho( A_\brho( \cdots A_\brho( v )))$, where the operator $A_\brho$ is applied to $v$ recursively for $T$ steps to obtain the updated representation $v'$.
The attractor dynamics ensure that every memory 
$\rho_i, i \in \iset{k},$ forms a ``basin of attraction'', and with enough recursions $T$, any latent point $v$ will usually converge to exactly one of these memories $\rho_i$, which thus act as cluster centers. Further,
the recursive dynamics is differentiable, with the memories learned via standard backpropagation.

\begin{algorithm}[H]
\caption{\dclam Algorithm}
\label{alg:dcam-com}
\DontPrintSemicolon
{\footnotesize
\SetKwProg{train}{Train$(S, k, N, T, \eps_\encoder, \eps_\decoder, \eps_\brho, \gamma)$}{}{end}
\SetKwProg{infer}{Infer$(S, \encoder, \decoder, \brho)$}{}{end}

\train{}{
Pretrain $(\encoder, \decoder)$ as autoencoder, minimizing $\loss_r(\encoder, \decoder)$\;
$\brho \gets \{\encoder(x), x \in M \}$, $M$ are random $k$ samples from S\;
\For{epoch $n = 1, \ldots, N$}{
  \For{batch $B \in S$}{
    Batch loss $\bar \loss  \gets 0$\;
    \For{example $x \in B$}{
      $v \gets \encoder(x)$\tcp*[l]{\!\!\!encode}
      $v' \gets A_\brho^T(v)$\tcp*[l]{\!\!\!energy descent}
      $\bar \ell \gets \| x - \decoder(v') \|^2$\tcp*[l]{\!\!\!loss}
      $\bar \loss \gets \bar \loss + \bar \ell$\;
    }
    $\rho_i \gets \rho_i - \eps_\brho \nabla_{\!\rho_i} {\bar \loss},\,\, \forall i \in \iset{ k }$ \\[0.2em] 
    $\encoder \gets \encoder - \eps_\encoder \nabla_{\encoder} {\bar \loss}$ \\[0.2em] 
    $\decoder \gets \decoder - \eps_\decoder \nabla_{\decoder} {\bar \loss}$\\[0.2em] 
  }
}
\KwRet{$\encoder, \decoder, \brho$}
}
\BlankLine
\infer{}{
Cluster assignments $C \gets \emptyset$\;
\For{$x \in S$}{
  $v' \gets A_\brho^T( \encoder(x) )$\\
  $C \gets C \cup \left \{ \argmin_{i \in \iset{k}} \| \rho_i - v' \|^2 \right\} $\;
}
\KwRet{Per-point cluster assignments $C$}
}
}
\end{algorithm}

\subsection{\dclam Algorithm}
Alg.~\ref{alg:dcam-com} shows the pseudo-code for \dclam. It first pretrains encoder $\encoder$ and decoder $\decoder$, and starts from $k$ random prototypes $\brho$.
The cluster assignment is done with $T$ recursion of the AM attractor dynamics operator $A_\brho$ parameterized with the centers $\brho = \{ \rho_i, i \in \iset{k} \}$.
The per-sample loss $\bar \ell$ of \dclam (line 10) is added to the batch loss.
We optimize for $N$ epochs via gradient descent, with learning rates $\{\eps_\encoder, \eps_\decoder, \eps_\brho\}$ for $\encoder, \decoder, \brho$ respectively. 
Upon solving \cref{eq:dc-am}, we obtain a trained encoder and decoder, and memories in the latent space, and we can utilize them to obtain the final partition the data (see the {\bf Infer} subroutine in Alg.~\ref{alg:dcam-com}).

\dclam provides various advantages over previous deep clustering formulations: (i)
Our novel per-sample loss $\bar \ell(x, \encoder, \decoder, \brho)$
does not involve a separate clustering loss thus obviating the need for the balancing hyperparameter $\gamma$. (ii)~The updates for all the parameters in \dclam are more explicitly tied together with the $\decoder \circ A_\brho^T \circ \encoder$ composition in the $\decoder( A_\brho^T( \encoder( x ) ) )$ term. {\em This ties the representation learning and clustering objectives more intricately.} (iii)~\dclam continues to have all the advantages of 
traditional deep clustering, being end-to-end differentiable since all operators in the above composition are differentiable, and so is the  discrete cluster center assignment via $T$ recursions of the attractor dynamics operator $A_\brho$. (iv)~It is architecture agnostic -- we can select a problem dependent encoder and decoder. For example, convolutional or residual networks for images or fully-connected feed-forward networks for text or tabular data. Furthermore, this setup can easily handle already trained encoders (for example, one trained via contrastive learning~\citep{chen2020simple, van2020scan}. 
In essence, \dclam introduces an inductive bias over the latent space via AM, which defines an energy function and utilizes the attractor dynamics to help cluster the points.

\section{Empirical Evaluation}
\begin{table*}[!ht]
\caption{Per-method best SC across all architectures (while RRL is within 10\% of the respective pretrained AE loss), comparing \dclam to baselines. Best for each dataset is in bold. {\em Higher SC but lower RRL is better.} The top set of rows are vision datasets, and the bottom set are text datasets. 
A `-' indicates not applicable (NA); e.g., DCEC, DEKM, SCAN, NNM work only on image datasets. 
We report SCAN and NNM results only on C-10, C-100, and STL, as pretrained contrastive encoders are available only for these datasets.
x\txsps{\down} indicates negative RRL which means the RL of the method is x\% less than the pretrained AE loss.}
\label{tab:q1-sc}
\label{tab:q1-sc-rebuttal}
\centering
\resizebox{\textwidth}{!}
{\footnotesize\begin{tabular}{l|ccccccccc|cccc}
\toprule
Dataset & \multicolumn{9}{c|}{SC} & \multicolumn{4}{c}{RRL}\\
\cmidrule(lr){2-14}
& $k$-means & Agglo. & \clam & DCEC & DEKM & EDC & SCAN & NNM & \dclam & DCEC & DEKM & EDC & \dclam\\

\midrule
FM &  0.257 & 0.201 & 0.279 & \textcolor{black}{0.923} & \textcolor{black}{0.260} & 0.483  & - & - & \textcolor{black}{\textbf{0.970}} & 9.8 & \textcolor{black}{\textbf{13.9\txsps{\down}}} & 10 & 1.6\txsps{\down} \\

C-10 & 0.084 & 0.372 & 0.208 & 0.787 & \textcolor{black}{0.116} & 0.511  & 0.541 & 0.587 & \textbf{0.863} & 9.6 & \textcolor{black}{8.6} & 10 & \textcolor{black}{\textbf{19.5\txsps{\down}}} \\

C-100 & 0.015 & 0.149 & 0.053 & \textcolor{black}{0.470} & \textcolor{black}{-0.007} & 0.311  & 0.321 & 0.358 & \textcolor{black}{\textbf{0.598}} & \textcolor{black}{7.5} & \textcolor{black}{\textbf{34.3\txsps{\down}}} & 10 & \textcolor{black}{1.4\txsps{\down}} \\

USPS & 0.195 & 0.158 & 0.194 & \textcolor{black}{\textbf{0.935}} & \textcolor{black}{0.217} & 0.461  & - & - & \textcolor{black}{0.891} & \textcolor{black}{\textbf{5.3\txsps{\down}}} & \textcolor{black}{4.3} & 0.0 & \textcolor{black}{8.7} \\

STL & 0.079 & 0.270 & 0.108 & \textcolor{black}{0.259} & \textcolor{black}{0.082} & 0.411 & 0.552 & 0.540 & \textbf{0.891} & \textcolor{black}{9.2} & \textcolor{black}{0.6} & \textbf{4.9\txsps{\down}} & 10 \\

CBird & -0.019 & 0.094 & -0.026 & 0.311 & -0.032 & 0.171 & - & - & \textbf{0.448} & 10 & \textbf{0.0} & 10 & 9.1 \\

\midrule
R-10k & -0.010 & 0.114 & -0.002 & - & - & 0.023 & - & - & \textbf{0.564} & - & -  & 10 & \textbf{10} \\

20NG & -0.021 & 0.114 & -0.008 & - & - & 0.101  & - & - &  \textbf{0.197} & - & - & 10 & \textbf{10}\\
\bottomrule
\end{tabular}}
\end{table*}

\begin{table*}[!ht]
\caption{Per-method best RRL across all architectures (while SC is within 10\% of the best SC of the method).}
\label{tab:q1-rl}
\label{tab:q1-rl-rebuttal}
\centering
\resizebox{\textwidth}{!}
{\footnotesize\begin{tabular}{l|ccccccccc|cccc}
\toprule
Dataset & \multicolumn{9}{c|}{SC} & \multicolumn{4}{c}{RRL}\\
\cmidrule(lr){2-14}
& $k$-means & Agglo. & \clam & DCEC & DEKM & EDC & SCAN & NNM & \dclam & DCEC & DEKM & EDC & \dclam\\

\midrule
FM &  0.257 & 0.201 & 0.279 & \textcolor{black}{0.898} & \textcolor{black}{0.785} & 0.521  & - & - & \textcolor{black}{\textbf{0.922}} & \textcolor{black}{\textbf{9.8\txsps{\down}}} & \textcolor{black}{321} & 143 & \textcolor{black}{42.2} \\

C-10 & 0.084 & 0.372 & 0.208 & \textcolor{black}{0.786} & \textcolor{black}{0.622} & 0.541  & 0.541 & 0.587 & \textbf{0.809} & \textcolor{black}{0.9\txsps{\down}} & \textcolor{black}{180} & 74.3 & \textcolor{black}{\textbf{20.4\txsps{\down}}} \\

C-100 & 0.015 & 0.149 & 0.053 & \textcolor{black}{0.572} & \textcolor{black}{0.047} & 0.337  & 0.321 & 0.358 & \textcolor{black}{\textbf{0.921}} & \textcolor{black}{\textbf{18.6}} & \textcolor{black}{870} & 33.3 & \textcolor{black}{27.5} \\

USPS & 0.195 & 0.158 & 0.194 & \textcolor{black}{\textbf{0.929}} & \textcolor{black}{0.843} & 0.491  & - & - & \textcolor{black}{0.914} & \textcolor{black}{\textbf{26.3\txsps{\down}}} & \textcolor{black}{2326} & 40 & \textcolor{black}{8.7} \\

STL & 0.079 & 0.270 & 0.108 & \textcolor{black}{0.812} & \textcolor{black}{0.804} & 0.431 & 0.552 & 0.540 & \textbf{0.923} & \textcolor{black}{79.2} & \textcolor{black}{234} & 155 & \textbf{27.7} \\

CBird & -0.019 & 0.094 & -0.026 & 0.282 & 0.018 & 0.188 & - & - & \textbf{0.413} & 286 & 1036 & 102 & \textbf{1.8} \\

\midrule
R-10k & -0.010 & 0.114 & -0.002 & - & - & 0.035 & - & - & \textbf{0.673} & - & - & \textbf{60} & 120 \\

20NG & -0.021 & 0.114 & -0.008 & - & - & 0.099  & - & - &  \textbf{0.287} & - & - & \textbf{25\txsps{\down}}  & 50 \\
\bottomrule
\end{tabular}}
\end{table*}
\label{sec:exp}
We evaluate the performance of \dclam on a diverse set of 6 images and 2 text datasets, ranging in size from 296 to 49152 (raw) features and containing 2007 to 60000 samples. The selection of the number of clusters for each dataset is based on its intrinsic class count, with no reliance on class information during clustering or hyperparameter selection (see dataset details in \cref{asec:exp-details:data}). 

\paragraph{Evaluation Metrics:}
A common metric to evaluate and benchmark deep clustering algorithms is by computing the overlap between the clusters in the latent space 
and the partitions obtained from some ground-truth labels, e.g., Normalized Mutual Information (NMI)~\citep{vinh2009information}, Adjusted Rand Index (ARI)~\citep{hubert1985comparing} or Accuracy (ACC). Nevertheless, {\em it is critical to ensure that NMI/ARI/ACC (or any other label-dependent metric) is not utilized for hyperparameter selection} since that leaks supervision into the unsupervised task of deep clustering.
Unfortunately, for many of reported results, 
it is not clear how hyperparameters are selected without being influenced by these label-dependent metrics.
Furthermore, existing works typically report these metrics without explicitly discussing reconstruction loss, which may not align with the primary goals of deep clustering. 

Given the unsupervised nature of the deep clustering,  hyperparameters should be selected based on unsupervised metrics that do not utilize ground-truth labels to evaluate clustering quality.
Thus, we report results on optimizing for the unsupervised Silhouette Coefficient (SC)~\citep{rousseeuw1987silhouettes} metric, while keeping the reconstruction loss (RL) below some user-defined threshold. We do also report NMI results in \cref{asec:emp:all-metrics}, along with other metrics.

\paragraph{Baseline Methods:}
We conduct a comparative analysis of \dclam against 
deep clustering methods like 
DCEC~\citep{guo2017deep}, DEKM~\citep{guo2021deep} and EDCWRN (or EDC)~\cite{golzari2023edcwrn} in the latent space. We also compare \dclam with state-of-the-art SimCLR~\citep{chen2020simple} based (contrastive learning) SCAN~\citep{van2020scan} and NNM~\citep{dang2021nearest} deep clustering schemes. Finally, we compare with $k$-means~\citep{lloyd1982least}, agglomerative clustering (or Agglo.)~\citep{mullner2011modern} and \clam~\citep{saha2023end}, both in the ambient space (denoted as NAE for No AutoEncoder) as well as in the latent space
obtained through a pretrained Convolutional Autoencoder (CAE) from DCEC~\citep{guo2017deep}. 

To test effect of different autoencoders, for DCEC amd DEKM, we consider a ResNet-based AE (RAE)~\citep{wickramasinghe2021resnet} along with their original CAE.
For \dclam, we extend our exploration to include not only the CAE and RAE architectures but also EDCWRN-based~\citep{golzari2023edcwrn} Autoencoder (EAE) (originally proposed by \citet{guo2017improved}) to analyze its impact on the algorithm. The architectural details of the different autocoders are given in \cref{asec:exp-details:parameter}.

\paragraph{Implementation Details and Tuning:}
We implement \dclam using the Tensorflow~\citep{abadi2016tensorflow} library while employing {\tt scikit-learn}~\citep{pedregosa2011scikit} for quality metrics, and for baseline schemes $k$-means and agglomerative.
For other methods we use the implementations provided by the authors.
We train all models on a single node with 1 NVIDIA RTX A6000 (48GB RAM) and a 16-core 2.4GHz Intel Xeon(R) Silver 4314 CPU. Hyperparameters are tuned individually for each method and each dataset to maximize the Silhouette Coefficient~\citep{rousseeuw1987silhouettes} or the other metrics.
Implementation details for the different methods are given in \cref{asec:expt-details:implementation}, and hyperparameter tuning details in \cref{asec:add-exp:baseline-hpo}. For \dclam, we use curriculum learning to determine $T$, the number of AM or energy descent steps in latent space, leaving only one main hyperparameter $\beta$ that needs to be tuned, as described in \cref{asec:add-exp:dclam-hpo}.

\paragraph{Comparison with Baselines:}
We present the best Silhouette Coefficient or SC achieved (while constraining the reconstruction loss or RL to be within 10\% of the pretrained AE loss) for \dclam and the baselines for all 8 datasets in \cref{tab:q1-sc}.
As it is hard to compare the raw RL numbers if the base AE is different for different methods, we define the relative RL (RRL) metric as follows:
\begin{equation}
RRL = (RL - RL_p)/RL_p
\label{eq:rrl}
\end{equation}
where $RL_p$ is the pretrained/base RL. We report the best SC per method with $RRL <= 10\%$.

From \cref{tab:q1-sc}, we see across both image and text datasets, \dclam consistently outperforms traditional and deep clustering baselines in terms of SC while keeping RRL relatively low. 
To provide a comprehensive view alongside SC, we also present the best RRL results (while constraining the SC to be within 10\% of the best/peak SC of the method) in \cref{tab:q1-rl}. 
Note that SCAN and NNM do not have a reconstruction loss term as they work on the pretrained (pretext) model by SimCLR~\citep{chen2020simple} and utilize only the encoder (discarding the decoder) for clustering purpose.
We observe that \dclam has the best SC values across all datasets except for USPS where DCEC performs the best. Its RL remains competitive. For example, it outperforms DCEC in 4 our of the 6 image datasets. While DEKM has better RRL, its SC values are not very high.
These results demonstrate that \dclam excels not only in achieving the best SC but also in simultaneously minimizing RL compared to the baselines. 
It is important to note that the \clam, $k$-means and Agglomerative clustering results are the best from either using no autoencoder, or those from applying them in latent space on the points obtained from the pretrained autoencoder. In particular, we can see that \dclam vastly outperforms a straightforward application of \clam in latent space.

For additional insights, in \cref{asec:emp:all-metrics}, we present the best SC (while keeping RL within 10\% of the pretrained AE loss) and its corresponding NMI, RL, cluster size and balance metrics obtained by all schemes in \cref{tab:scgt}, and in
\cref{tab:rlgt} we report the best RL (while keeping SC within 10\% of the best SC of the method) and its associated SC, NMI, and other metrics.
Finally, in \cref{tab:nmigt} we further report the best NMI obtained along with the associated SC, RL, and other metrics.
These results clearly show that \dclam offers the best clustering performance in terms of SC, as well as having low reconstruction loss. It also performs very well on the supervised NMI metric. In fact, for NMI, 
 \dclam has the best value in 5 out of the 8 datasets (see \cref{tab:nmigt}).

For a fair comparison with SimCLR-based methods SCAN and NNM, we use SimCLR-pretrained features as input to \dclam. Given these features are already clustering-friendly, we apply a simple one-layer MLP autoencoder (input\_dim $\to$  ld $\to$  input\_dim) to reduce dimensionality, where ld is set to same as the number of clusters. \cref{tab:simclr} shows \dclam consistently outperforms pretext + $k$-means and remains competitive with SCAN and NNM. However, our primary goal is optimizing unsupervised metrics—silhouette coefficient (SC) and reconstruction loss (RL)—rather than relying on ground-truth-dependent metrics (NMI, ACC or ARI), thus making \dclam especially suitable for fully unsupervised settings.
\begin{table}[!ht]
\caption{NMI, ACC and ARI comparison for Cifar-10, Cifar-100-20 and STL, comparing \dclam to baselines with SimCLR based architectures. Best for each dataset is in bold.}
\label{tab:simclr}
\centering
{\footnotesize\begin{tabular}{l|lll|lll|lll}
\toprule
Dataset & \multicolumn{3}{c|}{C-10} & \multicolumn{3}{c|}{C-100-20} & \multicolumn{2}{c}{STL} \\
\cmidrule(lr){1-10}
Metrics & NMI & ACC & ARI & NMI & ACC & ARI & NMI & ACC & ARI \\
\midrule
$k$-means & 0.647 & 0.746 & 0.541 & 0.398 & 0.393 & 0.216 & 0.529 & 0.515 & 0.353 \\
SCAN & 0.715 & 0.816 & 0.665 & 0.449 & 0.440 & 0.283 & 0.673 & 0.792 & 0.618 \\
NNM & \textbf{0.748} & \textbf{0.843} & \textbf{0.709} & \textbf{0.484} & \textbf{0.477} & \textbf{0.316} & \textbf{0.694} & \textbf{0.808} & \textbf{0.650} \\
\dclam & 0.697 & 0.767 & 0.596 & 0.436 & 0.438 & 0.247 & 0.621 & 0.646 & 0.451 \\
\bottomrule
\end{tabular}}
\end{table}
\begin{table}[!ht]
\caption{SC for image datasets, comparing \dclam to baselines with different encoder/decoder architectures. Best for each dataset is in bold. See text for details. {\em Higher is better.}}
\label{tab:q1-q2-images}
\label{tab:q1-q2-images-rebuttal}
\centering
{\footnotesize\begin{tabular}{l|lll|lll|ll}
\toprule
Dataset & \multicolumn{3}{c|}{Convolutional AE} & \multicolumn{3}{c|}{ResNet AE} & \multicolumn{2}{c}{EAE} \\
\cmidrule(lr){2-9}
& DCEC & DEKM & \dclam & DCEC & DEKM & \dclam & EDC & \dclam \\
\midrule
FM  
& \textcolor{black}{0.923} & \textcolor{black}{0.785} & \textcolor{black}{\textbf{0.970}}  
& \textcolor{black}{0.824} & \textcolor{black}{0.742} & \textcolor{black}{0.922}
& 0.521 & 0.715 \\

C-10 
& 0.787 & \textcolor{black}{0.622} & \textbf{0.863}
& \textcolor{black}{0.667} & \textcolor{black}{0.461} & \textcolor{black}{0.697}
& 0.541 & 0.731 \\

C-100 
& \textcolor{black}{0.572} & \textcolor{black}{0.047} & \textcolor{black}{0.598} 
& \textcolor{black}{0.557} & \textcolor{black}{0.036} & \textcolor{black}{\textbf{0.921}}
& 0.337 & 0.636 \\

USPS 
& \textcolor{black}{\textbf{0.935}} & \textcolor{black}{0.882} & \textcolor{black}{0.914}
& \textcolor{black}{0.909} & \textcolor{black}{0.843} & \textcolor{black}{0.914}
& 0.491 & 0.911 \\

STL 
& \textcolor{black}{0.766} & \textcolor{black}{0.745} & 0.919
& \textcolor{black}{0.812} & \textcolor{black}{0.804} & \textcolor{black}{0.865}
& 0.431 & \textbf{0.923} \\

CBird
& 0.386 & 0.018 & \textbf{0.448}
& 0.282 & 0.035 & 0.377
& 0.188 & 0.446 \\
\bottomrule
\end{tabular}}
\end{table}
%

\paragraph{Effect of AE Architecture:}
\cref{tab:q1-q2-images}
shows that the performance improvement achieved by \dclam is independent of the Autoencoder (AE) architecture choice.
\dclam with {\em all three architectures} -- CAE, EAE, and RAE -- consistently outperforms their respective baselines, DCEC, DEKM and EDCWRN with similar architecture. That is, within each type of AE, \dclam has better results than DCEC and DEKM, or EDC. This not only underscores the superiority of the internal algorithm of \dclam over the corresponding baselines but also suggests the potential for further improvement with some more advanced AE architecture.

\begin{figure}[!ht]
\begin{center}
\includegraphics[width=.08\columnwidth]{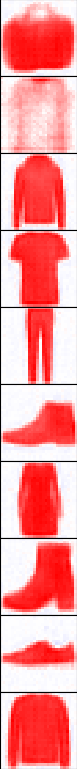}
\!\!\!
\includegraphics[width=.08\columnwidth]{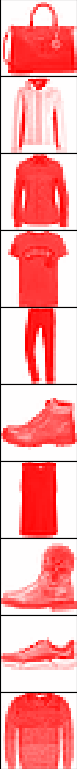}
\!\!\!
\includegraphics[width=.08\columnwidth]{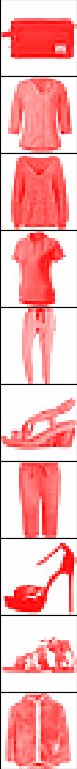}
\hskip 30pt
\includegraphics[width=.08\columnwidth]{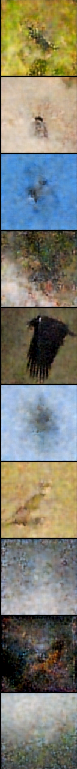}
\!\!\!
\includegraphics[width=.08\columnwidth]{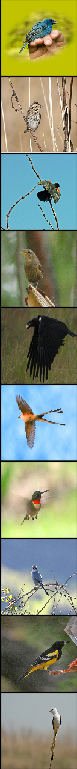}
\!\!\!
\includegraphics[width=.08\columnwidth]{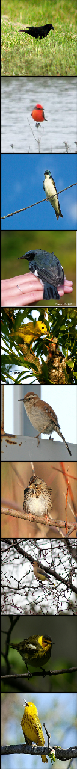}
\end{center}
\caption{Visualizing the decoded images for the learned cluster prototypes (leftmost column in block) and the corresponding closest (center column in block) and farthest (right column in block) members in each cluster for Fashion MNIST (left block) and Caltech Birds (right block).}
\label{fig:mcf}
\end{figure}

\paragraph{Qualitative Evaluation:}
We qualitatively evaluate the latent cluster prototypes found by \dclam in \cref{fig:mcf} for Fashion MNIST (10 clusters) and Caltech Birds (10 out of 200 clusters). The figure shows the decoded prototypes or cluster centers, i.e., $\decoder(\rho_i)$, as well as their corresponding decoded closest and farthest cluster members ( as measured in the latent space) from the centers. Generally, the prototypes $\rho_i$ form an average image that matches the closest images well. The farthest cluster members still appear similar to their prototypes in most cases, with some exceptions: (i)~In the 7th row for FMNIST an image that looks like a pant is grouped with dresses though the overall image shape is still similar. (ii)~In the 5th row for CBird, the memory and the closest image are very similar but the farthest image appears significantly different.

In addition to the above, we discuss our thorough empirical evaluation in \cref{asec:emp}, reporting various clustering metrics, visualizing the evolution of the latent memories (cluster centers), studying the impact of latent dimensionality, and further details of selecting the best results via Pareto analysis along the SC and RL axes.

\section{Discussion}
We introduce a fresh integration of associative memories within an innovative deep clustering algorithm \dclam that leverages the energy-based attractor dynamics in latent space. Our findings demonstrate that \dclam significantly surpasses standard prototype-based and existing deep clustering methods. 
Our future work aims to extend it to multimodal deep clustering. 
Leveraging its flexibility to add other encoder/decoder frameworks with \dclam, we aim to explore transformer-based AE approaches. Additionally, 
we plan to explore how the energy landscape may help estimate the number of clusters directly from the data. 

\bibliography{dclam}
\bibliographystyle{unsrtnat}

\newpage
\appendix
\onecolumn
\section{Experimental Details}
\label{asec:exp}

\subsection{Dataset details} 
\label{asec:exp-details:data}
To evaluate \dclam, we conducted our experiments on eight standard benchmark datasets, including USPS
\citep{hull1994database}, Fashion-MNIST
\citep{xiao2017fashion}, CIFAR-10 and CIFAR-100
\citep{krizhevsky2009learning}, STL-10
\citep{coates2011analysis}, Caltech\_birds2010
\citep{WelinderEtal2010}, 20-NG from Sklearn
and Reuters-10k from TensorFlow datasets.
The later two are text datasets, whereas the others are image datasets. For both text datasets, we calculate TFIDF~\citep{ref1} features based on the 2000 most frequent words, following a similar approach as \citet{golzari2023edcwrn}. 
However, 
we consider the original number of categories as the true number of clusters, which is 46 for Reuters-10k and 20 for 20-NG. For Caltech\_birds2010, as there are images of various shapes, we resize all images to (128, 128, 3) for uniformity and ease of implementation. Table \ref{tab:data} provides the statistics for the datasets used in our experiments.
\begin{table}[ht]
\caption{Descriptions of datasets}
\label{tab:data}
\begin{center}
{\footnotesize
\begin{tabular}{llcccc}
\toprule
Dataset & Short name & \# Points & \ Shape & \# Classes & \# Type \\
\midrule
Fashion MNIST & FM & 60000 & (28, 28, 1) & 10 & Image\\
CIFAR-10 & C-10 & 50000 & (32, 32, 3) & 10 & Image\\
CIFAR-100 & C-100 & 50000 & (32, 32, 3) & 100 & Image\\
USPS & USPS & 2007 & (16, 16, 1) & 10 & Image\\
STL-10 & STL & 5000 & (96, 96, 3) & 10 & Image\\
Caltech\_birds2010 & CBird & 3000 & (128, 128, 3) & 200 & Image\\
\midrule
Reuters-10k & R-10k & 11228 & 2000 & 46 & Text\\
20-NG & 20NG & 18846 & 2000 & 20 & Text \\
\bottomrule
\end{tabular}}
\end{center}
\end{table}

\subsection{Metrics used} \label{asec:expt-details:metric}
To assess the performance of \dclam, we utilize the Silhouette Coefficient (SC)~\citep{rousseeuw1987silhouettes} as an unsupervised metric for measuring clustering quality. SC scores range from $-1$ to 1, where 1 indicates perfect clustering and $-1$ indicates completely incorrect labels. A score close to 0 suggests the existence of overlapping clusters. We also employ Normalized Mutual Information (NMI)~\citep{vinh2009information} \& Adjusted Rand Index (ARI)~\citep{hubert1985comparing} to evaluate the alignment between the partition obtained by \dclam and the ground truth clustering labels. NMI scores range from 0 (completely incorrect) to 1 (perfect clustering) and  ARI scores range from $-1$ to $1$ where the interpretation is same as SC. 
Additionally, we compute Reconstruction Loss (RL), representing the mean squared error between original and reconstructed points, where lower is better. Entropy (ETP)~\citep{bein2006entropy} and Cluster Size (CS) are computed to assess cluster balance. In clustering, higher entropy (the highest value is $\log_2(k)$ for each dataset, where $k$ is the number of true clusters) indicates more balanced clusters, while lower values suggest potential imbalance, possibly involving singleton or very small clusters. Entropy ($H(X)$) is calculated based on the distribution of data points across clusters:
\[
H(X) = -\sum_{i=1}^{k} P(C_i) \; \log_2 P(C_i)
\]
where, \(P(C_i)\) is the proportion of data points in cluster \(C_i\) relative to the total number of data points. Cluster Size (CS) indicates the largest and smallest clusters (in terms of the number of data points) identified in the dataset (more balanced clustering is better).

\subsection{Parameter settings} 
\label{asec:exp-details:parameter}
 
For Convolutional AE or CAE, for $k$-means, Agglomerative, \clam, DCEC, DEKM, and \dclam, we adopt an architecture identical to DCEC. The encoder network structure follows conv$^5_{32}$ $\to$ conv$^5_{64}$ $\to$ conv$^3_{128}$ $\to$ FC$_{d}$, where conv$^k_{n}$ represents a convolutional layer with $n$ filters and a kernel size of $k \times k$, and FC$_d$ denotes a fully connected layer of dimension $d$. Here, $d$ is the number of true clusters in the dataset, and serves as the latent dimension. The decoder mirrors the encoder.

The ResNet AE or RAE approach draws inspiration from the standard ResNet block described by~\citet{wickramasinghe2021resnet}.
For DCEC, DEKM, and \dclam a streamlined configuration is employed using two filters with sizes 32 and 64.
The size of the embedded representation is maintained at $d$, corresponding to the number of clusters in the dataset, as in the previous setup. In this experiment, the number of repeating layers in the ResNet block is set to 2. To enhance model performance, batch normalization and leakyReLU are incorporated. For a given number of repeats ($f$), the total number of hidden layers is calculated as 2 + ($f$ * number of filters), resulting in 6 layers in our case. 

The EDCWRN AE or EAE, is that from \citet{golzari2023edcwrn}, so for both EDC and \dclam, we follow the proposed architecture, where the encoder network is configured as a fully connected multilayer perceptron (MLP) with dimensions $i$-500-500-2000-$d$ for all datasets, where $i$ represents the dimension of the input space (features), and $d$ is the number of clusters in the dataset. Similarly, the decoder network mirrors the encoder, constituting an MLP with dimensions $d$-2000-500-500-$i$. All internal layers, except for the input, output, and embedding layers, use the ReLU activation function.

All three architectures described above are pretrained end-to-end for 100 epochs using Adam~\citep{kingma2014adam} with default parameters. {\color{black} 
The number of clusters $k$ is not a hyperparameter, but rather is taken as the true number of classes in each dataset.
Also, as noted above, we set the latent dimensionality $d$ (or $m$) the same as the number of true classes $k$ in the dataset, i.e., $m=d=k$. }

\subsection{Implementation details} \label{asec:expt-details:implementation}

We implement and evaluate \dclam using the Tensorflow~\citep{abadi2016tensorflow} library while employing {\tt scikit-learn}~\citep{pedregosa2011scikit} for clustering quality metrics. We train our models on a single node with 1 NVIDIA RTX A6000 (48GB RAM) and a 16-core 2.4GHz Intel Xeon(R) Silver 4314 CPU. Hyperparameters are tuned individually for each dataset to maximize the Silhouette Coefficient~\citep{rousseeuw1987silhouettes}. Table \ref{tab:hp_dclam} lists the hyperparameters, their roles, and respective values/ranges. 

\begin{table}[!ht]
\caption{Hyperparameters, their roles and range of values for \dclam.}
\label{tab:hp_dclam}
\begin{center}
\small
{\footnotesize
\begin{tabular}{l l}
\toprule
\multicolumn{1}{c}{\bf Hyperparameter} &\multicolumn{1}{c}{\bf Used Values} \\
\midrule
Inverse temperature, $\beta$ & $[10^{-5}, ..., 5]$ \\
\midrule
Batch size  & $[16, 32, 64, 128, 256]$ \\
Initial learning rate (AM), $\epsilon_{am}$ & $[10^{-4}, 10^{-3}, 10^{-2}, 10^{-1}]$ \\
Initial learning rate (AE), $\epsilon_{ae}$ & $[10^{-7}, 10^{-6}, 10^{-5}, 10^{-4}, 10^{-3}, 10^{-2}, 10^{-1}]$ \\
Reduce LR patience (epochs) & $[5, 10, 15]$ \\
\bottomrule
\end{tabular}}
\end{center}
\end{table}

For baseline schemes like $k$-means and agglomerative, we use the {\tt scikit-learn} library implementation, adjusting hyperparameters for optimal performance on each dataset. For DCEC~\citep{guo2017deep} and DEKM~\citep{guo2021deep}, we leverage their Tensorflow implementation\footnote{https://github.com/XifengGuo/DCEC} \footnote{https://github.com/spdj2271/DEKM/blob/main/DEKM.py}, for EDCWRN~\citep{golzari2023edcwrn}, we utilze their Python implementation\footnote{https://github.com/Amin-Golzari-Oskouei/EDICWRN}, and for \clam~\citep{saha2023end} we use their Tensorflow implementation\footnote{https://github.com/bsaha205/clam}.
Likewise we use the author provide implementations for SCAN~\citep{van2020scan}~\footnote{https://github.com/wvangansbeke/Unsupervised-Classification} and NNM~\citep{dang2021nearest}\footnote{https://github.com/ZhiyuanDang/NNM}.

\subsection{Hyperparameters for \dclam} \label{asec:add-exp:dclam-hpo}
\begin{figure}[!ht]
\centering
\begin{subfigure}[t]{0.49\textwidth}
\centering
    \includegraphics[width=1\textwidth,height=2in]{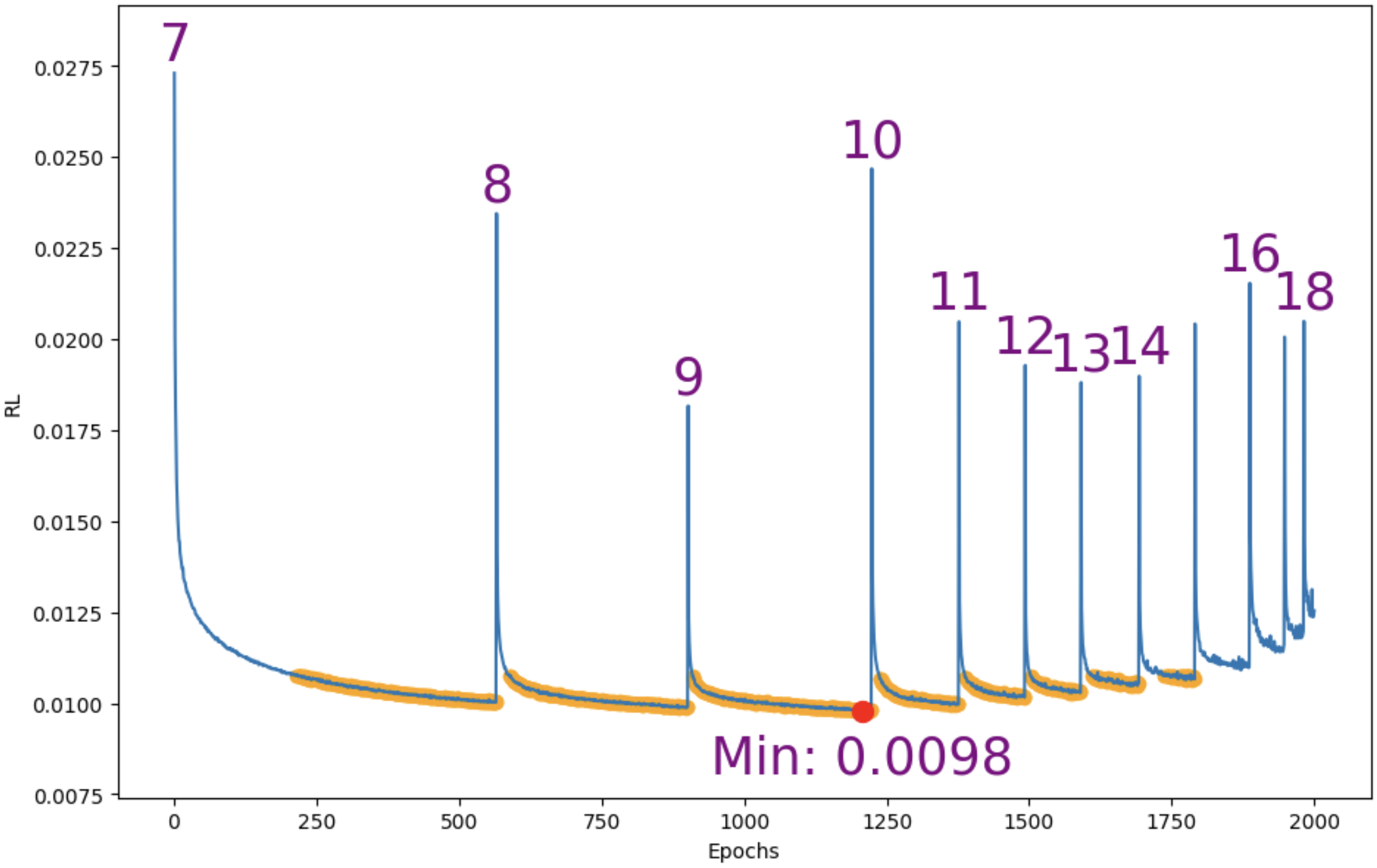}
  \caption{RL vs Training}
  \label{fig:st:fm-st-rl}
\end{subfigure}%
~
\begin{subfigure}[t]{0.49\textwidth}
\centering
  \includegraphics[width=1\textwidth, height=2in]{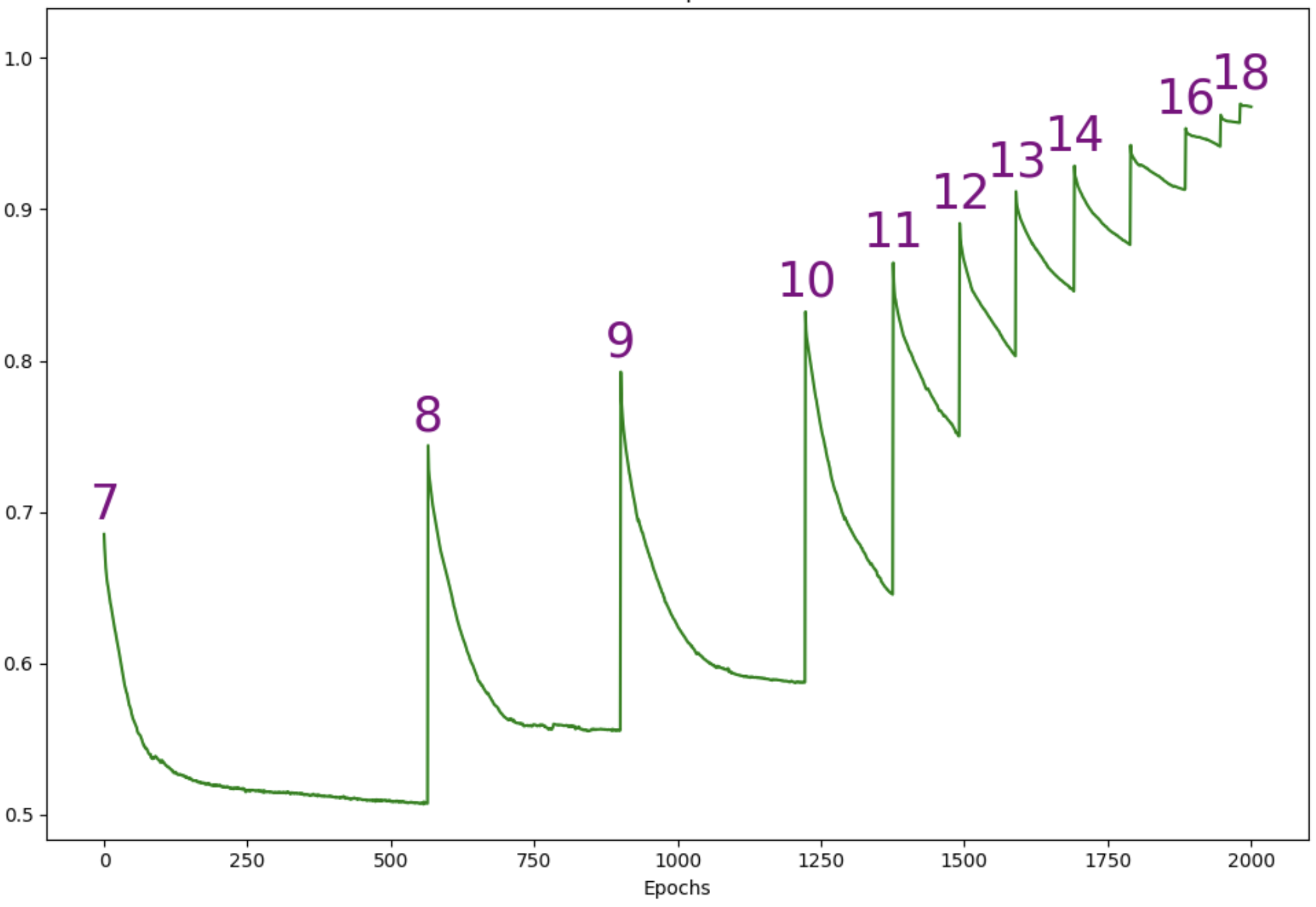}
  \caption{SC vs Training}
  \label{fig:st:fm-st-sc}
\end{subfigure}
\caption{Reconstruction loss (RL) and clustering quality (SC) for varying number of steps (T) for FMNIST. Red point in \cref{fig:st:fm-st-rl} indicates the lowest reconstruction loss and orange points indicate the reconstruction loss within 10\% of this lowest reconstruction loss.
} 
\label{fig:st-plots}
\end{figure}

We extensively tune all hyperparameters (Table \ref{tab:hp_dclam}) for the optimal results in \dclam. We found that the inverse temperature $\beta$ serves as the most critical hyperparameter, which we explore in the range of $[10^{-5}, ..., 5]$ for tuning. 
We employ curriculum learning for the number of AM steps $T$ from the data, where $T$ starts with a small value (e.g., 1) and increases up to 20 based on the reconstruction loss. \cref{fig:st-plots} visualizes this curriculum learning for $T$ for FMNIST dataset where $T$ starts with 7 and ends at 18.
We employ the Adam optimizer while keeping separate initial learning rates for the AM and AE networks. If the training loss does not improve for a certain number of epochs, we decrease the learning rate by a factor of 0.8. We do this for a certain patience value (curriculum patience) after that we increase $T$ by 1. This process continues
until the training loss reaches a minimum threshold ($10^{-9}$) or $T$ reaches its maximum value ($20$).  In \cref{fig:st-plots}, the red point indicates the lowest training loss (reconstruction error) and orange points indicate the reconstruction losses within 10\% of the lowest reconstruction loss. We can see that $T=14$ has reconstruction loss within 10\% of the lowest reconstruction loss and a high value of silhouette coefficient $(> 0.9)$. In this case, we select 14 as the optimal value of $T$ as it has a good trade-off between the reconstruction loss and silhouette coefficient. By doing curriculum learning, we avoid treating $T$ as a separate  hyperparameter.
The best hyperparameter values for various datasets for \dclam are detailed in Table \ref{tab:bhp-rae}.

\begin{table}[!ht]
\caption{Best hyperparameters for different datasets for \dclam. `-' denotes NA.
}
\label{tab:bhp-rae}
\begin{center}
\resizebox{\textwidth}{!}
{\small
\begin{tabular}{l|ccc|ccc|ccc|ccc|ccc|ccc}
\toprule
\multirow{2}{*}{\bf Dataset}  &\multicolumn{3}{c}{\bf Inverse temperature, $\beta$} &\multicolumn{3}{c}{\bf Layers, $T$} &\multicolumn{3}{c}{\bf Batch size} &\multicolumn{3}{c}{\bf Learning rate(AM)}  &\multicolumn{3}{c}{\bf Learning rate ($\encoder$)} & \multicolumn{3}{c}{\bf Learning rate ($\decoder$)}\\
\cmidrule{2-19}
& CAE & RAE & EAE & CAE & RAE & EAE & CAE & RAE & EAE & CAE & RAE & EAE & CAE & RAE & EAE & CAE & RAE & EAE \\
\midrule
FM & 0.5 & 0.09 & 0.7 & 15 & 15 & 10 & 64 & 64 & 64 & 0.001 & 0.001 & 0.1 & 0.000001 & 0.000001 & 0.000001 & 0.001 & 0.001 & 0.001\\

C-10 & 2 & 0.02 & 0.5 & 15 & 15 & 12 & 64 & 64 & 64 & 0.001 & 0.001 & 0.01 & 0.000001 &  0.000001 &  0.000001 & 0.001 & 0.001 & 0.001\\

C-100 & 1 & 0.005 & 5 & 10 & 10 & 10 & 64 & 64 & 64 & 0.001 & 0.001 & 0.001 & 0.000001 & 0.000001 & 0.000001 & 0.001 & 0.001 & 0.001\\

USPS & 0.5 & 1 & 1 & 15 & 10 & 15 & 64 & 64 & 32 & 0.01 & 0.01 & 0.1 & 0.000001 & 0.000001 & 0.000001  & 0.001 & 0.001 & 0.01\\

STL & 0.5 & 0.003 & 0.1 & 15 & 10 & 12 & 64 & 64 & 128 & 0.001 & 0.01 & 0.1 & 0.000001 & 0.000001 & 0.000001 & 0.001 & 0.001 & 0.0001\\

CBird & 0.05 & 0.00015 & 0.005 & 15 & 10 & 15 & 64 & 64 & 64 & 0.01 & 0.001 & 0.1 & 0.000001 & 0.000001 & 0.000001 & 0.001 & 0.001 & 0.001\\
\midrule
R-10K & - & - & 10 & - & - & 10 & - & - & 64 & - & - & 0.01 & - & - & 0.000001 & - & - & 0.1 \\
20-NG & - & - & 1.5 & - & - & 15 & - & - & 64 & - & - & 0.1 & - & - & 0.000001 & - & - & 0.1 \\
\bottomrule
\end{tabular}
}
\end{center}
\end{table}

\subsection{Hyperparameters for Baselines} \label{asec:add-exp:baseline-hpo}
We compare \dclam with baseline clustering schemes $k$-means and agglomerative from {\tt scikit-learn}, \clam, DCEC, DEKM and EDCWRN. For $k$-means and agglomerative, we perform a comprehensive search for tuning different hyperparameters available in {\tt scikit-learn} and pick the best results. For DCEC, DEKM and EDCWRN, we tuned all related hyperparams to obtain the best SC (for 10\% RRL) and best RL (for 10\% of best SC). For \clam we precisely replicate the hyperparameter search criteria outlined in its respective paper, which closely aligns with our approach for \dclam, as detailed in \cref{tab:hp_dclam}. \cref{tab:hp_baselines} provides a brief description of the hyperparameters and their roles in the baseline schemes. 

\begin{table}[!ht]
\caption{Hyperparameters (HPs), their roles and range of values for the baseline clustering schemes.}
\label{tab:hp_baselines}
\begin{center}
{\scriptsize
\begin{tabular}{l l| p{5.5cm}| p{3.5cm}}
\toprule
\multicolumn{1}{l}{\bf Baseline}  &\multicolumn{1}{c}{\bf HP} &\multicolumn{1}{c}{\bf Role} &\multicolumn{1}{c}{\bf Used Values} \\
\midrule
\multirow{3}{*}{$k$-means} & init & Initialization method & [`k-means++', 'random'] \\
& n\_init & Number of time the k-means algorithm will be run  & 1000 \\
\midrule
\multirow{4}{*}{Agglomerative} & affinity & Metric used to compute the linkage & [`euclidean', `l1', `l2', `manhattan', `cosine'] \\
& linkage & Linkage criterion to use & [`single',  `average', `complete', `ward'] \\
\midrule

\multirow{8}{*}{DCEC} & batch\_size & Size of each batch & [64, 128, 256] \\
& maxiter & Maximum number of iteration & [2e4, 3e4] \\
& alpha & Degree of freedom of student's t-distribution & 1 \\
& gamma & Coefficient of clustering loss & [0.01, 0.1, 1, 10]  \\
& update\_interval & Interval at which the predicted and target distributions are updated & [1, 2, 3, 4, 5, 10, 20, 30, 40, 50, 75, 100, 125, 140, 150, 200] \\
& tol & Tolerance rate & 0.001 \\
\midrule

\multirow{8}{*}{DEKM} & batch\_size & Size of each batch & [64, 128, 256] \\
& maxiter & Maximum number of iteration & 2e4 \\
& hidden\_units & Number of latent dimension & Number of true cluster as per dataset \\
& update\_interval & Interval at which the predicted and target distributions are updated & [10, 20, 30, 40, 50, 75, 100, 125, 140, 150, 200] \\
& tol & Tolerance rate & 0.000001 \\
\midrule

\multirow{8}{*}{EDCWRN} & batch\_size & Size of each batch & [64, 128, 256] \\
& maxiter\_pretraining & Maximum number of iteration in pertaining & 500*batch\_size \\
& maxiter\_clustering & Maximum number of iteration in clustering & [8000, 16000, 24000] \\
& gamma & Coefficient of clustering loss & [0.01, 0.1, 1, 10] \\
& update\_interval & Interval at which the predicted and target distributions are updated & [1, 2, 3, 4, 5, 10, 20, 30, 40, 50, 75, 100, 125, 140, 150, 200] \\
& tol & Tolerance rate & 0.0001 \\
\midrule
\multirow{12}{*}\clam & $\beta$ & Inverse temperature & [$10^{-5}$ - 5] \\
& $T = \nicefrac{1}{\alpha} =\nicefrac{\tau}{dt}$ & Number of layers & [2-20] \\
& batch\_size & Size of each batch & [8, 16, 32, 64, 128, 256] \\
& $\epsilon$ & Adam initial learning rate & [$10^{-4}$, $10^{-3}$, $10^{-2}$, $10^{-1}$] \\
& $\epsilon$\_factor & Reduce LR by factor & 0.8 \\
& $\epsilon$\_patience & Reduce LR patience (epochs) & 5 \\
& $\epsilon$\_min& Minimum LR & $10^{-5}$ \\
& $\epsilon$\_loss\_threshold & Reduce LR loss threshold & $10^{-3}$ \\
& max\_epochs & Maximum Number of epochs & 200 \\
& restart & Number of restart & 10 \\
& mask\_prob & Mask probability & [0.1, 0.12, 0.15, 0.2, 0.25, 0.3] \\
& mask\_val & Mask value & [`mean', `min', `max'] \\
\bottomrule
\end{tabular}
}
\end{center}
\end{table}

\section{Detailed and Additional Experimental Results} \label{asec:emp}

\subsection{Pretrained losses for all architecture and all datasets}
\cref{tab:pretrained} records the pretrained reconstruction losses (RL) for all architectures and all datasets. These are the base RL values $RL_p$ used when computing RRL in \cref{eq:rrl}.

\begin{table}[!ht]
\color{black}
\caption{Per-dataset, per-architecture pretrained loss. Note abbreviations Conv-AE$\to$CAE, Resnet-AE$\to$RAE, and EDCWRN-AE$\to$EAE. Further, `-' denotes NA.}
\label{tab:pretrained}
\centering
{\footnotesize\begin{tabular}{l|ccc}
\toprule
\multirow{2}{*}{\bf Dataset}  &\multicolumn{3}{c}{\bf Architecture}\\
\cmidrule{2-4}
& CAE & RAE & EAE\\
\midrule
FM & 0.0122 & 0.0083 & 0.0087  \\
C-10 & 0.0220 & 0.0180 & 0.0167  \\
C-100 & 0.0070 & 0.0040 & 0.0096  \\
USPS & 0.0019 & 0.0023 & 0.0005  \\
STL & 0.0179 & 0.0173 & 0.0206  \\
CBird & 0.0055 & 0.0036 & 0.0187  \\
R-10K & - & - & 0.0010  \\
20NG & - & - & 0.0008  \\

\bottomrule
\end{tabular}}
\end{table}

\begin{table*}[!ht]
\small 
\caption{{\bf Metrics obtained by \dclam and baselines corresponding to the best SC (RL within 10\% of the pretrained AE loss).} The best performance for each dataset is in {\bf boldface}. (note abbreviations DCEC$\to$DC, EDCWRN$\to$EDC, Entropy$\to$ETP, Cluster-size$\to$CS, No-AE$\to$NAE, Conv-AE$\to$CAE, EDCWRN-AE$\to$EAE, Resnet-AE$\to$RAE). `-' denotes NA. x\txsps{\down} indicates negative RRL which means the RL of the method is x\% less than the pretrained AE loss.
}
\label{tab:scgt}
\label{tab:scgt-rebuttal}
\begin{center}
\resizebox{\textwidth}{!}
{\footnotesize
\begin{tabular}{lc|cc|cc|cc|cc|cc|c|ccc}
\toprule
\multirow{2}{*}{\bf Data}  &\multirow{2}{*}{\bf Met} &\multicolumn{2}{c}{\bf Kmeans} &\multicolumn{2}{c}{\bf Agglo} &\multicolumn{2}{c}{\bf \clam} &\multicolumn{2}{c}{\bf DC} 
&\multicolumn{2}{c}{\bf DEKM} 
&\multirow{2}{*}{\bf EDC} 
& \multicolumn{3}{c}{\bf \dclam}\\ 
\cmidrule{3-12} \cmidrule{14-16}
& & \bf NAE & \bf CAE & \bf NAE & \bf CAE & \bf NAE & \bf CAE & \bf CAE & \bf RAE & \bf CAE & \bf RAE & & \bf CAE & \bf EAE & \bf RAE\\

\midrule
\multirow{5}{*}{FM} & SC & 0.154 & 0.257 & 0.109 & 0.201 & 0.158 & 0.279 & \textcolor{black}{0.923} & \textcolor{black}{0.726} & \textcolor{black}{0.260} & \textcolor{black}{0.258} & 0.483 & \textcolor{black}{\textbf{0.970}} & 0.663 & \textcolor{black}{0.712} \\
& NMI & 0.511 & 0.643 & 0.534 & 0.624 & 0.521 & 0.622 & \textcolor{black}{0.558} & 0.624  & \textcolor{black}{0.551} & \textcolor{black}{0.609} & 0.495 & 0.472 & 0.511 & \textcolor{black}{0.488}  \\
& RL & - & 0.0122 & - & 0.0122 & - & 0.0122 & 0.0134 & \textcolor{black}{0.0091}  & \textcolor{black}{0.0105} & \textcolor{black}{0.0097} & 0.0096 & 0.0120 & 0.0096 & 0.0091\\
& RRL & - & 0.0 & - & 0.0 & - & 0.0 & 9.8 & \textcolor{black}{9.6}  & \textcolor{black}{\textbf{13.9\txsps{\down}}} & \textcolor{black}{18.1} & 10 & 1.6\txsps{\down} & 10 & 9.6 \\
& ETP & 3.17 & 3.17 & 3.14 & 3.2  & 2.81 & 2.80 & 3.23 & 3.23 & 3.14 & 3.15 & 3.11 & 2.83 & 3.14 & \textcolor{black}{3.11} \\
& CS  & 9617-2361  & 11145-2744 & 11830-1860  & 10298-2544 & 19032-1524  & 15679-2 & \textcolor{black}{10208-2733} & \textcolor{black}{8914-3338}  & \textcolor{black}{12360-2310} & \textcolor{black}{10974-2724} & 12118-1478 & \textcolor{black}{20448-504} & 11734-2251 & \textcolor{black}{15906-2082} \\
\midrule

\multirow{5}{*}{C-10} & SC & 0.050 & 0.084 & 0.158 & 0.372 & 0.073 & 0.208 & 0.787 & \textcolor{black}{0.659} & \textcolor{black}{0.116} & \textcolor{black}{0.082} & 0.511 & \textbf{0.863} & 0.632 & \textcolor{black}{0.697} \\
& NMI & 0.078 & 0.122 & 0.0005 & 0.0004 & 0.073 & 0.015 & 0.074 & 0.094 & 0.123 & \textcolor{black}{0.129} & 0.112 & 0.075 & 0.061 & 0.079 \\
& RL & - & 0.0220 & - & 0.0220 & - & 0.0220 & 0.0241 & \textcolor{black}{0.0197} & \textcolor{black}{0.0239} & \textcolor{black}{0.0199} & 0.0184 & \textcolor{black}{0.0178} & 0.0184 & \textcolor{black}{0.0170} \\
& RRL & - & 0.0 & - & 0.0 & - & 0.0 & 9.6 & \textcolor{black}{8.9} & \textcolor{black}{8.6} & \textcolor{black}{11.1} & 10 & \textcolor{black}{\textbf{19.5\txsps{\down}}} & 10 & \textcolor{black}{5\txsps{\down}} \\
& ETP & 3.27 & 3.19 & 0.006 & 0.003 & 2.50 & 0.24 & 3.22 & 2.99 & \textcolor{black}{3.25} & 3.15 & 3.24 & 2.83 & 2.65 & \textcolor{black}{2.81} \\
& CS & 7105-2734 & 9779-2524 & 49979-1 & 49991-1 & 23544-582 & 48234-1 & 8511-2610 & \textcolor{black}{11229-1724} &  \textcolor{black}{7905-3245} & 11731-2107 & 8198-2632 & \textcolor{black}{17521-425} & 13771-570 & \textcolor{black}{17121-569} \\
\midrule

\multirow{5}{*}{C-100} & SC & 0.015 & -0.020 & 0.028 & 0.149 & 0.018 & 0.053 & \textcolor{black}{0.314} & \textcolor{black}{0.470} & -0.007 & \textcolor{black}{-0.016} & 0.311 & \textcolor{black}{\textbf{0.598}} & 0.536 & \textcolor{black}{0.482} \\
& NMI & 0.161 & 0.183 & 0.036 & 0.004 & 0.153 & 0.156 & \textcolor{black}{0.104} & 0.119 & \textcolor{black}{0.238} & 0.184 & 0.181 & 0.110 & 0.202 & 0.125 \\
& RL & - & 0.0070 & - & 0.0070 & - & 0.0070 & \textcolor{black}{0.0059} & \textcolor{black}{0.0043} & \textcolor{black}{0.0046}  & \textcolor{black}{0.0041} & 0.0106 & \textcolor{black}{0.0069} & 0.0099 & 0.0044 \\
& RRL & - & 0.0 & - & 0.0 & - & 0.0 & \textcolor{black}{15.7\txsps{\down}} & \textcolor{black}{7.5} & \textcolor{black}{\textbf{34.3\txsps{\down}}} & \textcolor{black}{2.5} & 4.3 & \textcolor{black}{1.4\txsps{\down}} & 3.1 & 10 \\
& ETP & 6.53 & 6.48 & 0.940 & 0.052 & 6.51 & 4.38 & \textcolor{black}{5.54} & \textcolor{black}{4.52} & \textcolor{black}{6.25} & 6.46 & 6.49 & 4.16 & 5.85 & \textcolor{black}{4.03} \\
& CS & 1160-129 & 1395-23 & 38814-1 & 49834-1 & 1317-177 & 13950-11 & \textcolor{black}{2255-325} & \textcolor{black}{12721-122} & \textcolor{black}{1715-5} & \textcolor{black}{1322-87} & 999-216 & \textcolor{black}{11085-112} & 4116-32 & \textcolor{black}{8245-112}\\
\midrule

\multirow{5}{*}{USPS} & SC & 0.143 & 0.195 & 0.124 & 0.158 & 0.144 & 0.194 & \textcolor{black}{\textbf{0.935}} & \textcolor{black}{0.758} & \textcolor{black}{0.195} & \textcolor{black}{0.217} & 0.461 & \textcolor{black}{0.820} & 0.872 & \textcolor{black}{0.891} \\
& NMI & 0.573 & 0.628  & 0.627 & 0.680  & 0.475 & 0.619 & \textcolor{black}{0.732} & \textcolor{black}{0.761} & \textcolor{black}{0.631} & \textcolor{black}{0.665} & 0.467 & 0.444 & 0.347 & 0.428 \\
& RL & - & 0.0019 & - & 0.0019 & - & 0.0019 & \textcolor{black}{0.0018} & \textcolor{black}{0.0019} & \textcolor{black}{0.0020} & 0.0024 & 0.0005 & 0.0021 & 0.0006 & 0.0025 \\
& RRL & - & 0.0 & - & 0.0 & - & 0.0 & \textcolor{black}{5.3\txsps{\down}} & \textcolor{black}{\textbf{17.4\txsps{\down}}} & \textcolor{black}{5.3} & 4.3 & 0.0 & 10 & 10 & 8.7 \\
& ETP & 3.27 & 3.23 & 3.26 & 3.27 & 3.10 & 3.16 & 3.26 &  \textcolor{black}{3.29} & 3.23 & 3.25 & 3.29 & 3.12 & 2.78 & 2.99 \\
& CS & 284-121 & 359-89 & 333-121 & 328-104 & 420-53 & 375-64 & \textcolor{black}{287-106}  &  \textcolor{black}{281-138} & \textcolor{black}{379-90} & \textcolor{black}{319-99} & 295-134 & \textcolor{black}{442-71} & 841-76 & \textcolor{black}{524-49}  \\
\midrule

\multirow{5}{*}{STL} & SC & 0.039 & 0.079 & 0.158 & 0.270 & 0.051 & 0.108 &  \textcolor{black}{0.132} & \textcolor{black}{0.259} & \textcolor{black}{0.082} & \textcolor{black}{0.081} & 0.411 & \textcolor{black}{0.475} & \textbf{0.891} & \textcolor{black}{0.615} \\
& NMI & 0.127 & 0.152 & 0.007 & 0.004 & 0.106 & 0.139 & \textcolor{black}{0.180} & \textcolor{black}{0.162} & \textcolor{black}{0.167} & \textcolor{black}{0.167} & 0.066 & \textcolor{black}{0.077} & 0.073 & \textcolor{black}{0.119} \\
& RL & - & 0.0179 & - & 0.0179  & - & 0.0179 & \textcolor{black}{0.0204} & \textcolor{black}{0.0189} & \textcolor{black}{0.0180} & \textcolor{black}{0.0174} & 0.0196 & \textcolor{black}{0.0187} & 0.0227 & 0.0190 \\
& RRL & - & 0.0 & - & 0.0 & - & 0.0 & \textcolor{black}{13.9} & \textcolor{black}{9.2}  & \textcolor{black}{0.6} & \textcolor{black}{0.6} & \textbf{4.9\txsps{\down}} & \textcolor{black}{4.5} & 10 & 9.8 \\
& ETP & 3.26 & 3.25 & 0.069 & 0.025 & 2.43 & 1.4 & \textcolor{black}{3.19} & \textcolor{black}{3.17} & \textcolor{black}{3.21} & \textcolor{black}{3.19} & 2.92 & 2.48 & 2.99 & 2.87 \\
& CS & 764-312 & 830-287 & 4969-1 & 4991-1 & 2586-82 & 3888-38 & \textcolor{black}{931-239} & \textcolor{black}{1003-263} & \textcolor{black}{844-191} & \textcolor{black}{804-258} & 2611-33 & \textcolor{black}{2076-21} & 912-45 & \textcolor{black}{1219-113} \\
\midrule

\multirow{5}{*}{CBird} & SC  & -0.019 & -0.021 & 0.037 & 0.094 & -0.026 & -0.062 & 0.311 & 0.251 & -0.032 & -0.037 & 0.171 & \textbf{0.448} & 0.446 & 0.312 \\
& NMI & 0.412 & 0.353 & 0.206 & 0.132  & 0.423 & 0.485 & 0.347 & 0.299 & 0.372 & 0.370 & 0.471 & 0.221 & 0.467 & 0.211 \\
& RL & - & 0.0055 & - & 0.0055  & - & 0.0055 & 0.0061 & 0.0040 & 0.0055 & 0.0036 & 0.0206 & 0.0060 & 0.0115 & 0.0039 \\
& RRL & - & 0.0 & - & 0.0 & - & 0.0 & 10 & 10 & 0.0 & 0.0 & 10 & 9.1 & \textbf{39\txsps{\down}} & 8.3 \\
& ETP & 6.34 & 5.59 & 2.71 & 0.958  & 6.56 & 7.21 & 5.41 & 5.04 & 5.81 & 5.80 & 7.41 & 5.68 & 7.02 & 5.07 \\
& CS & 131-1 & 245-1 & 1722-1 & 2773-1 & 101-2 & 99-2 & 241-1 & 291-1 & 168-1 & 197-1 & 37-2 & 213-1 & 99-1 & 676-1 \\
\midrule

\multirow{5}{*}{R-10k} & SC & -0.010 & - & 0.114 & - & -0.002 & - & - & - & - & - &  0.023 & - & \textbf{0.564} & -  \\
& NMI & 0.398 & - & 0.012 & - & 0.383 & - & - & - & - & - & 0.152 &  - & 0.367 & -\\
& RL & - & - & - & - & - & - & - & - & - & - & 0.0011 & - & 0.0011 & -\\
& RRL & - & - & - & - & - & - & - & -  & - & - & \textbf{10} & - & \textbf{10} & - \\
& ETP & 5.13 & - & 0.072 & - & 5.10 & - & - & - & - & - & 5.51 & - & 4.77 & -\\
& CS & 916-20 & - & 11172-1 & - & 885-18 & - & - & - & - & - & 721-51 & - & 1046-1 & -\\
\midrule

\multirow{5}{*}{20NG} & SC & -0.021 & - & 0.114 & - & -0.008 & - & - & - & - & - & 0.101 & - & \textbf{0.197} & -  \\
& NMI & 0.155 & - & 0.003 & - & 0.166 & - & - & - & - & - & 0.019 & - & 0.181 & -  \\
& RL & - & - & - & - & - & - & - & - & - & - & 0.0009 & - & 0.0009 & -\\
& RRL & - & - & - & - & - & - & - & -  & - & - & \textbf{10} & - & \textbf{10} & - \\
& ETP & 4.03 & - & 0.022 & - & 3.86 & - & - & - & - & - & 4.32 & - & 4.21 & -\\
& CS & 2217-107 & - & 18818-1 & - & 3428-26 & - & - & - & - & - & 1131-599 & - & 1812-199 & -\\
\bottomrule
\end{tabular}
}
\end{center}
\end{table*}

\begin{table*}[!ht]
\small 
\caption{{\bf Metrics obtained by \dclam and baselines corresponding to the best RL (SC within 10\% of the best SC of the method).} The best performance for each dataset is in {\bf boldface}. (note abbreviations DCEC$\to$DC, EDCWRN$\to$EDC, Entropy$\to$ETP, Cluster-size$\to$CS, No-AE$\to$NAE, Conv-AE$\to$CAE, EDCWRN-AE$\to$EAE, Resnet-AE$\to$RAE). `-' denotes NA. x\txsps{\down} indicates negative RRL which means the RL of the method is x\% less than the pretrained AE loss.}
\label{tab:rlgt}
\label{tab:rlgt-rebuttal}
\begin{center}
\resizebox{\textwidth}{!}
{\footnotesize
\begin{tabular}{lc|cc|cc|cc|cc|cc|c|ccc}
\toprule
\multirow{2}{*}{\bf Data}  &\multirow{2}{*}{\bf Met} &\multicolumn{2}{c}{\bf Kmeans} &\multicolumn{2}{c}{\bf Agglo} &\multicolumn{2}{c}{\bf \clam} &\multicolumn{2}{c}{\bf DC} 
&\multicolumn{2}{c}{\bf DEKM} 
&\multirow{2}{*}{\bf EDC} 
& \multicolumn{3}{c}{\bf \dclam}\\ 
\cmidrule{3-12} \cmidrule{14-16}
& & \bf NAE & \bf CAE & \bf NAE & \bf CAE & \bf NAE & \bf CAE & \bf CAE & \bf RAE & \bf CAE & \bf RAE & & \bf CAE & \bf EAE & \bf RAE\\

\midrule
\multirow{5}{*}{FM} & SC & 0.154 & 0.257 & 0.109 & 0.201 & 0.158 & 0.279 & \textcolor{black}{0.898} & \textcolor{black}{0.824} & \textcolor{black}{0.785} & \textcolor{black}{0.742} & 0.521 & \textcolor{black}{0.891} & 0.715 & \textcolor{black}{\textbf{0.922}} \\
& NMI & 0.511 & 0.643 & 0.534 & 0.624 & 0.521 & 0.622 & 0.561 & 0.623  & \textcolor{black}{0.571} & \textcolor{black}{0.633} & 0.493 & 0.472 & 0.522 & \textcolor{black}{0.401}  \\
& RL & - & 0.0122 & - & 0.0122 & - & 0.0122 & \textcolor{black}{0.0109} & \textcolor{black}{0.0105}  & \textcolor{black}{0.0514} & \textcolor{black}{0.0516} & 0.0211 & \textcolor{black}{0.0102} & 0.0131 & \textcolor{black}{0.0118}\\
& RRL & - & 0.0 & - &  0.0 & - &  0.0 & \textcolor{black}{9.8\txsps{\down}} & \textcolor{black}{26.5}  & \textcolor{black}{321} & \textcolor{black}{522} & 143 & \textcolor{black}{\textbf{16.4\txsps{\down}}} & 54.0 & \textcolor{black}{42.2} \\
& ETP & 3.17 & 3.17 & 3.14 & 3.2  & 2.81 & 2.80 & \textcolor{black}{3.21} & \textcolor{black}{3.6} & \textcolor{black}{3.15} & 3.16 & 3.09 & 2.83 & 3.16 & 2.98 \\
& CS  & 9617-2361  & 11145-2744 & 11830-1860  & 10298-2544 & 19032-1524  & 15679-2 & \textcolor{black}{11307-2766} & \textcolor{black}{9450-3132}  & \textcolor{black}{12720-2478} & \textcolor{black}{11178-2658} & 13199-1391 & \textcolor{black}{17040-504} & 11886-2148 & \textcolor{black}{11830-1290} \\
\midrule

\multirow{5}{*}{C-10} & SC & 0.050 & 0.084 & 0.158 & 0.372 & 0.073 & 0.208 & \textcolor{black}{0.786} & \textcolor{black}{0.667} & \textcolor{black}{0.622} & \textcolor{black}{0.461} & 0.541 & \textbf{0.809} & 0.731 & \textcolor{black}{0.697} \\
& NMI & 0.078 & 0.122 & 0.0005 & 0.0004 & 0.073  & 0.015 & \textcolor{black}{0.099}& 0.094 & \textcolor{black}{0.092} & \textcolor{black}{0.119} & 0.111 & \textcolor{black}{0.079} & 0.060 & 0.082 \\
& RL & - & 0.0220 & - & 0.0220 & - & 0.0220 & \textcolor{black}{0.0217} & \textcolor{black}{0.0217} & \textcolor{black}{0.0616} & \textcolor{black}{0.0502} & 0.0291 & \textcolor{black}{0.0175} & 0.0252 & \textcolor{black}{0.0171} \\
& RRL & - &  0.0 & - &  0.0 & - &  0.0 & \textcolor{black}{0.9\txsps{\down}} & \textcolor{black}{20}  & \textcolor{black}{180} & \textcolor{black}{179} & 74.3 & \textcolor{black}{\textbf{20.4\txsps{\down}}} & 50.9 & \textcolor{black}{5\txsps{\down}} \\
& ETP & 3.27 & 3.19 & 0.006 & 0.003 & 2.50 & 0.24 & \textcolor{black}{3.15} & 2.99 & \textcolor{black}{2.01} & \textcolor{black}{3.07} & 3.25 & 2.83 & 2.64 & 2.50 \\
& CS & 7105-2734 & 9779-2524 & 49979-1 & 49991-1 & 23544-582 & 48234-1 & \textcolor{black}{7145-4055} & \textcolor{black}{11025-1542} & \textcolor{black}{23420-26} & \textcolor{black}{14530-2505} & 8172-2562 & \textcolor{black}{17520-390} & 14890-120 & \textcolor{black}{17121-455} \\
\midrule

\multirow{5}{*}{C-100} & SC & 0.015 & -0.020 & 0.028 & 0.149 & 0.018 & 0.053 & \textcolor{black}{0.572} & \textcolor{black}{0.557} & \textcolor{black}{0.047} & \textcolor{black}{0.036} & 0.337 & \textcolor{black}{0.540} & 0.617 & \textcolor{black}{\textbf{0.921}} \\
& NMI & 0.161 & 0.183 & 0.036 & 0.004 & 0.153 & 0.156 & \textcolor{black}{0.158} & 0.119 & 0.162 & \textcolor{black}{0.221} & 0.186 & 0.112 & 0.201 & \textcolor{black}{0.094} \\
& RL & - & 0.0070 & - & 0.0070 & - & 0.0070 & \textcolor{black}{0.0083} & \textcolor{black}{0.0047} & \textcolor{black}{0.0679}  & \textcolor{black}{0.0494} & 0.0128 & \textcolor{black}{0.0061} & 0.0092 & \textcolor{black}{0.0051} \\
& RRL & - &  0.0 & - &  0.0 & - & 0.0 & \textcolor{black}{18.6} & \textcolor{black}{17.5}  & \textcolor{black}{870} & \textcolor{black}{1135} & 33.3 & \textcolor{black}{\textbf{12.9\txsps{\down}}} & 4.2\txsps{\down} & \textcolor{black}{27.5} \\
& ETP & 6.53 & 6.48 & 0.940 & 0.052 & 6.51 & 4.38 & \textcolor{black}{5.8} & 4.06 & \textcolor{black}{6.18} & \textcolor{black}{6.11} & 6.51 & 4.02 & 5.83 & 3.22 \\
& CS & 1160-129 & 1395-23 & 38814-1 & 49834-1 & 1317-177 & 13950-11 & \textcolor{black}{2540-115} & \textcolor{black}{13736-12} & \textcolor{black}{1950-10} & \textcolor{black}{1980-10} & 996-156 & \textcolor{black}{11010-25} & 4350-10 & \textcolor{black}{22480-1}\\
\midrule

\multirow{5}{*}{USPS} & SC & 0.143 & 0.195 & 0.124 & 0.158 & 0.144 & 0.194 & \textcolor{black}{\textbf{0.929}} & \textcolor{black}{0.909} & \textcolor{black}{0.882} & \textcolor{black}{0.843} & 0.491 & \textcolor{black}{0.914} & 0.911 & \textcolor{black}{0.914} \\
& NMI & 0.573 & 0.628  & 0.627 & 0.680  & 0.475 & 0.619 & \textcolor{black}{0.717} & 0.736 & \textcolor{black}{0.691} & \textcolor{black}{0.684} & 0.451 & \textcolor{black}{0.477} & 0.339 & 0.437 \\
& RL & - & 0.0019 & - & 0.0019 & - & 0.0019 & \textcolor{black}{0.0014} & \textcolor{black}{0.0029} & \textcolor{black}{0.0487} & \textcolor{black}{0.0558} & 0.0007 & \textcolor{black}{0.0025} & 0.0013 & \textcolor{black}{0.0025} \\
& RRL & - &  0.0 & - &  0.0 & - &  0.0 & \textcolor{black}{\textbf{26.3\txsps{\down}}} & \textcolor{black}{26.1} & \textcolor{black}{2463} & \textcolor{black}{2326} & 40 & \textcolor{black}{31.6} & 160 & \textcolor{black}{8.7} \\
& ETP & 3.27 & 3.23 & 3.26 & 3.27 & 3.10 & 3.16 & 3.27 & 3.27 & 3.24 & \textcolor{black}{3.25} & 3.29 & \textcolor{black}{3.11} & 2.55 & 2.99 \\
& CS & 284-121 & 359-89 & 333-121 & 328-104 & 420-53 & 375-64 & \textcolor{black}{283-106}  & \textcolor{black}{283-127} & \textcolor{black}{334-87} & \textcolor{black}{312-97} & 294-156 & \textcolor{black}{463-35} & 947-27 & \textcolor{black}{513-49}  \\
\midrule

\multirow{5}{*}{STL} & SC & 0.039 & 0.079 & 0.158 & 0.270 & 0.051 & 0.108 & \textcolor{black}{0.766} & \textcolor{black}{0.812} & \textcolor{black}{0.745} & \textcolor{black}{0.804} & 0.431 & 0.919 & \textbf{0.923} & \textcolor{black}{0.865} \\
& NMI & 0.127 & 0.152 & 0.007 & 0.004 & 0.106 & 0.139 & \textcolor{black}{0.181} & \textcolor{black}{0.170} & \textcolor{black}{0.149} & \textcolor{black}{0.152} & 0.065 & 0.144 & 0.072 & 0.107 \\
& RL & - & 0.0179 & - & 0.0179  & - & 0.0179 & \textcolor{black}{0.0242} & \textcolor{black}{0.0310} & \textcolor{black}{0.0711} & \textcolor{black}{0.0578} & 0.0525 & 0.0354 & 0.0263 & \textcolor{black}{0.0255} \\
& RRL & - &  0.0 & - &  0.0 & - &  0.0 & \textcolor{black}{35.8} & \textcolor{black}{79.2}  & \textcolor{black}{297} & \textcolor{black}{234} & 155 & 97.8 & \textbf{27.7} & \textcolor{black}{47.4} \\
& ETP & 3.26 & 3.25 & 0.069 & 0.025 & 2.43 & 1.4 & \textcolor{black}{3.23} & \textcolor{black}{3.26} & \textcolor{black}{1.15} & \textcolor{black}{3.22} & 2.90 & 2.48 & 2.98 & 2.86 \\
& CS & 764-312 & 830-287 & 4969-1 & 4991-1 & 2586-82 & 3888-38 & \textcolor{black}{725-229} & \textcolor{black}{741-299} & \textcolor{black}{4064-16} & \textcolor{black}{821-261} & 2641-23 & 2280-27 & 929-34 & 1466-69 \\
\midrule

\multirow{5}{*}{CBird} & SC  & -0.019 & -0.021 & 0.037 & 0.094 & -0.026 & -0.062 & 0.386 & 0.282 & 0.018 & 0.035 & 0.188 & 0.413 & \textbf{0.441} & 0.377 \\
& NMI & 0.412 & 0.353 & 0.206 & 0.132  & 0.423 & 0.485 & 0.333 & 0.297 & 0.316 & 0.273 & 0.484 & 0.222 & 0.466 & 0.209 \\
& RL & - & 0.0055 & - & 0.0055  & - & 0.0055 & 0.0229 & 0.0139 & 0.0625 & 0.0560 & 0.0377 & 0.0056 & 0.0104 & 0.0039 \\
& RRL & - &  0.0 & - &  0.0 & - &  0.0 & 316 & 286  & 1036 & 1455 & 102 & 1.8 & \textbf{44.4\txsps{\down}} & 8.3 \\
& ETP & 6.34 & 5.59 & 2.71 & 0.958  & 6.56 & 7.21 & 5.51 & 5.03 & 5.16 & 4.47 & 7.43 & 5.68 & 7.01 & 5.06 \\
& CS & 131-1 & 245-1 & 1722-1 & 2773-1 & 101-2 & 99-2 & 248-1 & 297-1 & 312-1 & 519-1 & 35-2 & 211-1 & 100-1 & 701-1 \\
\midrule

\multirow{5}{*}{R-10k} & SC & -0.010 & - & 0.114 & - & -0.002 & - & - & - & - & - &  0.035 & - & \textbf{0.673} & -  \\
& NMI & 0.398 & - & 0.012 & - & 0.383 & - & - & - & - & - & 0.147 &  - & 0.378 & -\\
& RL & - & - & - & - & - & - & - & - & - & - & 0.0016 & - & 0.0022 & -\\
& RRL & - & - & - & - & - & - & - & -  & - & - & \textbf{60} & - & 120 & - \\
& ETP & 5.13 & - & 0.072 & - & 5.10 & - & - & - & - & - & 5.55 & - & 4.79 & -\\
& CS & 916-20 & - & 11172-1 & - & 885-18 & - & - & - & - & - & 727-56 & - & 1026-1 & -\\
\midrule

\multirow{5}{*}{20NG} & SC & -0.021 & - & 0.114 & - & -0.008 & - & - & - & - & - & 0.099 & - & \textbf{0.287} & -  \\
& NMI & 0.155 & - & 0.003 & - & 0.166 & - & - & - & - & - & 0.018 & - & 0.180 & -  \\
& RL & - & - & - & - & - & - & - & - & - & - & 0.0006 & - & 0.0012 & -\\
& RRL & - & - & - & - & - & - & - & -  & - & - & \textbf{25\txsps{\down}} & - & 50 & - \\
& ETP & 4.03 & - & 0.022 & - & 3.86 & - & - & - & - & - & 4.31 & - & 4.19 & -\\
& CS & 2217-107 & - & 18818-1 & - & 3428-26 & - & - & - & - & - & 1142-582 & - & 1809-197 & -\\
\bottomrule
\end{tabular}
}
\end{center}
\end{table*}
%

\begin{table*}[!ht]
\caption{{\bf Metrics obtained by \dclam and baselines corresponding to the best NMI.} The best performance for each dataset is in {\bf boldface}. (note abbreviations DCEC$\to$DC, EDCWRN$\to$EDC, Entropy$\to$ETP, Cluster-size$\to$CS, No-AE$\to$NAE, Conv-AE$\to$CAE, EDCWRN-AE$\to$EAE, Resnet-AE$\to$RAE). `-' denotes NA. 
x\txsps{\down} indicates negative RRL which means the RL of the method is x\% less than the pretrained AE loss.
}
\label{tab:nmigt}
\begin{center}
\resizebox{\textwidth}{!}
{\footnotesize

\begin{tabular}{lc|cc|cc|cc|cc|cc|c|ccc}
\toprule
\multirow{2}{*}{\bf Data}  &\multirow{2}{*}{\bf Met} &\multicolumn{2}{c}{\bf Kmeans} &\multicolumn{2}{c}{\bf Agglo} &\multicolumn{2}{c}{\bf \clam} &\multicolumn{2}{c}{\bf DC} 
&\multicolumn{2}{c}{\bf DEKM} 
&\multirow{2}{*}{\bf EDC} 
& \multicolumn{3}{c}{\bf \dclam}\\ 
\cmidrule{3-12} \cmidrule{14-16}
& & \bf NAE & \bf CAE & \bf NAE & \bf CAE & \bf NAE & \bf CAE & \bf CAE & \bf RAE & \bf CAE & \bf RAE & & \bf CAE & \bf EAE & \bf RAE\\

\midrule
\multirow{5}{*}{FM} & SC & 0.154 & 0.251 & 0.109 & 0.201 & 0.140 & 0.262  & 0.861 & 0.716 & 0.819 & 0.784 & 0.430 & 0.817 & 0.619 & 0.825 \\
& NMI & 0.511 & 0.643 & 0.534 & 0.625 & 0.525 & 0.631 & 0.629 & \textbf{0.668} & 0.586 & 0.639  & 0.457 & 0.610 & 0.534  & 0.597 \\
& RL & - & 0.0122 & - & 0.0122 & - & 0.0122 & 0.0138 & 0.0139 & 0.0574 & 0.0596 & 0.0263 & 0.0406 & 0.0327 & 0.0387 \\
& RRL & - & 0.0 & - & 0.0 & - & 0.0 & 13.1 & 67.5 & 370 & 618 & 202 & 233 & 276 & 366 \\
& ETP & 3.17 & 3.17 & 3.14 & 3.20 & 3.13 & 2.98 & 3.22 & 3.20 & 3.07 & 3.16 & 3.00 & 3.16 & 3.22 & 3.18 \\
& CS & 9617-2361 & 11145-2744 & 11830-1860 & 10298-2544 & 14068-2435 & 15262-2100 & 10886-3030 & 9734-2847 & 12974-1191 & 11023-2652  & 17140-1578 & 11028-2658 & 10332-3054 & 10404-2610 \\
\midrule

\multirow{5}{*}{C-10} & SC & 0.050 & 0.072 & 0.014 & 0.020 & 0.064 & 0.101  & 0.118 & 0.653 & 0.276 & 0.262 & 0.541 & 0.713 & 0.632 & 0.420 \\
& NMI & 0.078 & 0.122 & 0.071 & 0.101 & 0.086 & 0.105  & 0.121 & 0.120 & 0.116 & 0.122 & 0.111 & \textbf{0.123} & 0.114 & 0.119  \\
& RL& - & 0.0220 & - & 0.0220 & - & 0.0220 & 0.0221 & 0.0245 & 0.0426 & 0.0362 & 0.0291 & 0.0403 & 0.0379 & 0.0326 \\
& RRL & - & 0.0 & - & 0.0 & - & 0.0 & 0.5 & 36.1 & 93.6 & 101 & 74.3 & 83.2 & 127 & 81.1 \\
& ETP & 3.27 & 3.19 & 3.17 & 3.02 & 3.23 & 2.21  & 3.07 & 3.21 & 3.19 & 3.11 & 3.25 & 3.18 & 2.98 & 3.28 \\
& CS & 7105-2734 & 9779-2524 & 10505-1650 & 11278-1764 & 9587-2925 & 26395-361 & 11022-3374 & 10235-1968 & 10275-2454 & 13746-2168 & 8172-2562 & 8595-2365  & 10721-289 & 6843-3144 \\
\midrule

\multirow{5}{*}{C-100} & SC & 0.015 & -0.014 & -0.018 & -0.043 & 0.018 & 0.001  & 0.048 & 0.002 & -0.011 & -0.028 & 0.308 & 0.354 & 0.200 & 0.130 \\
& NMI & 0.161 & 0.183 & 0.150 & 0.167 & 0.153 & 0.170 & 0.162 & 0.179 & 0.186 & 0.189 & 0.186 & 0.219 & 0.225 & \textbf{0.239} \\
& RL & - & 0.0070 & - & 0.0070 & - & 0.0070 & 0.0072 & 0.0049 & 0.0112 & 0.0074 & 0.0398 & 0.0257 & 0.0250 & 0.0226 \\
& RRL & - & 0.0 & - & 0.0 & - & 0.0 & 2.9 & 22.5  & 60 & 85 & 315 & 267 & 160 & 465 \\
& ETP & 6.53 & 6.48 & 6.45 & 6.30 & 6.51 & 6.27  & 6.41 & 6.41 & 5.23 & 6.45 & 5.51 & 6.33 & 6.33 & 6.36 \\
& CS & 1160-129 & 1395-23 & 1299-77 & 2308-17 & 1317-177 & 2535-39 & 1623-14 & 1380-21 & 2213-32 & 1440-60 & 996-156 & 1210-5 & 2105-15 & 1315-5 \\
\midrule

\multirow{5}{*}{USPS} & SC & 0.143 & 0.195 & 0.124 & 0.159 & 0.142 & 0.180  & 0.920 & 0.896 & 0.946 & 0.465 & 0.43 & 0.865 & 0.660 & 0.857 \\
& NMI & 0.573 & 0.628 & 0.627 & 0.680 & 0.564 & 0.640  & \textbf{0.737} & 0.736 & 0.728 & 0.701 & 0.451 & 0.689 & 0.583 & 0.660 \\
& RL & - & 0.0019 & - & 0.0019 & - & 0.0019 & 0.0074 & 0.0039 & 0.0748 & 0.0374 & 0.0006 & 0.0451 & 0.0322 & 0.0409 \\
& RRL & - & 0.0 & - & 0.0 & - & 0.0 & 289 & 69.6 & 3836 & 1526 & 20 & 2274 & 6340 & 1678 \\
& ETP & 3.27 & 3.23 & 3.26 & 3.27 & 3.27 & 3.21  & 3.27 & 3.27 & 3.24 & 3.24 & 3.29 & 3.11 & 3.24 & 3.23 \\
& CS & 284-121 & 359-89 & 333-121 & 328-104 & 290-132 & 343-73 & 284-108 & 282-107 & 298-80 & 318-91 & 294-156 & 396-35 & 385-107 & 308-72 \\
\midrule

\multirow{5}{*}{STL} & SC & 0.039 & 0.074 & 0.024 & 0.021 & 0.042 & 0.069 & 0.822 &  0.837 & 0.109 & 0.079 & 0.332 & 0.388 & 0.597 & 0.280 \\
& NMI & 0.127 & 0.152 & 0.121 & 0.138 & 0.130 & 0.169  & \textbf{0.188} & 0.165 & 0.170 & 0.166 & 0.103 & 0.149 & 0.151 & 0.159 \\
& RL & - & 0.0179 & - & 0.0179 & - & 0.0179 & 0.0328 & 0.0362 & 0.0315 & 0.0174 & 0.0433 & 0.0409 & 0.0454 & 0.0364 \\
& RRL & - & 0.0 & - & 0.0 & - & 0.0 & 83.2 & 109  & 76.0 & 0.6 & 110 & 128 & 120 & 110 \\
& ETP & 3.26 & 3.25 & 3.02 & 3.02 & 3.24 & 2.82 & 3.24 & 3.28 & 3.21 & 3.20 & 2.62 & 3.13 & 3.18 & 3.15 \\
& CS & 764-312 & 830-287 & 1379-205 & 1373-130 & 945-317 & 1212-2 & 849-232 & 671-326 & 807-250 & 876-264 & 2173-121 & 982-46 & 929-232 & 938-181 \\
\midrule

\multirow{5}{*}{CBird} & SC & -0.019 & -0.021 & -0.018 & -0.064 & -0.026 & -0.062 & 0.248 & 0.152 & -0.041 & -0.038 & 0.188 & 0.135 & 0.068 & 0.167 \\
& NMI & 0.412 & 0.353 & 0.469 & 0.439 & 0.423 & 0.485 & 0.356 & 0.320 & 0.364 & 0.370 & 0.484 & 0.421 & \textbf{0.493} & 0.385 \\
& RL & - & 0.0055 & - & 0.0055 & - & 0.0055 & 0.0229  & 0.0152 & 0.0066 & 0.0036 & 0.0377 & 0.0255 & 0.0237 & 0.0249 \\
& RRL & - & 0.0 & - & 0.0 & - & 0.0 & 316 & 322 & 20 & 0.0 & 102 & 364 & 26.7 & 592 \\
& ETP & 6.34 & 5.59 & 6.97 & 6.58 & 6.56 & 7.21  & 5.84 & 5.12 & 5.71 & 5.80 & 7.43 & 6.48 & 7.39 & 6.05 \\
& CS & 131-1 & 245-1 & 93-1 & 232-1 & 101-2 & 99-2 & 167-1 & 570-1 & 177-1 & 197-1 & 35-2 & 143-1 & 58-2 & 180-1 \\
\midrule

\multirow{5}{*}{R-10k} & SC & -0.010  & - & -0.012 & - & -0.007  & - & - & - & - & - & 0.013 & - & 0.647 & -  \\
& NMI & 0.398  & - & 0.404 & - & 0.394 & - & - & - & - & - & 0.169 & - & \textbf{0.414} & -  \\
& RL & -  & - & - & - & - & - & - & - & - & - & 0.0014 & - & 0.0020 & -\\
& RRL & - & - & - & - & - & - & - & -  & - & - & 40 & - & 100 & - \\
& ETP & 5.13  & - & 5.15 & - & 5.22 & - & - & - & - & - & 5.47 & - & 5.2 & -\\
& CS & 916-20  & - & 845-18 & - & 650-41 & - & - & - & - & - & 478-76 & - & 540-1 & -\\
\midrule

\multirow{5}{*}{20NG} & SC & -0.021  & - & -0.186 & - & -0.103 & - & - & - & - & - & 0.066 & - & 0.199 & -  \\
& NMI & 0.155  & - & 0.167 & - & 0.176 & - & - & - & - & - & 0.018 & - & \textbf{0.229} & -  \\
& RL & -  & - & - & - & - & - & - & - & - & - & 0.0006 & - & 0.0012 & -\\
& RRL & - & - & - & - & - & - & - & -  & - & - & 25\txsps{\down} & - & 50 & - \\
& ETP & 4.03  & - & 3.64 & - & 3.77 & - & - & - & - & - & 4.31 & - & 3.87 & -\\
& CS & 2217-107  & - & 4024-52 & - & 4227-103 & - & - & - & - & - & 1142-582 & - & 3203-105 & -\\
\bottomrule
\end{tabular}
}
\end{center}
\end{table*}

\begin{table*}[!ht]
\caption{{\bf Metrics obtained by \dclam and baselines corresponding to the best ARI.} The best performance for each dataset is in {\bf boldface}. (note abbreviations DCEC$\to$DC, EDCWRN$\to$EDC, Entropy$\to$ETP, Cluster-size$\to$CS, No-AE$\to$NAE, Conv-AE$\to$CAE, EDCWRN-AE$\to$EAE, Resnet-AE$\to$RAE).`-' denotes NA. 
}
\label{tab:arigt}
\begin{center}
{\footnotesize

\begin{tabular}{lc|cc|cc|cc|cc|c|ccc}
\toprule
\multirow{2}{*}{\bf Data}  &\multirow{2}{*}{\bf Met} &\multicolumn{2}{c}{\bf Kmeans} &\multicolumn{2}{c}{\bf Agglo} &\multicolumn{2}{c}{\bf \clam} &\multicolumn{2}{c}{\bf DC} 
&\multirow{2}{*}{\bf EDC} 
& \multicolumn{3}{c}{\bf \dclam}\\ 
\cmidrule{3-10} \cmidrule{12-14}
& & \bf NAE & \bf CAE & \bf NAE & \bf CAE & \bf NAE & \bf CAE & \bf CAE & \bf RAE & & \bf CAE & \bf EAE & \bf RAE\\

\midrule
FM & ARI & 0.347 & 0.491 & 0.350 & 0.448 & 0.352 & 0.472  & 0.416 & 0.470 & 0.295 & 0.475 & 0.384 & \textbf{0.486} \\
\midrule
C-10 & ARI & 0.041 & 0.061 & 0.034 & 0.049 & 0.047 & 0.045  & 0.071 & 0.070 & 0.031 & 0.074 & 0.058 & \textbf{0.075} \\
\midrule
C-100 & ARI & 0.021 & 0.025 & 0.018 & 0.020 & 0.023 & 0.024  & 0.026 & 0.026 & 0.012 & \textbf{0.028} & 0.025 & 0.026 \\
\midrule
USPS & ARI & 0.476 & 0.541 & 0.508 & 0.587 & 0.473 & 0.527  & 0.654 & 0.673 & 0.637 & 0.668 & 0.684 & \textbf{0.689} \\
\midrule
STL & ARI & 0.060 & 0.077 & 0.043 & 0.048 & 0.062 & 0.082  & 0.084 & 0.085 & 0.046 & \textbf{0.089} & 0.087 & 0.088 \\
\midrule
CBird & ARI & 0.008 & 0.008 & 0.012 & 0.009 & 0.008 & 0.011  & 0.008 & 0.009 & 0.005 & 0.010 & \textbf{0.012} & 0.011 \\
\bottomrule
\end{tabular}
}
\end{center}
\end{table*}
\subsection{Detailed results with various clustering quality metrics} \label{asec:emp:all-metrics}
Table \ref{tab:scgt} provides a comprehensive overview of the metrics (SC, NMI, RL, ETP, and CS) for \dclam, and corresponding baselines, focusing on the best SC in each method across various AE architectures where RL is constrained to 10\% of the pretrained AE loss. 
For $k$-means, Agglomerative and \clam, we apply them both in the original ambient space (No-AE or NAE) and in the latent space (utilizing CAE).
RL is not presented for $k$-means, Agglomerative and \clam for the original space and for CAE as it remains consistent across the three methods after pretraining. Similarly, Table \ref{tab:rlgt} provides a similar overview of the metrics (SC, NMI, RL, ETP, and CS) for \dclam, and corresponding baselines, focusing on the best Relative RL (RRL) in each method across various AE architectures where SC is constrained to 10\% of the best/peak SC of the method.
Table \ref{tab:nmigt}
represents all corresponding metrics focusing on the best NMI in each method. These tables highlight that \dclam exhibits strong performance not only in terms of SC and RL, but also when compared to the ground truth labels via NMI.
In fact, for NMI, \dclam has the best values in 5 out of the 8 datasets (DCEC has the best values on the other 3).
Additionally, \dclam clusters maintain reasonable entropy (ETP) and cluster size (CS), ensuring a balanced clustering outcome. 

For an understanding of the importance of ETP and CS in clustering, consider the case of Agglomerative clustering in the latent space (CAE) on the CIFAR-10 dataset (see Table \ref{tab:scgt}). In this instance, almost all points (49991 out of 50000) belong to one cluster, while the other 9 clusters contain only one data point each, indicating very poor clustering. The low entropy (0.003) further highlights the deficiency of the clustering.

In certain situations, when comparing two clustering methods, it can happen that a method performs better in terms of SC and RL but still exhibits a lower NMI compared to another method (see Table \ref{tab:scgt} for USPS where \dclam outperforms DCEC in both CAE and RAE architecture in both SC and RL, however, the NMI is worse than DCEC in both cases). This indicates that the alignment of semantic class (ground truth or true underlying structure) with the geometric characteristics of the data might not be consistent or straightforward.

\subsection{How interpretable are the memories of \dclam?} \label{asec:emp:mem-viz}
\begin{figure}[!ht]
\begin{center}
\begin{subfigure}{0.49\columnwidth}
\centering
\begin{subfigure}{0.12\columnwidth}
\centering
    \includegraphics[width=0.6\textwidth]{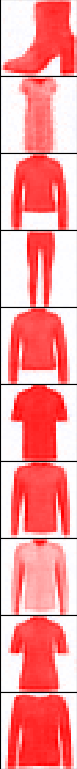}
   \captionsetup{labelformat=empty} 
   \caption{$n_0$}
  \label{fig:dcpp:fm:e0}
\end{subfigure}
~
\begin{subfigure}{0.12\columnwidth}
  \includegraphics[width=0.6\textwidth]{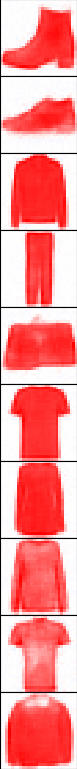}
   \captionsetup{labelformat=empty} 
   \caption{$n_5$}
  \label{fig:dcpp:fm:e5}
\end{subfigure}
~
\begin{subfigure}{0.12\columnwidth}
  \includegraphics[width=0.6\textwidth]{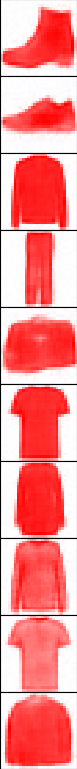}
   \captionsetup{labelformat=empty} 
   \caption{$n_{10}$}
  \label{fig:dcpp:fm:e10}
\end{subfigure}
~
\begin{subfigure}{0.12\columnwidth}
  \includegraphics[width=0.6\textwidth]{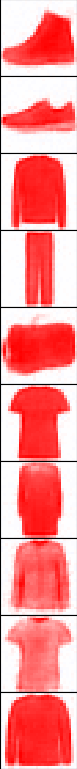}
   \captionsetup{labelformat=empty} 
   \caption{$n_{20}$}
  \label{fig:dcpp:fm:e20}
\end{subfigure}
~
\begin{subfigure}{0.12\columnwidth}
  \includegraphics[width=0.6\textwidth]{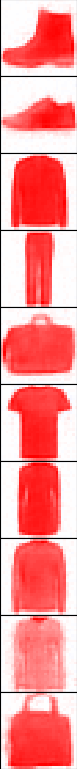}
   \captionsetup{labelformat=empty} 
   \caption{$n_{50}$}
  \label{fig:dcpp:fm:e50}
\end{subfigure}
~
\begin{subfigure}{0.12\columnwidth}
\centering
  \includegraphics[width=0.6\textwidth]{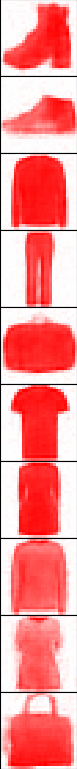}
   \captionsetup{labelformat=empty} 
   \caption{$n_{100}$}
  \label{fig:dcpp:fm:e100}
\end{subfigure}
\addtocounter{subfigure}{-6} 
\caption{FM}
\label{fig:memory:fm-dclam}
\end{subfigure}
\begin{subfigure}{0.49\columnwidth}
\centering
\begin{subfigure}{0.12\columnwidth}
\centering
    \includegraphics[width=0.6\textwidth]{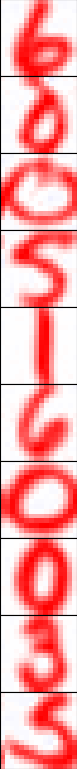}
   \captionsetup{labelformat=empty} 
   \caption{$n_0$}
  \label{fig:dcpp:usps:e0}
\end{subfigure}
~
\begin{subfigure}{0.12\columnwidth}
  \includegraphics[width=0.6\textwidth]{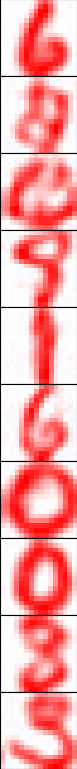}
   \captionsetup{labelformat=empty} 
   \caption{$n_5$}
  \label{fig:dcpp:usps:e5}
\end{subfigure}
~
\begin{subfigure}{0.12\columnwidth}
  \includegraphics[width=0.6\textwidth]{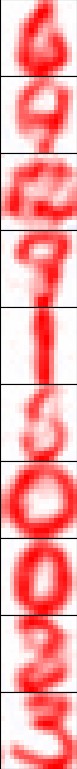}
   \captionsetup{labelformat=empty} 
   \caption{$n_{10}$}
  \label{fig:dcpp:usps:e10}
\end{subfigure}
~
\begin{subfigure}{0.12\columnwidth}
  \includegraphics[width=0.6\textwidth]{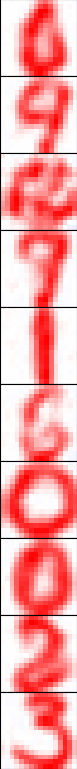}
   \captionsetup{labelformat=empty} 
   \caption{$n_{20}$}
  \label{fig:dcpp:usps:e20}
\end{subfigure}
~
\begin{subfigure}{0.12\columnwidth}
  \includegraphics[width=0.6\textwidth]{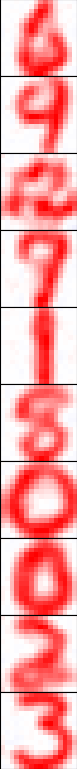}
   \captionsetup{labelformat=empty} 
   \caption{$n_{50}$}
  \label{fig:dcpp:usps:e50}
\end{subfigure}
~
\begin{subfigure}{0.12\columnwidth}
\centering
  \includegraphics[width=0.6\textwidth]{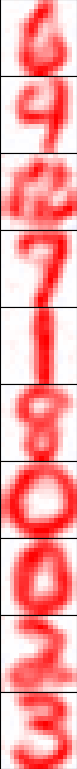}
   \captionsetup{labelformat=empty} 
   \caption{$n_{100}$}
  \label{fig:dcpp:usps:e100}
\end{subfigure}
\addtocounter{subfigure}{-6} 
\caption{USPS}
\label{fig:memory:usps-dclam}
\end{subfigure}
\end{center}
\caption{{\bf Evolution of prototypes for Fashion-MNIST and USPS for \dclam.} We visualize the prototypes at the $n^{\text{th}}$ training epoch for $n = 0, 5, 10, 20, 50, 100$ (with $T=10$).}
\label{fig:memory-all}
\end{figure}

We explore the representation of the learned prototypes in latent space for \dclam for Fashion-MNIST and USPS in \cref{fig:memory-all}. For Fashion-MNIST, the 60k images are partitioned into 10 clusters, and the evolution of cluster prototypes is visualized in \cref{fig:memory:fm-dclam} during the training process outlined in Algorithm~\ref{alg:dcam-com} for \dclam. In each sub-figure we observe the evolution over epochs. At epoch 0, there are no distinct prototypes for clustering; instead, there are pairs of pullover (rows 3 and 5), shirts (rows 7 and 8), and t-shirts/tops (rows 6 and 9). However, discernible patterns emerge at epoch 10, refining further by epoch 20. By epoch 100, all ten prototypes represent distinct shapes, representing different cluster centroids.

\begin{figure}[!ht]
\begin{center}
\begin{subfigure}{0.046\columnwidth}
\centering
    \includegraphics[width=1\textwidth]{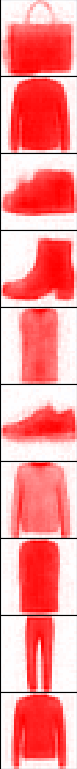}
   \caption{{M}}
  \label{fig:dclam:fm20}
\end{subfigure}
~
\begin{subfigure}{0.92\columnwidth}
\centering
  \includegraphics[width=1\textwidth]{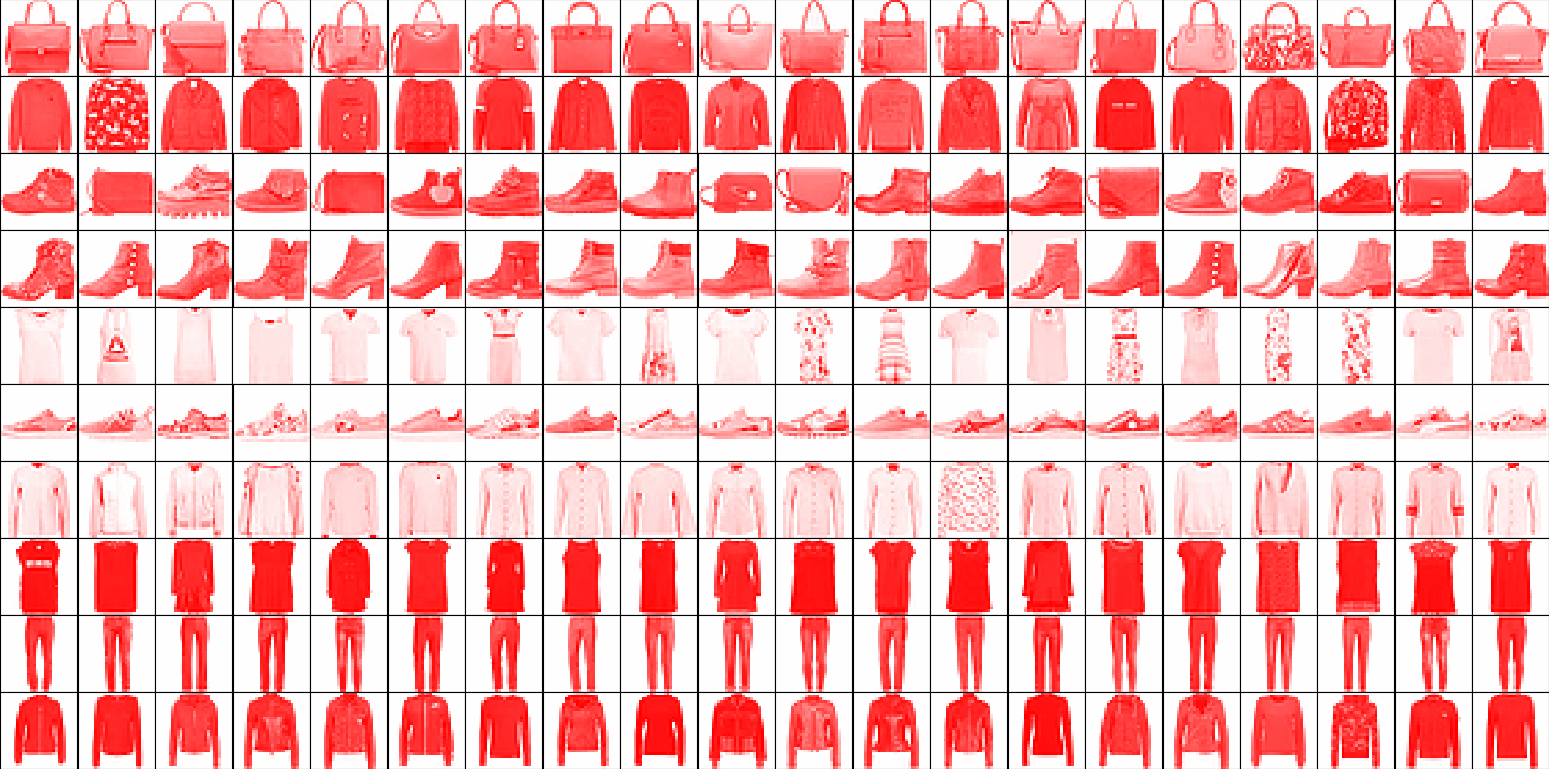}
   \caption{20-closest points}
  \label{fig:dclam:cbird20}
\end{subfigure}
\end{center}
\caption{\dclam: {\bf Final memories (left column) and the 20-closest points for each memory in F-MNIST.}}
\label{fig:dcpp-closest}
\end{figure}

Figure~\ref{fig:dcpp-closest} provides a visualization of the 20 data points from the Fashion-MNIST dataset that are closest to each learned prototype. In the figure, the leftmost column represents the 10 learned prototypes stacked vertically, with each corresponding to one cluster in the dataset. The subsequent columns show the 20 data points that are nearest to each prototype or cluster center in terms of the Euclidean distance within the learned latent space.
This visualization offers valuable insights into the nature of the learned prototypes and the clusters they represent. By examining the closest points, we can see how well the prototypes capture the underlying data distribution and whether they correspond to semantically meaningful clusters. For example, the learned prototypes often represent exemplar or central examples of specific fashion items (such as shirts, trousers, or shoes), while the nearest points reveal variations of these items that still fall within the same cluster.
Such visualizations not only highlight the interpretability of \dclam’s energy-based clustering but also demonstrate the model's ability to learn compact and meaningful representations of the data. This approach provides an intuitive way to assess the quality of the learned clusters and their alignment with the dataset's inherent structure.

\subsection{Effect of latent dimensionality } 
It is important to clarify that in \dclam the latent dimension $m$ is always set as the true number of classes per dataset, i.e., $m=k$. Indeed, most deep clustering schemes in the literature such as DCEC~\citep{guo2017deep}, DEC~\citep{xie2016unsupervised}, DEKM~\citep{guo2021deep}, and EDCWRN~\cite{golzari2023edcwrn} either follow this strategy or fix this to a specific number (e.g., 10), since latent representations are not only just good representations of the data points, they also represent the clusters. We thus always set this latent dimension as the number of clusters. By setting this exactly the same as the number of clusters, each latent dimension should ideally represent one specific cluster. If the latent dimensions are larger than the number of clusters, some dimensions might not align with any specific cluster, or multiple dimensions could end up representing the same cluster. This can introduce redundancy and result in a less efficient representation. On the other hand, if the latent dimensions are smaller than the number of clusters, some clusters may not be adequately represented. This forces multiple clusters to share the same dimension, making it challenging for the model to distinguish between them accurately.

We can also consider a spectral argument~\citep{von2007tutorial} for setting $m=k$ .
Given a set of $n$ points (in any representation) from $k$ ground-truth clusters, consider a graph (weighted or
unweighted and undirected) with each point as a node, and edges between points belonging to the same cluster, and no
inter-cluster edges. This graph would have $k$ connected components, and the Laplacian $L \in \mathbb{R}^{n \times n}$ of
this graph will have $k$ zero eigenvalues (for example, see Von Luxburg [5, Proposition 2]). Now consider the first $k$
eigenvectors $u_1, \ldots, u_k \in \mathbb{R}^n$ forming the columns of the matrix $U \in \mathbb{R}^{n \times k}$. Then
each row $z_i \in \mathbb{R}^k$ of $U$ can serve as a representation of the point $i$, and the points will be
well-separated into $k$ clusters in this representation. This is the intuition that forms the basis of various spectral clustering algorithms.

The above implies the existence of a $k$-dimensional space where the points (coming from $k$ ground-truth
clusters) are well-separated into $k$ clusters. Thus, a latent space of $k$ dimensions is necessary to obtain
well-separated clusters. As Euclidean clustering becomes more challenging with increasing representation dimensionality
(the representation in which the clustering is happening), the motivation is to keep the latent space dimension as low
as possible as long as we have enough dimensions to separate the clusters. For this reason, the latent space
dimensionality in most deep clustering methods usually matches the desired number of clusters.

A higher latent dimensionality will definitely help with the
reconstruction but can potentially hurt Euclidean clustering; a lower latent dimensionality would not be sufficient to
obtain $k$ well-separated clusters.
Fortunately, given the extremely expressive modern deep learning encoder and decoders, we are able to still get quite low reconstruction loss with a $m=k$ dimensional latent space.
%
\begin{figure}[!ht]
\begin{center}
\begin{subfigure}{0.49\columnwidth}
\centering
    \includegraphics[width=1\textwidth]{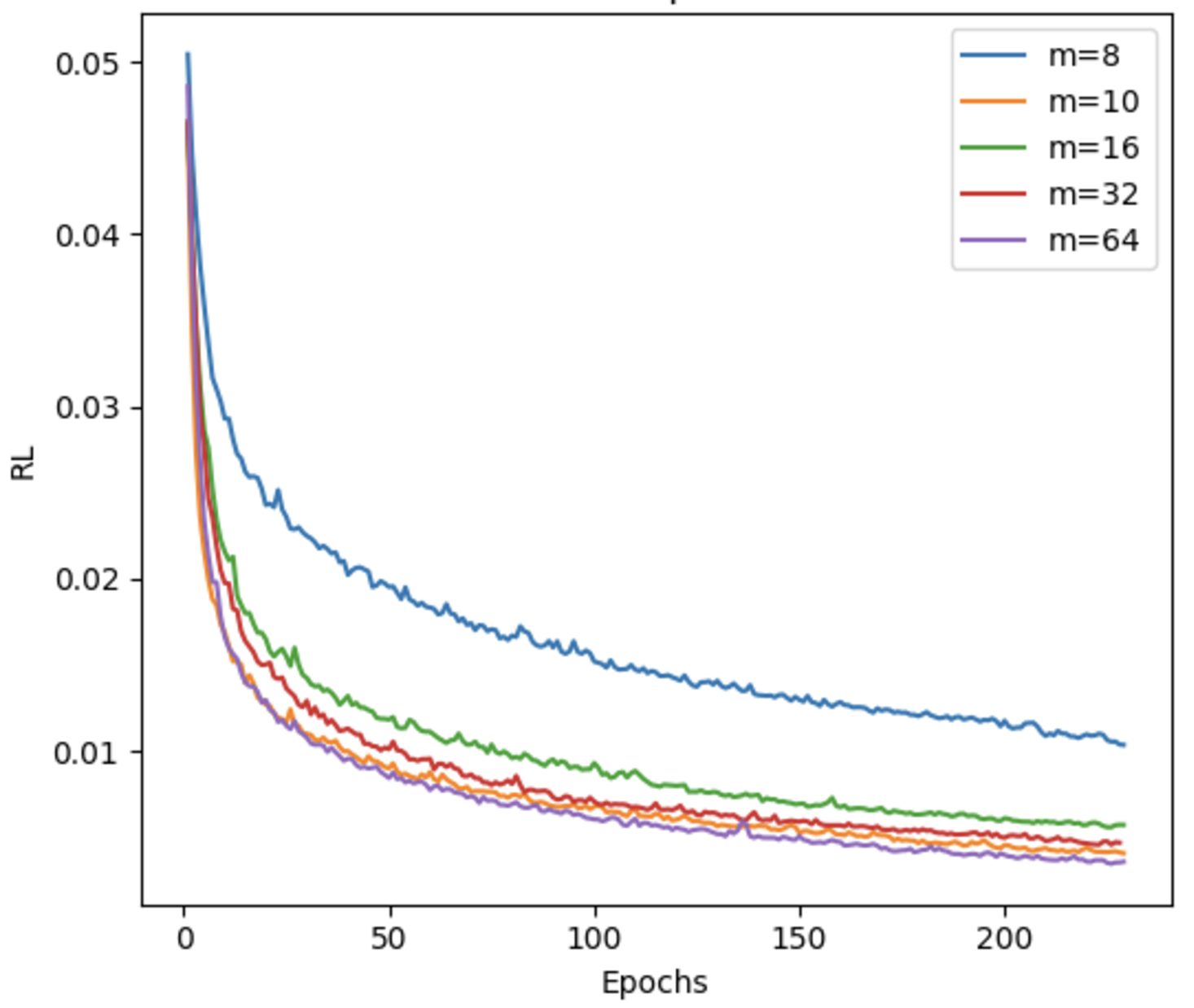}
  \caption{RL vs Training}
  \label{fig:ld:usps-ld-rl}
\end{subfigure}
~
\begin{subfigure}{0.49\columnwidth}
\centering
  \includegraphics[width=1\textwidth]{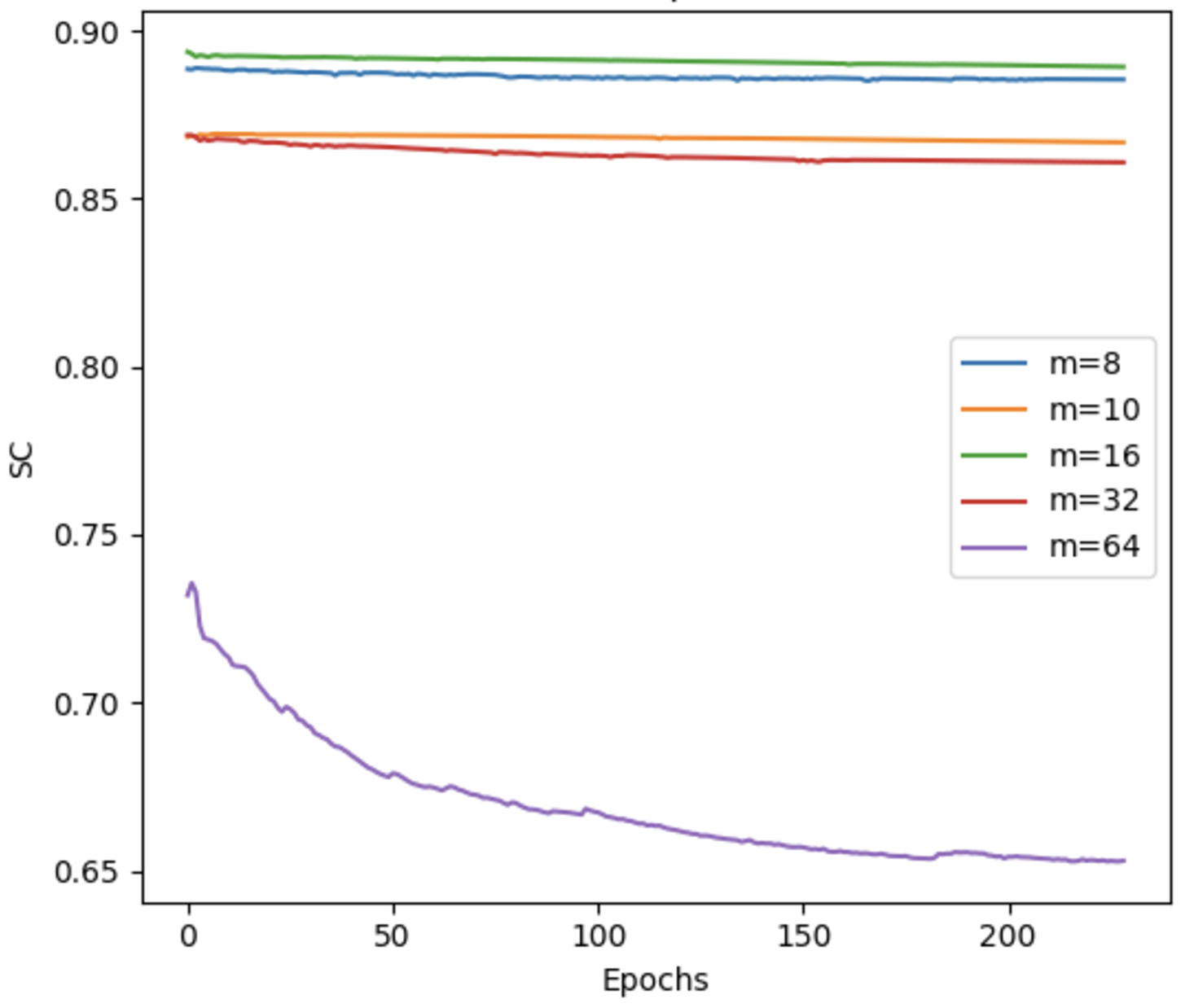}
  \caption{SC vs Training}
  \label{fig:ld:usps-ld-sc}
\end{subfigure}
\end{center}
\caption{\color{black} Reconstruction loss (RL) and clustering quality (SC) for varying latent dimension ($m$) for USPS ($T$ is set to 10). } 
\label{fig:usps-ld-plots}
\end{figure}

Despite the aforementioned caveats, Figure~\ref{fig:usps-ld-plots} illustrates the impact of varying the latent space dimensionality $m$ on the USPS dataset, which consists of 10 classes. The figure highlights the trade-offs between reconstruction loss (RL) and Silhouette Coefficient (SC) for different values of $m$. 
From the results, it is evident that setting $m=10$ provides an optimal balance between RL and SC, achieving a strong trade-off compared to other values of $m$. This observation further supports our claim for matching the latent dimensionality with the number of clusters in the dataset. By doing so, \dclam effectively captures the underlying structure of the data while maintaining compact and well-separated clusters.
This experiment underscores the importance of selecting an appropriate latent dimensionality in clustering tasks and demonstrates how \dclam leverages this alignment to deliver meaningful and interpretable partitions.
\subsection{Additional details on hyperparameter selection} 
\begin{figure}[!ht]%
\begin{center}
\begin{subfigure}{0.32\columnwidth}
\centering
    \includegraphics[width=0.85\textwidth]{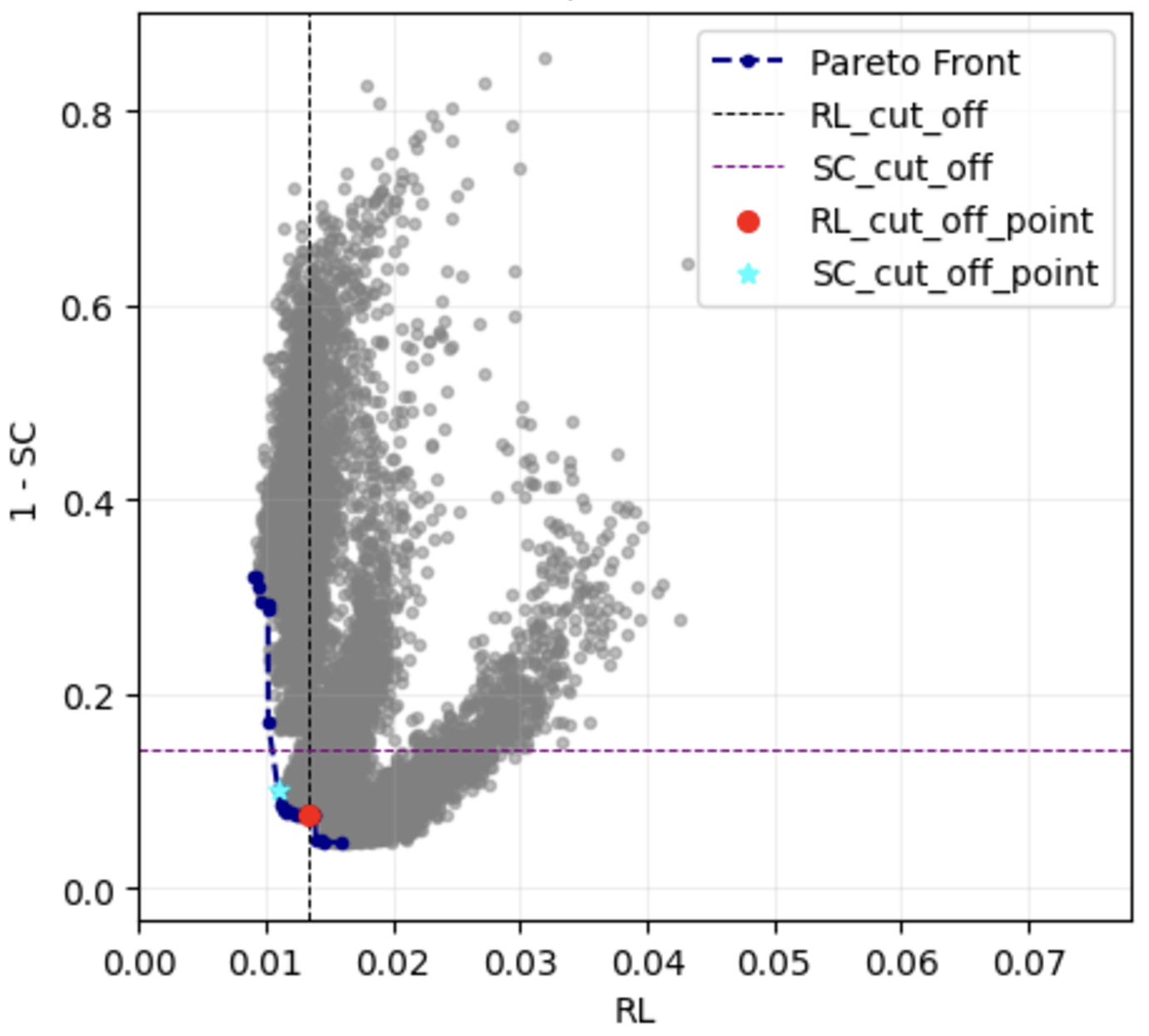}
  \caption{CAE: DCEC}
  \label{fig:hp:fm-cae-dcec}
\end{subfigure}
~
\begin{subfigure}{0.32\columnwidth}
\centering
  \includegraphics[width=0.85\textwidth]{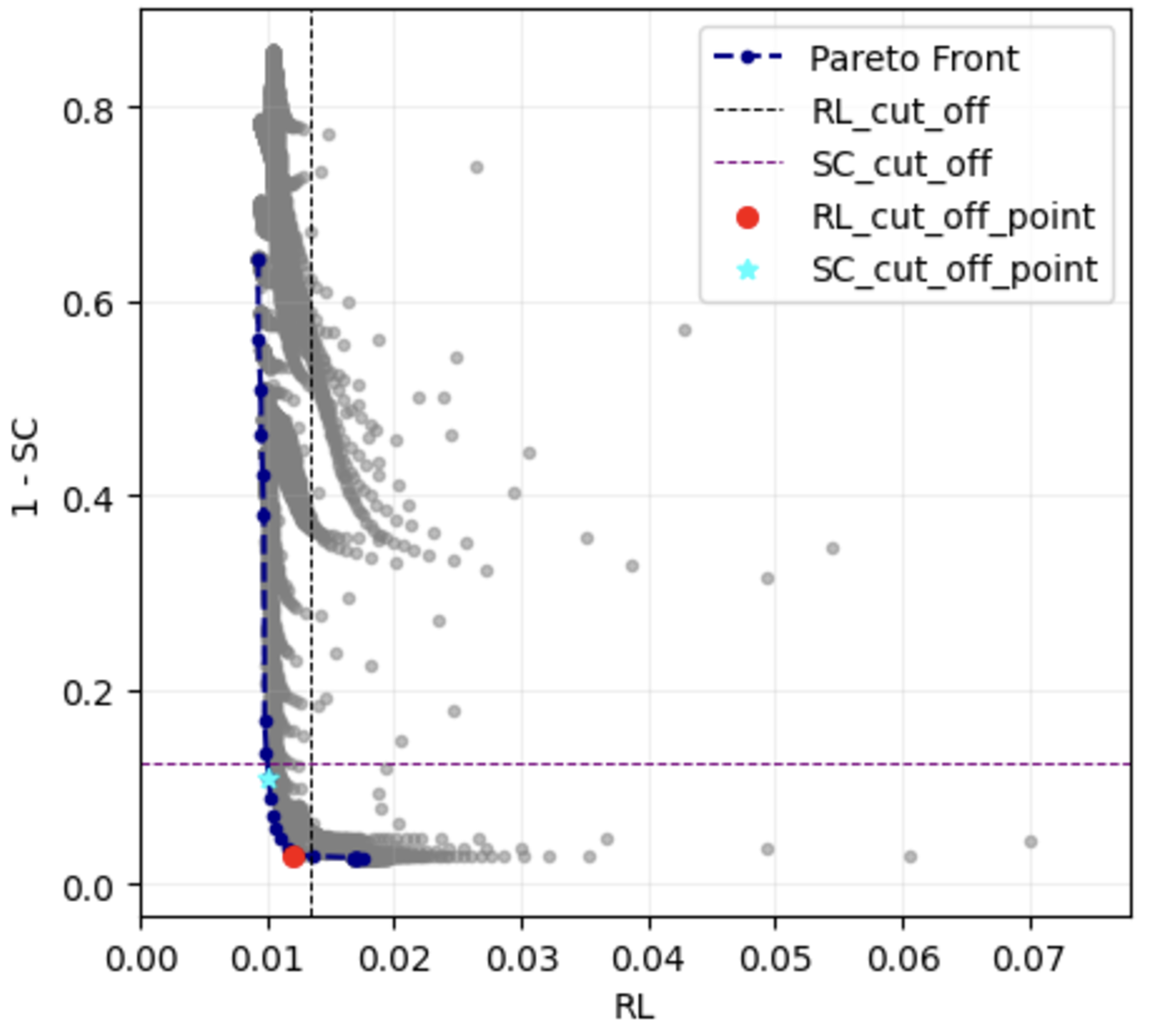}
  \caption{CAE: \dclam}
  \label{fig:hp:fm-cae-dclam}
\end{subfigure}
~
\begin{subfigure}{0.32\columnwidth}
\centering
    \includegraphics[width=0.85\textwidth]{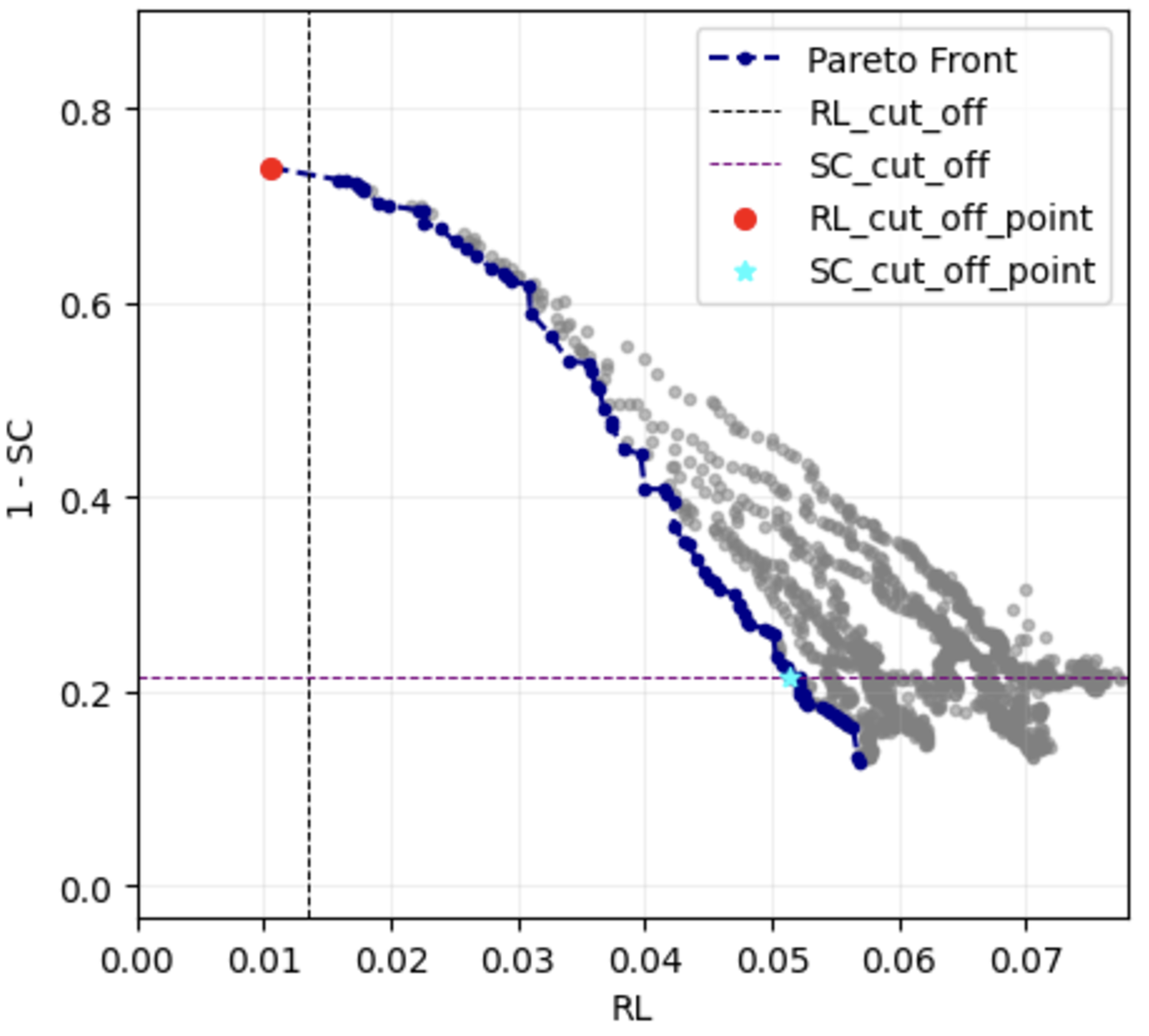}
  \caption{CAE: DEKM}
  \label{fig:hp:fm-cae-dekm}
\end{subfigure}
~
\begin{subfigure}{0.32\columnwidth}
\centering
  \includegraphics[width=0.85\textwidth]{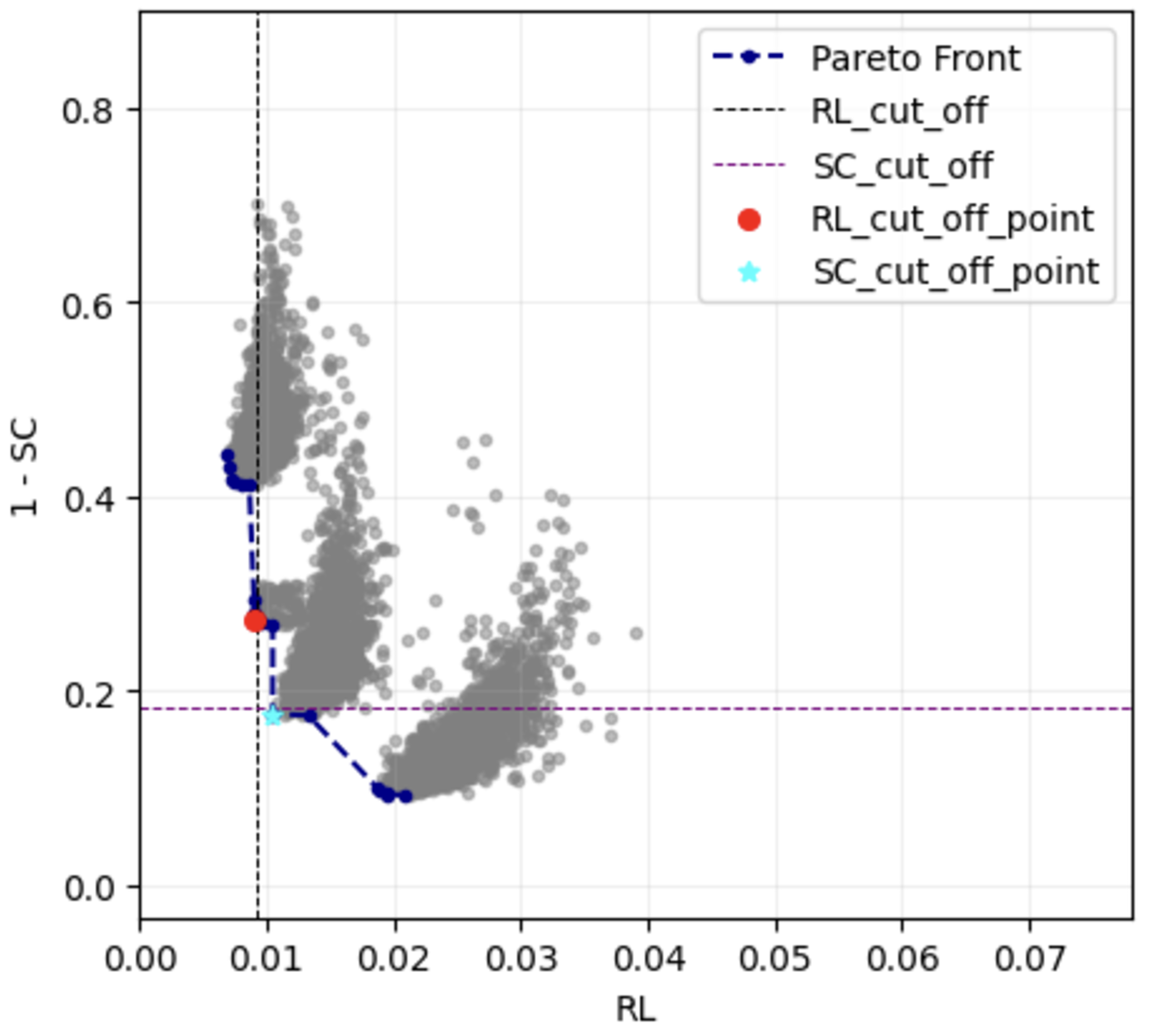}
  \caption{RAE: DCEC}
  \label{fig:hp:fm-rae-dcec}
\end{subfigure}
~
\begin{subfigure}{0.32\columnwidth}
\centering
    \includegraphics[width=0.85\textwidth]{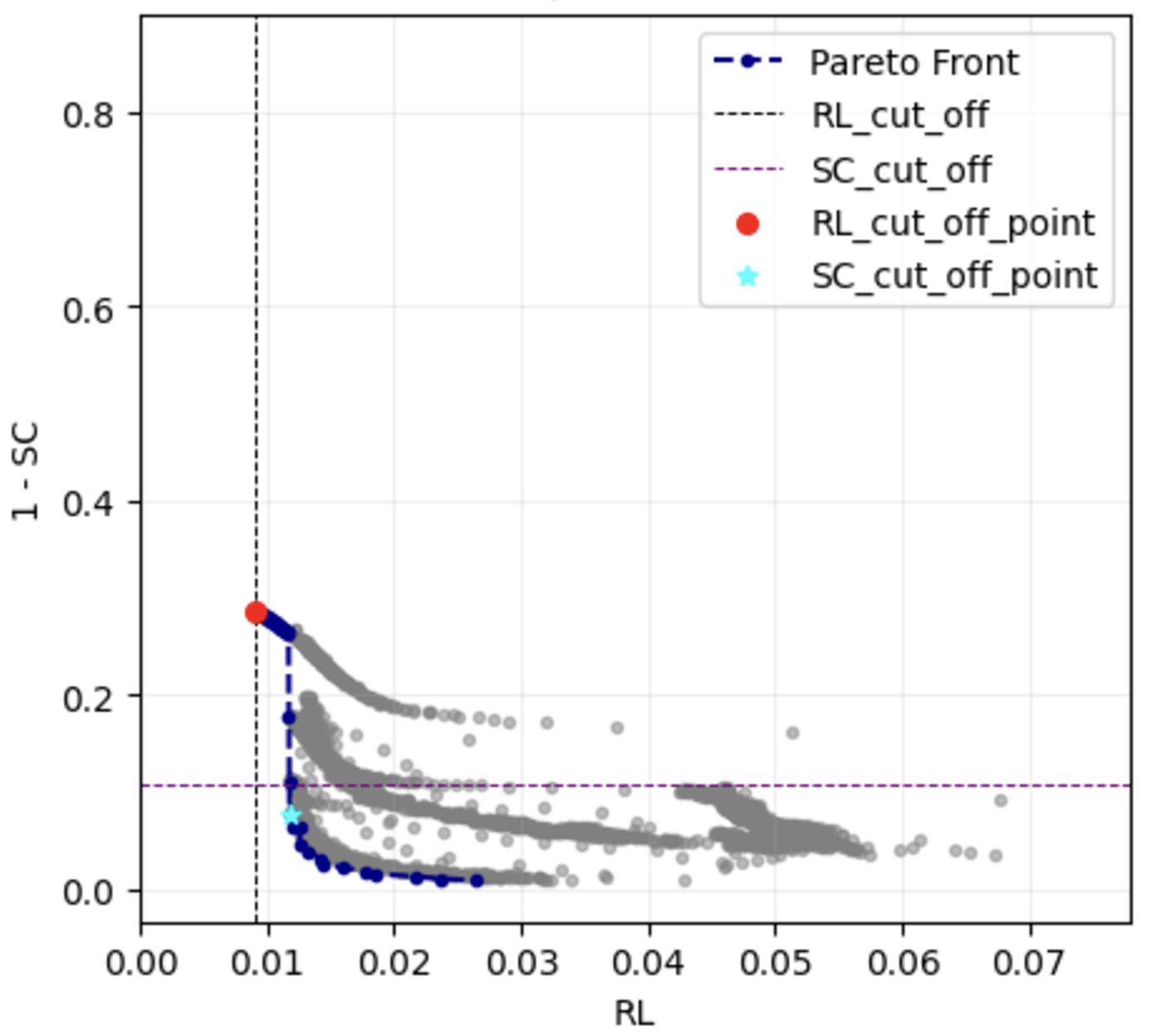}
  \caption{RAE: \dclam}
  \label{fig:hp:fm-rae-dclam}
\end{subfigure}
~
\begin{subfigure}{0.32\columnwidth}
\centering
  \includegraphics[width=0.85\textwidth]{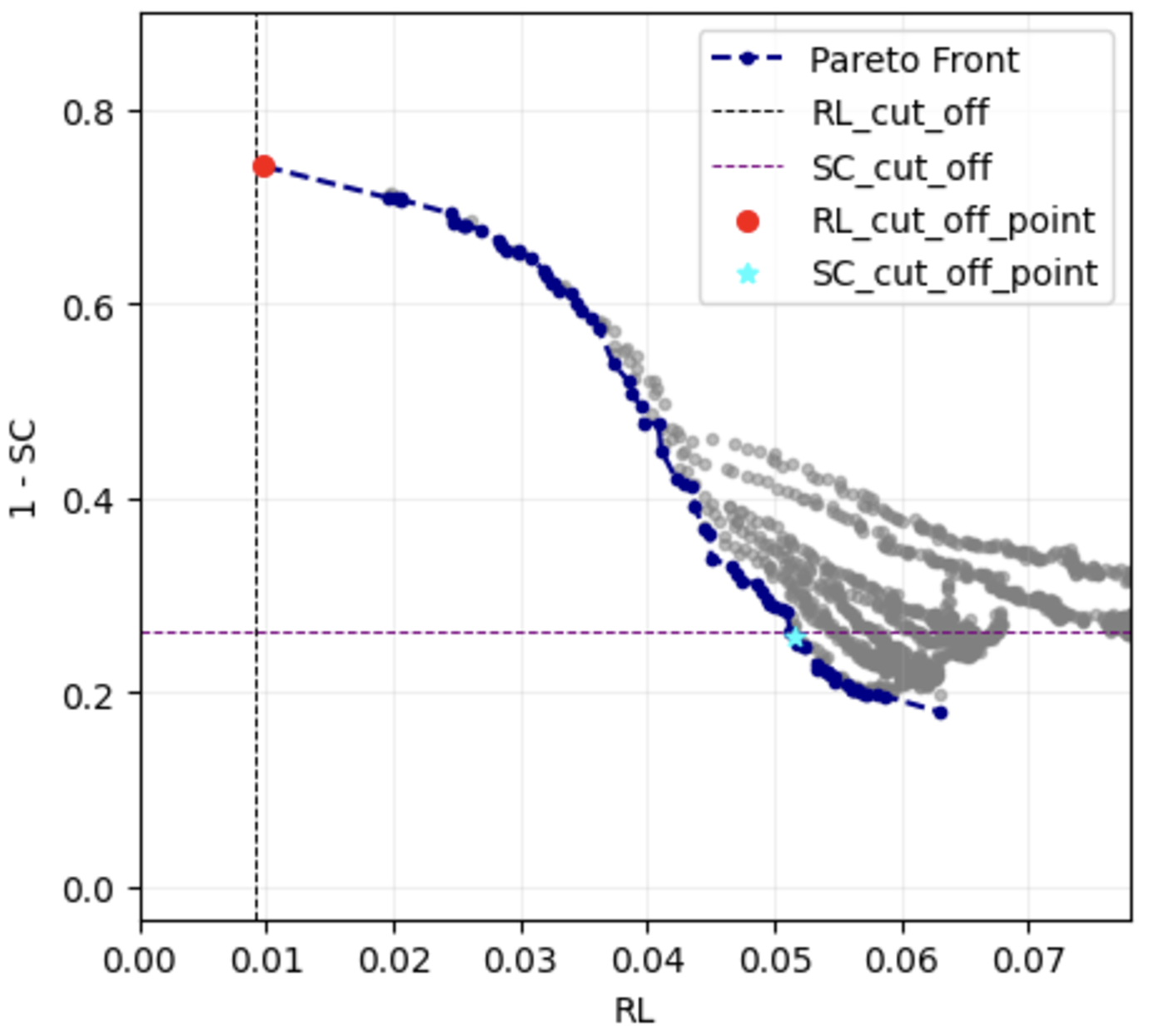}
  \caption{RAE: DEKM}
  \label{fig:hp:fm-rae-dekm}
\end{subfigure}
\end{center}
\caption{\color{black}\textbf{FMNIST:} Reconstruction loss and clustering quality (1-SC) for all hyperparameter configurations for DCEC, \dclam and DEKM with CAE and RAE architectures. {\em Lower is better for both axes}, since we plot 1-SC on the $y$-axis.
} 
\label{fig:fm-hp-plots}
\end{figure}

\begin{figure}[!ht]
\begin{center}
\begin{subfigure}{0.32\columnwidth}
\centering
    \includegraphics[width=0.85\textwidth]{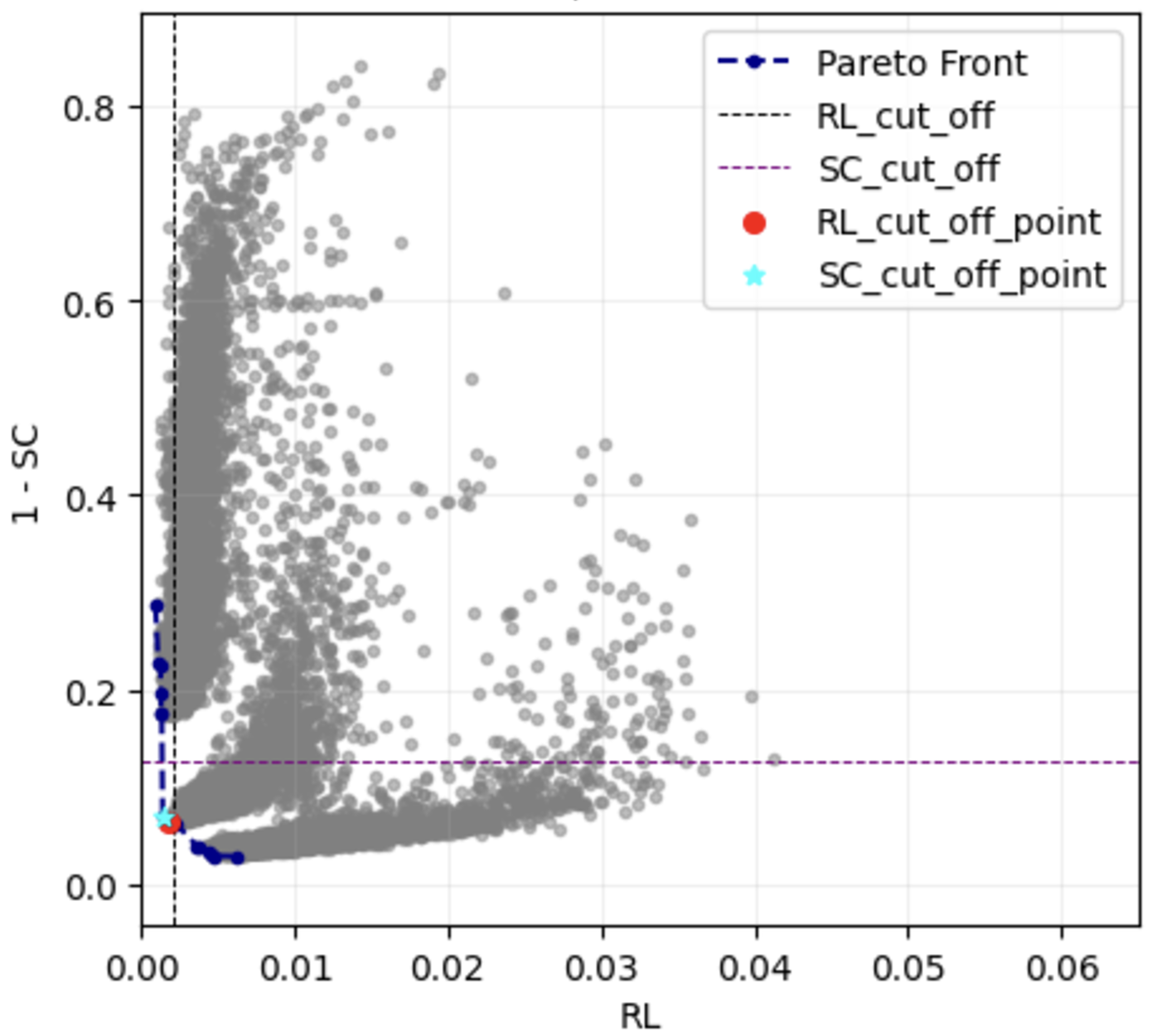}
  \caption{CAE: DCEC}
  \label{fig:hp:usps-cae-dcec}
\end{subfigure}
~
\begin{subfigure}{0.32\columnwidth}
\centering
  \includegraphics[width=0.85\textwidth]{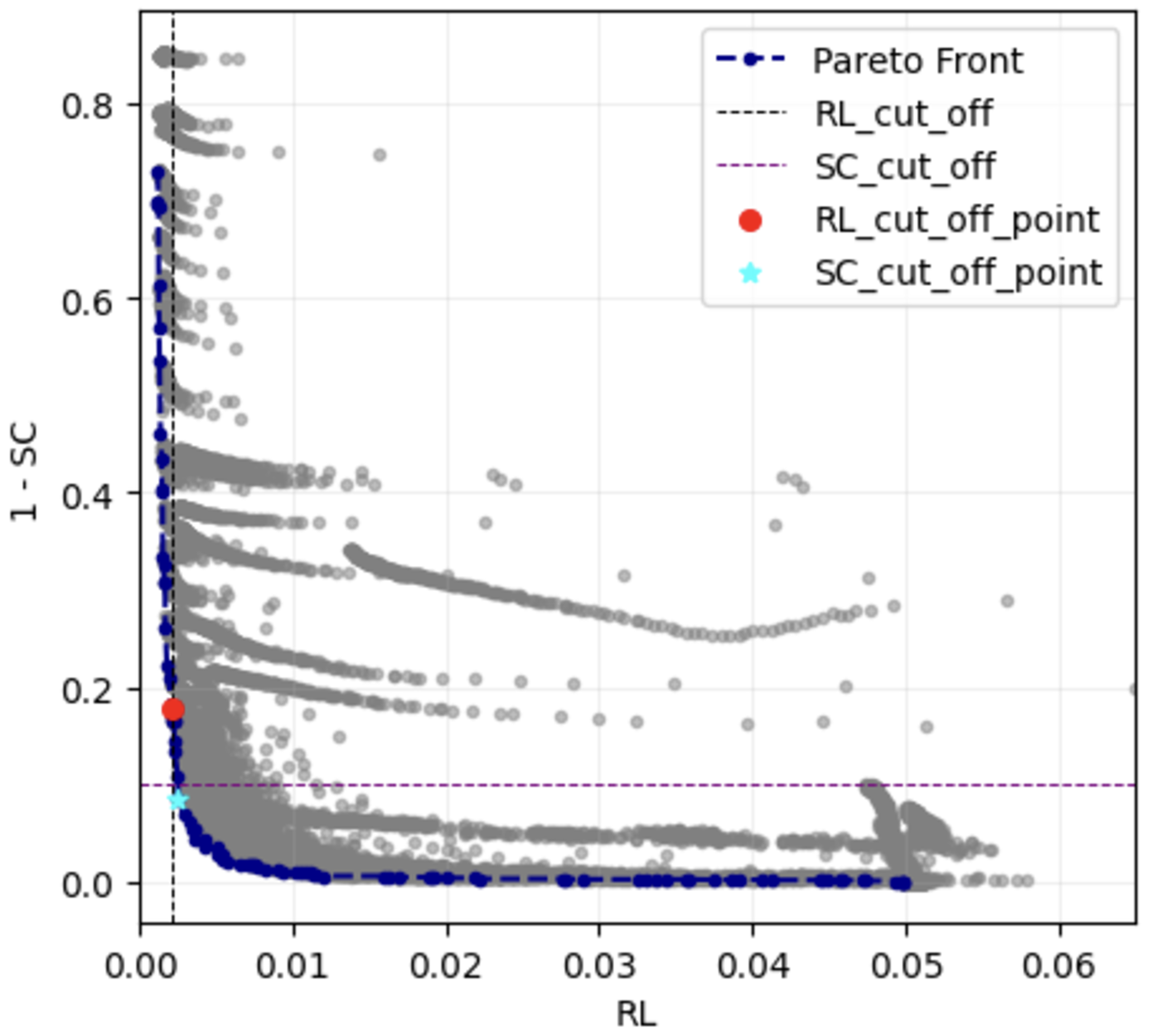}
  \caption{CAE: \dclam}
  \label{fig:hp:usps-cae-dclam}
\end{subfigure}
~
\begin{subfigure}{0.32\columnwidth}
\centering
    \includegraphics[width=0.85\textwidth]{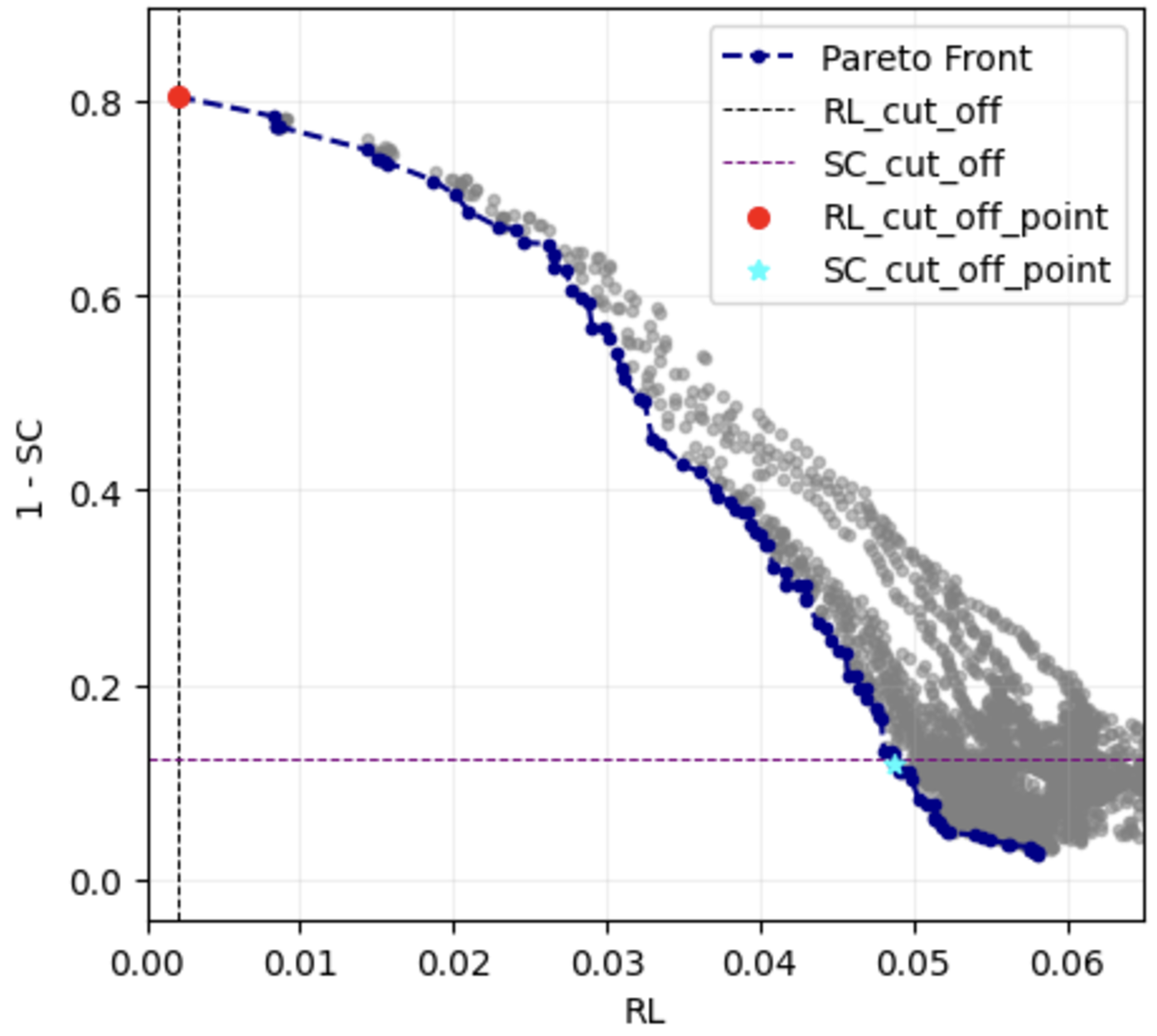}
  \caption{CAE: DEKM}
  \label{fig:hp:usps-cae-dekm}
\end{subfigure}
~
\begin{subfigure}{0.32\columnwidth}
\centering
  \includegraphics[width=0.85\textwidth]{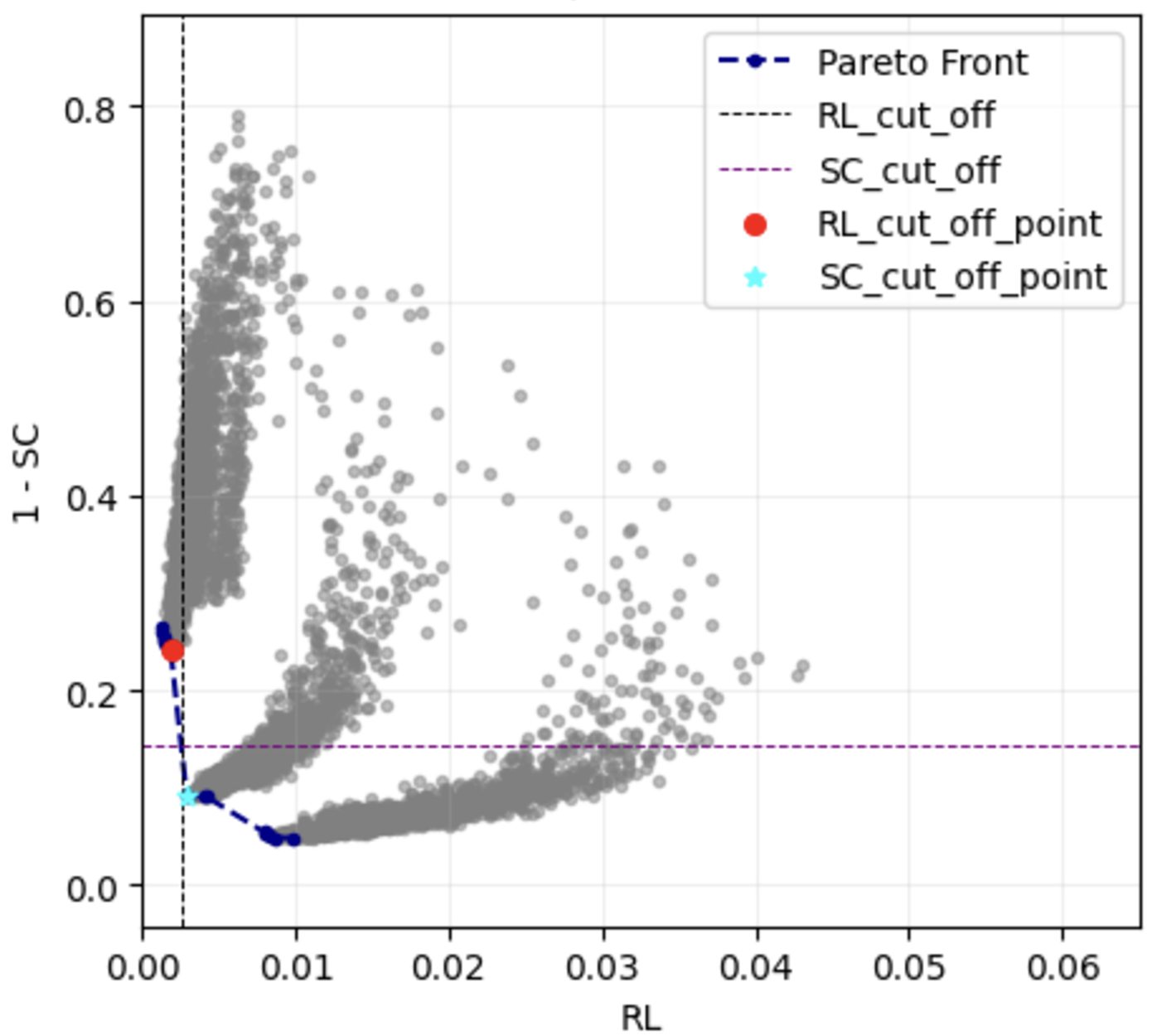}
  \caption{RAE: DCEC}
  \label{fig:hp:usps-rae-dcec}
\end{subfigure}
~
\begin{subfigure}{0.32\columnwidth}
\centering
    \includegraphics[width=0.85\textwidth]{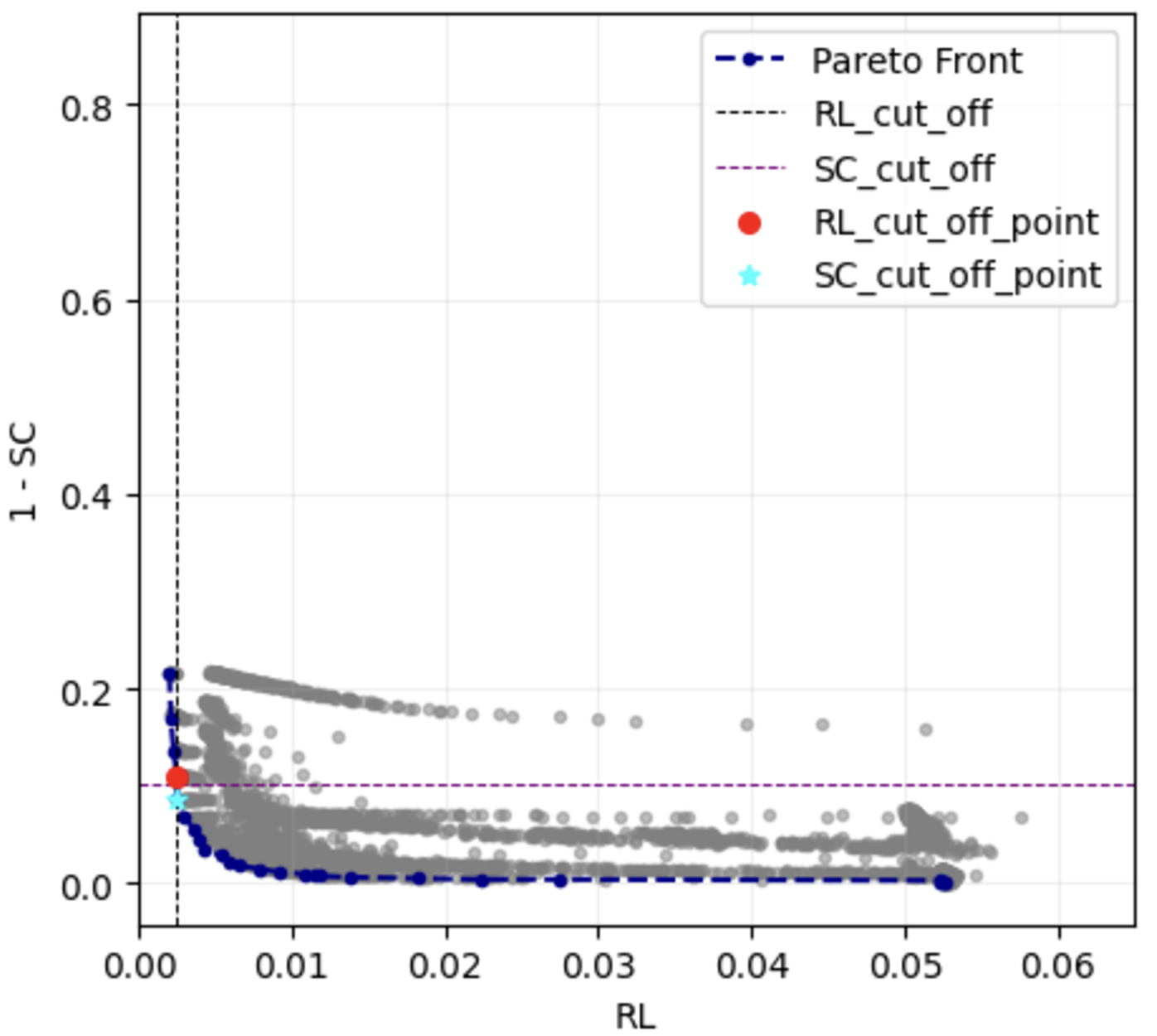}
  \caption{RAE: \dclam}
  \label{fig:hp:usps-rae-dclam}
\end{subfigure}
~
\begin{subfigure}{0.32\columnwidth}
\centering
  \includegraphics[width=0.85\textwidth]{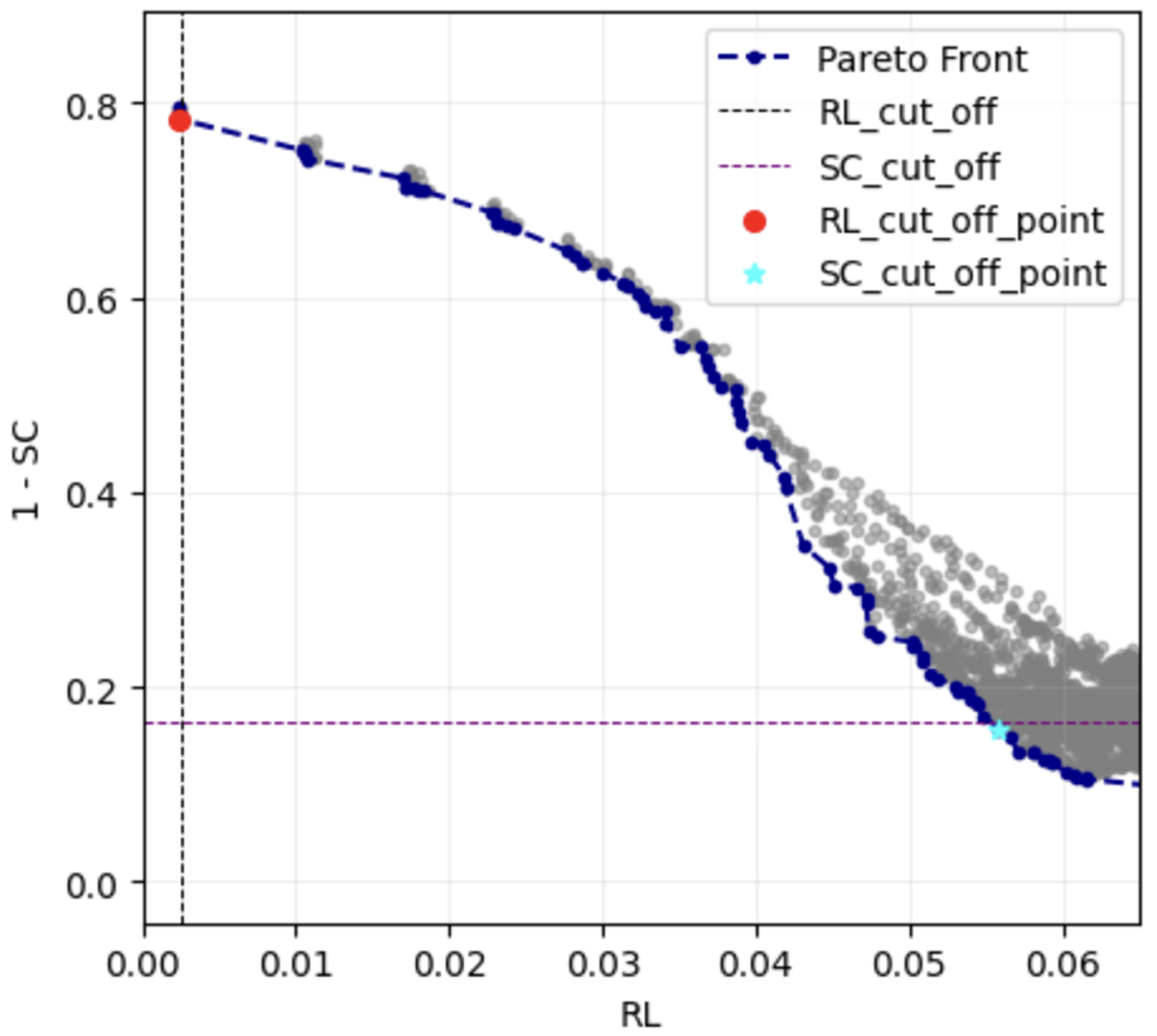}
  \caption{RAE: DEKM}
  \label{fig:hp:usps-rae-dekm}
\end{subfigure}
\end{center}
\caption{\color{black}\textbf{USPS:} Reconstruction loss and clustering quality (1-SC) for all hyperparameter configurations for DCEC, \dclam and DEKM with CAE and RAE architectures. {\em Lower is better for both axes}.
} 
\label{fig:usps-hp-plots}
\end{figure}

\begin{figure}[!ht]
\begin{center}
\begin{subfigure}{0.32\columnwidth}
\centering
    \includegraphics[width=0.85\textwidth]{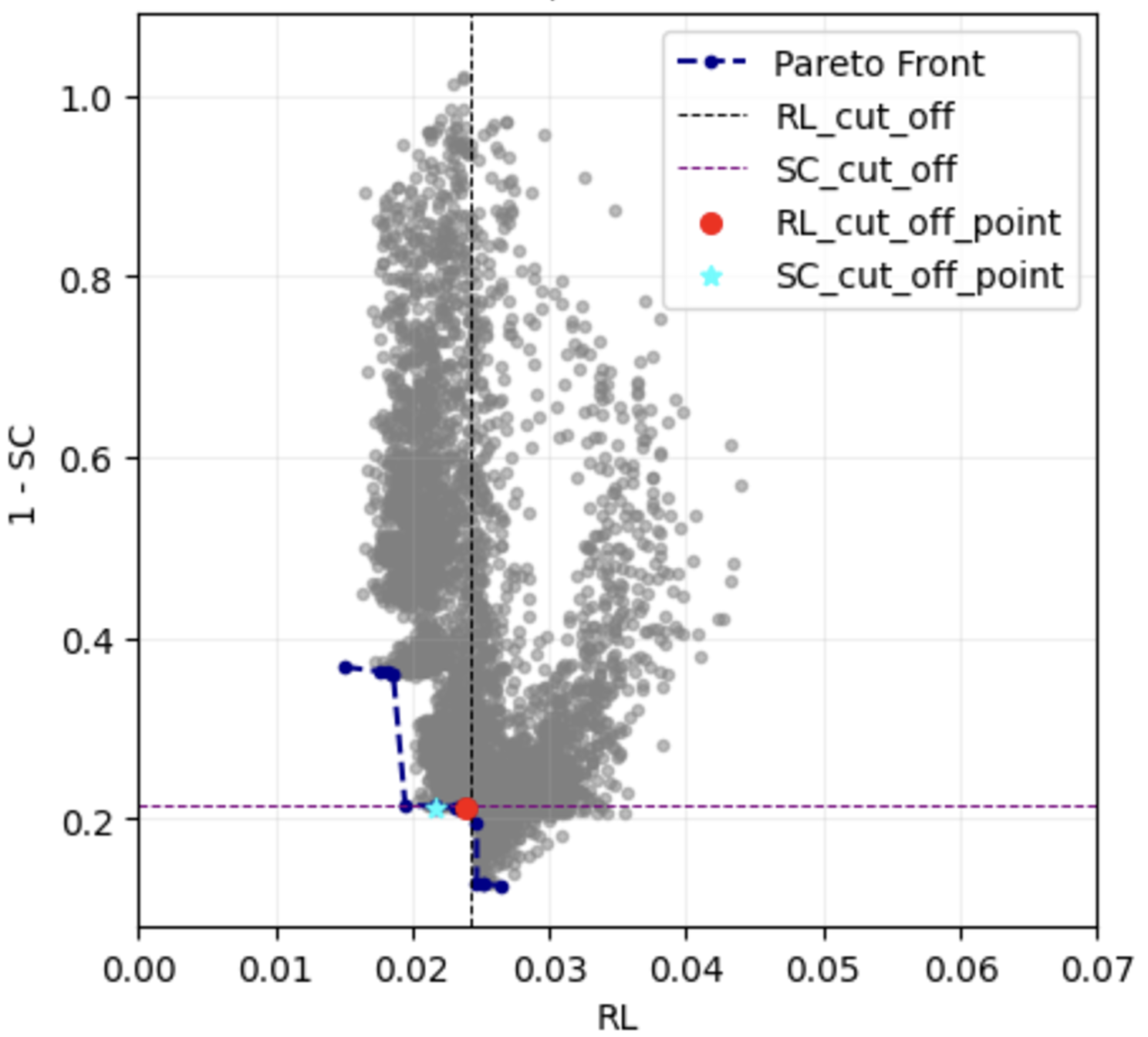}
  \caption{CAE: DCEC}
  \label{fig:hp:c10-cae-dcec}
\end{subfigure}
~
\begin{subfigure}{0.32\columnwidth}
\centering
  \includegraphics[width=0.85\textwidth]{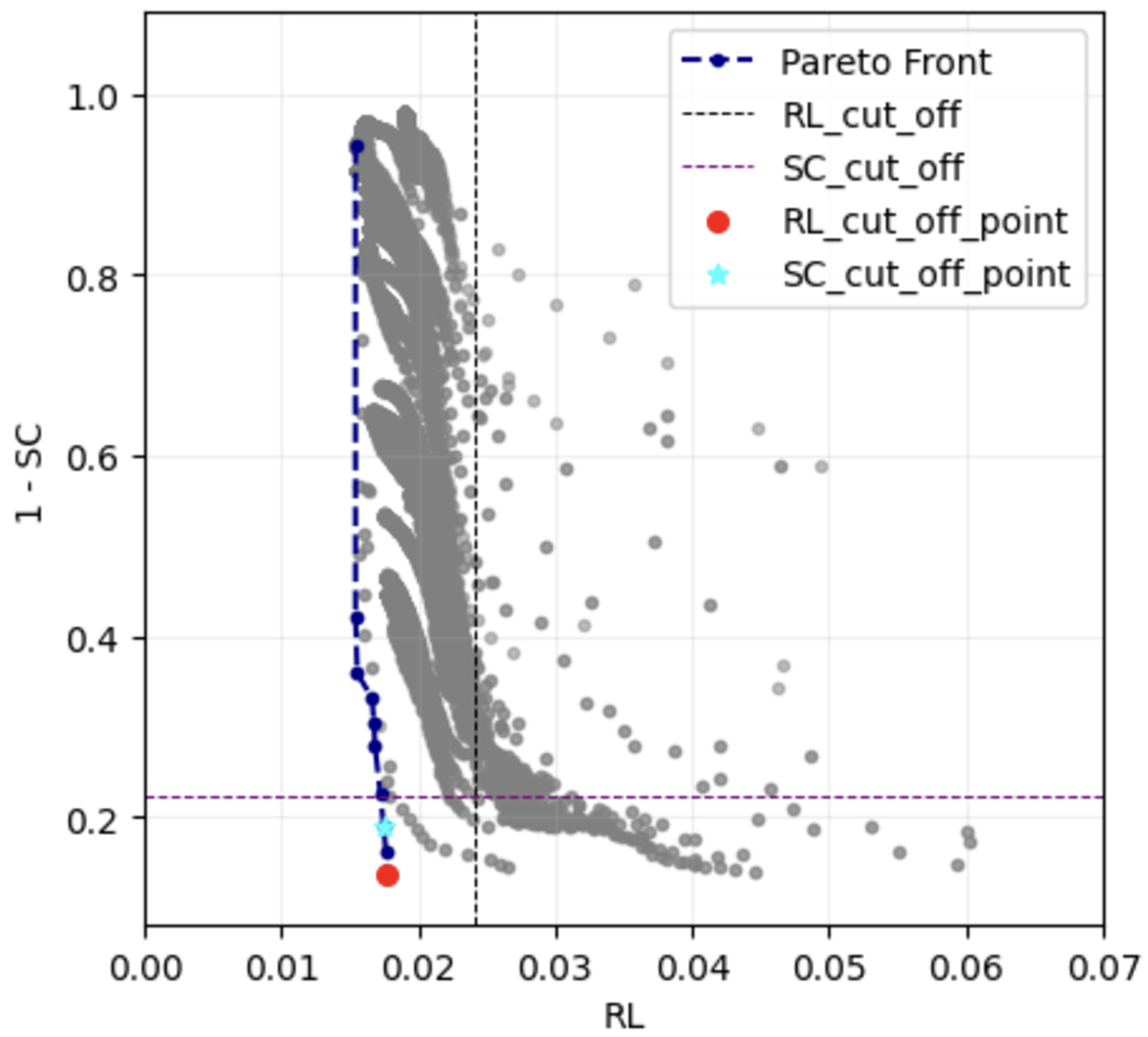}
  \caption{CAE: \dclam}
  \label{fig:hp:c10-cae-dclam}
\end{subfigure}
~
\begin{subfigure}{0.32\columnwidth}
\centering
    \includegraphics[width=0.85\textwidth]{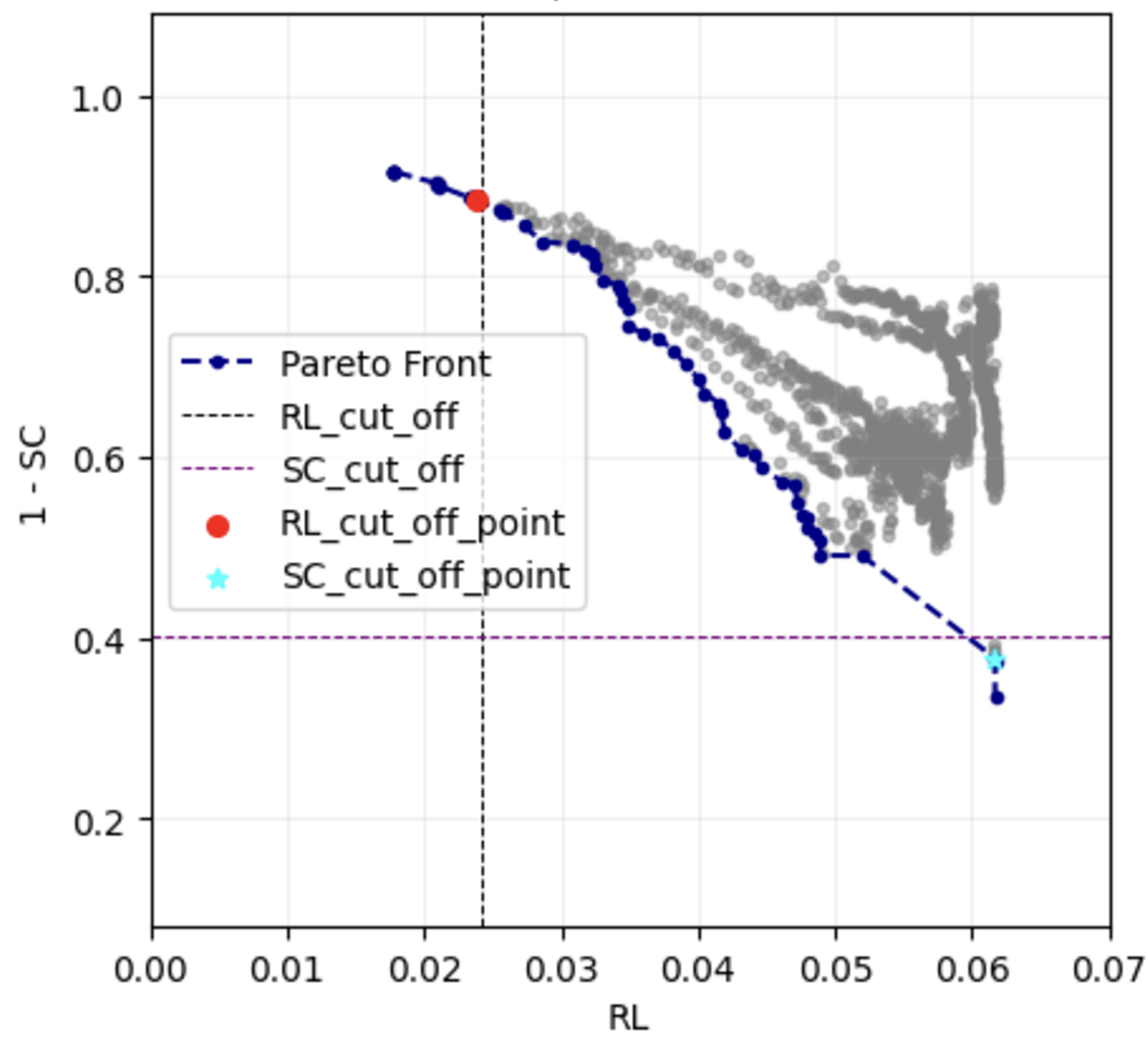}
  \caption{CAE: DEKM}
  \label{fig:hp:c10-cae-dekm}
\end{subfigure}
~
\begin{subfigure}{0.32\columnwidth}
\centering
  \includegraphics[width=0.85\textwidth]{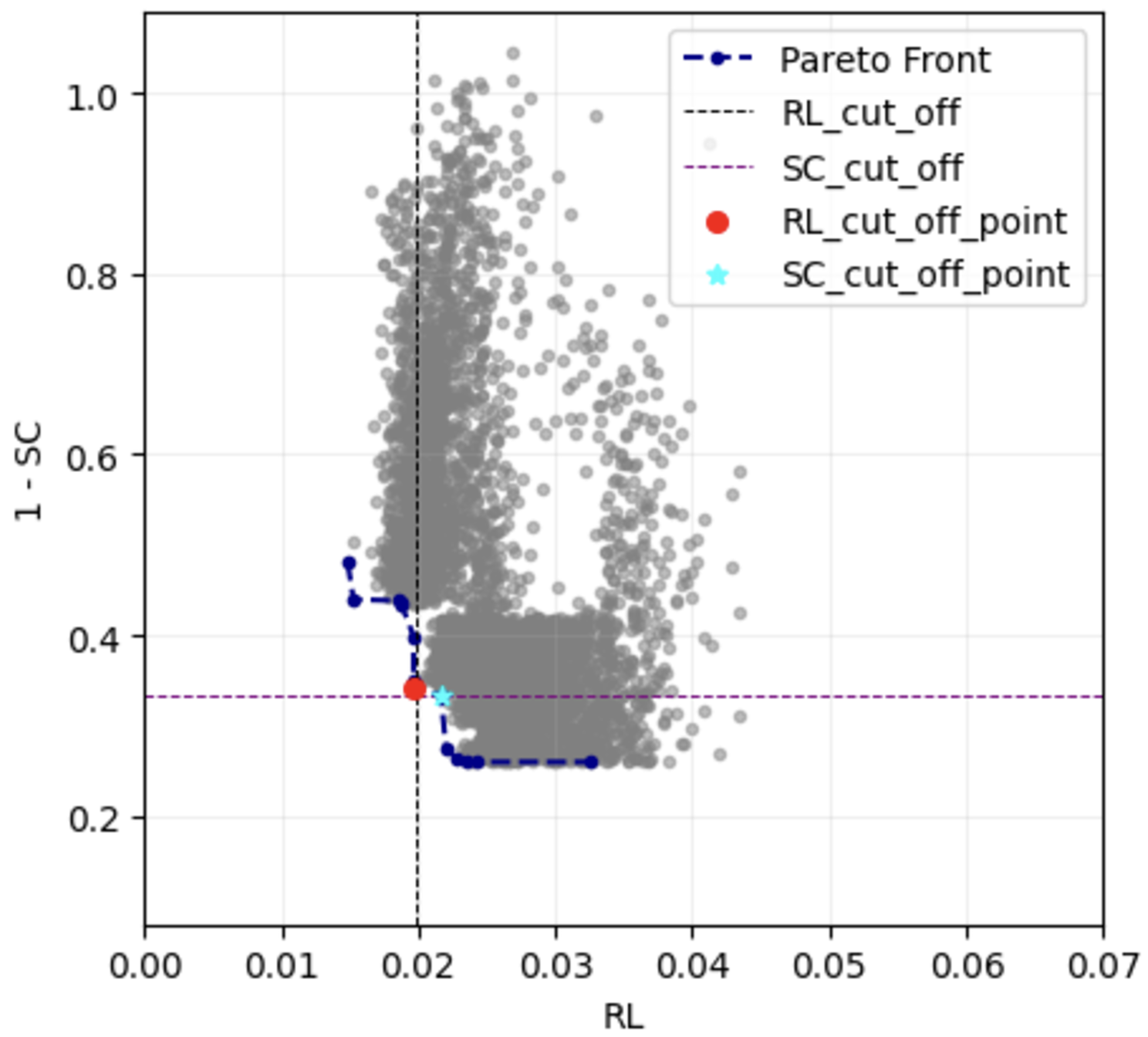}
  \caption{RAE: DCEC}
  \label{fig:hp:c10-rae-dcec}
\end{subfigure}
~
\begin{subfigure}{0.32\columnwidth}
\centering
    \includegraphics[width=0.85\textwidth]{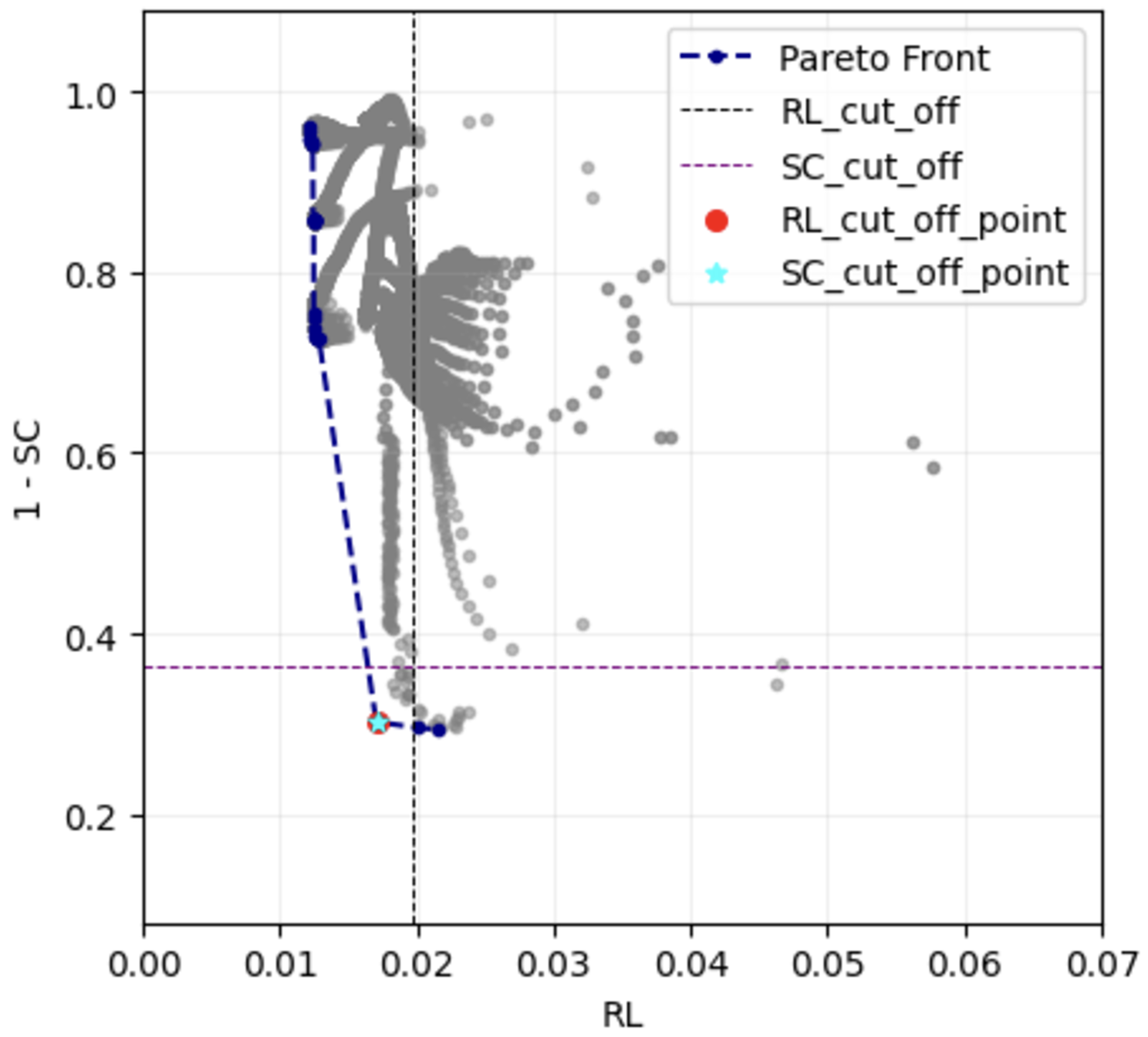}
  \caption{RAE: \dclam}
  \label{fig:hp:c10-rae-dclam}
\end{subfigure}
~
\begin{subfigure}{0.32\columnwidth}
\centering
  \includegraphics[width=0.85\textwidth]{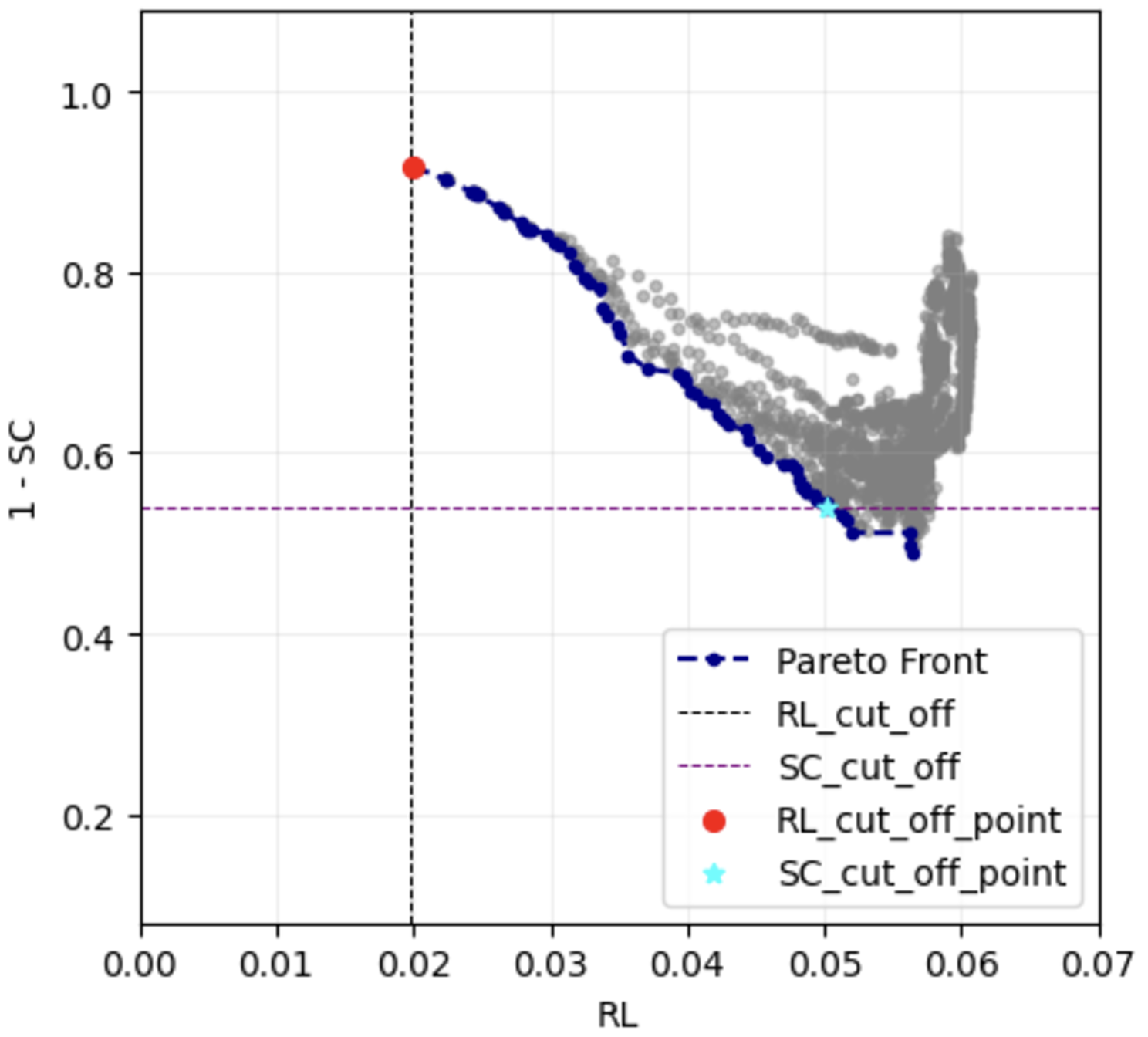}
  \caption{RAE: DEKM}
  \label{fig:hp:c10-rae-dekm}
\end{subfigure}
\end{center}
\caption{\color{black}\textbf{CIFAR10:} Reconstruction loss and clustering quality (1-SC) for all hyperparameter configurations for  DCEC, \dclam and DEKM with CAE and RAE architectures. {\em Lower is better for both axes}.
} 
\label{fig:c10-hp-plots}
\end{figure}

\begin{figure}[!ht]
\begin{center}
\begin{subfigure}{0.32\columnwidth}
\centering
    \includegraphics[width=0.85\textwidth]{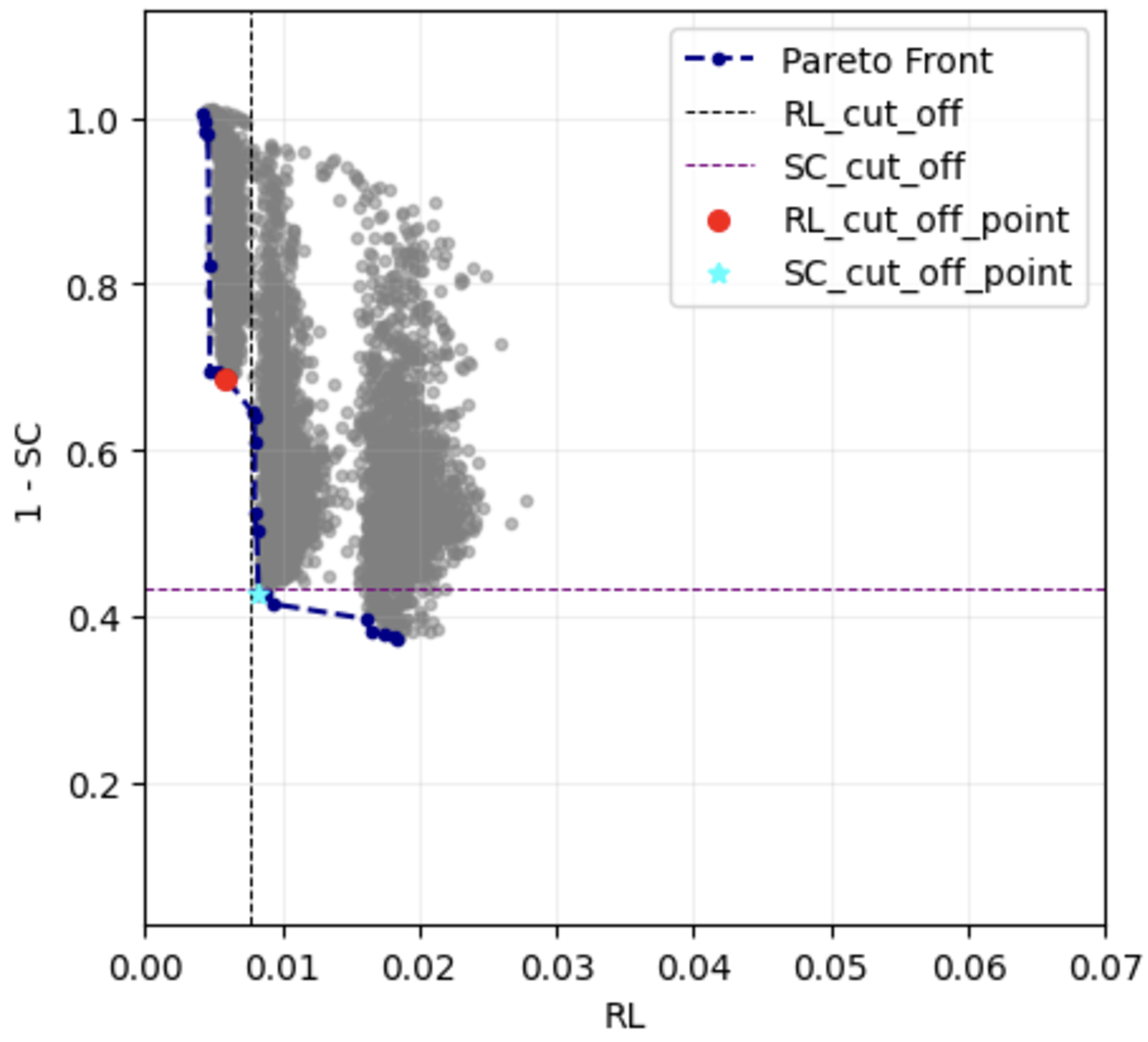}
  \caption{CAE: DCEC}
  \label{fig:hp:c100-cae-dcec}
\end{subfigure}
~
\begin{subfigure}{0.32\columnwidth}
\centering
  \includegraphics[width=0.85\textwidth]{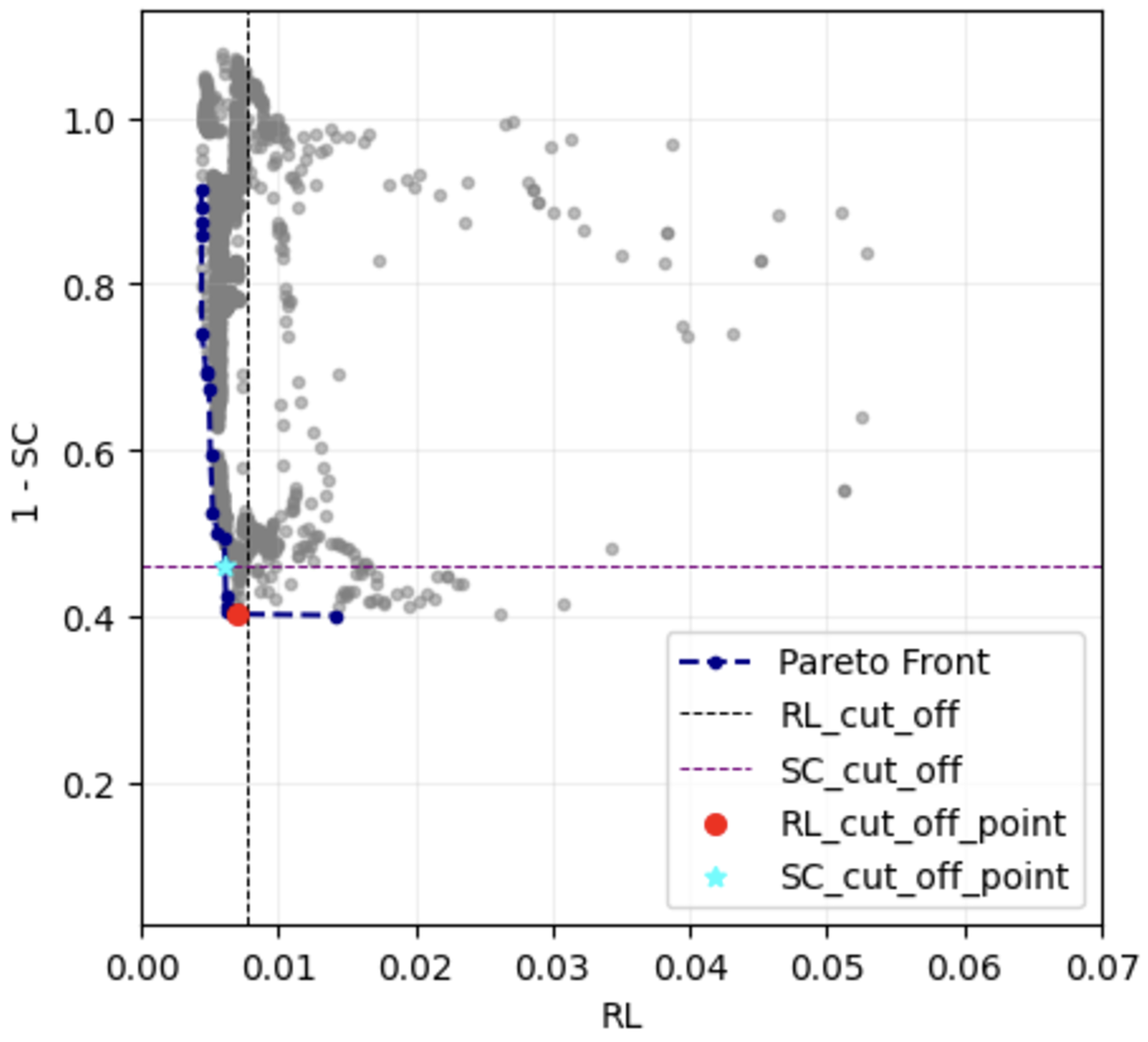}
  \caption{CAE: \dclam}
  \label{fig:hp:c100-cae-dclam}
\end{subfigure}
~
\begin{subfigure}{0.32\columnwidth}
\centering
    \includegraphics[width=0.85\textwidth]{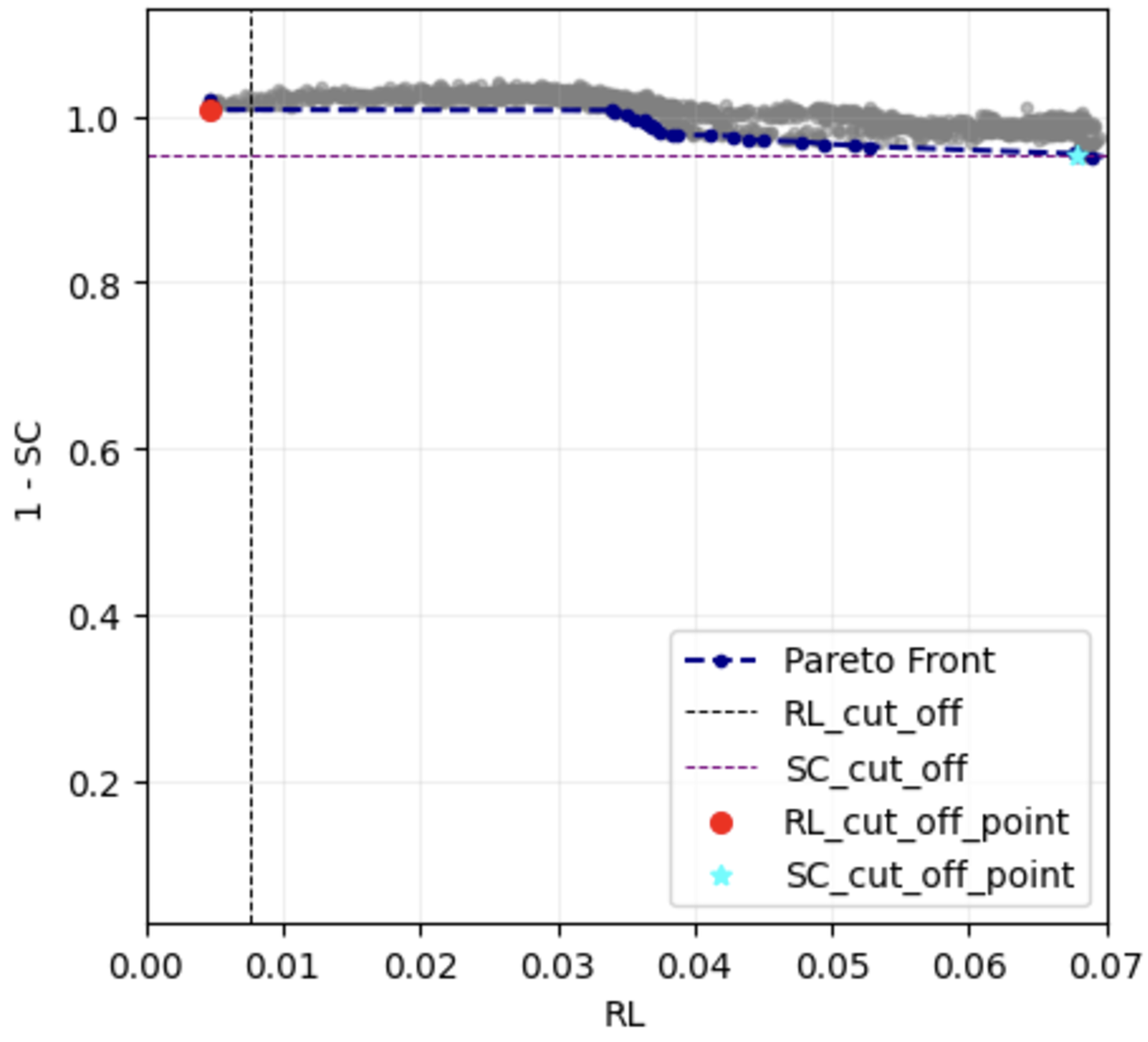}
  \caption{CAE: DEKM}
  \label{fig:hp:c100-cae-dekm}
\end{subfigure}
~
\begin{subfigure}{0.32\columnwidth}
\centering
  \includegraphics[width=0.85\textwidth]{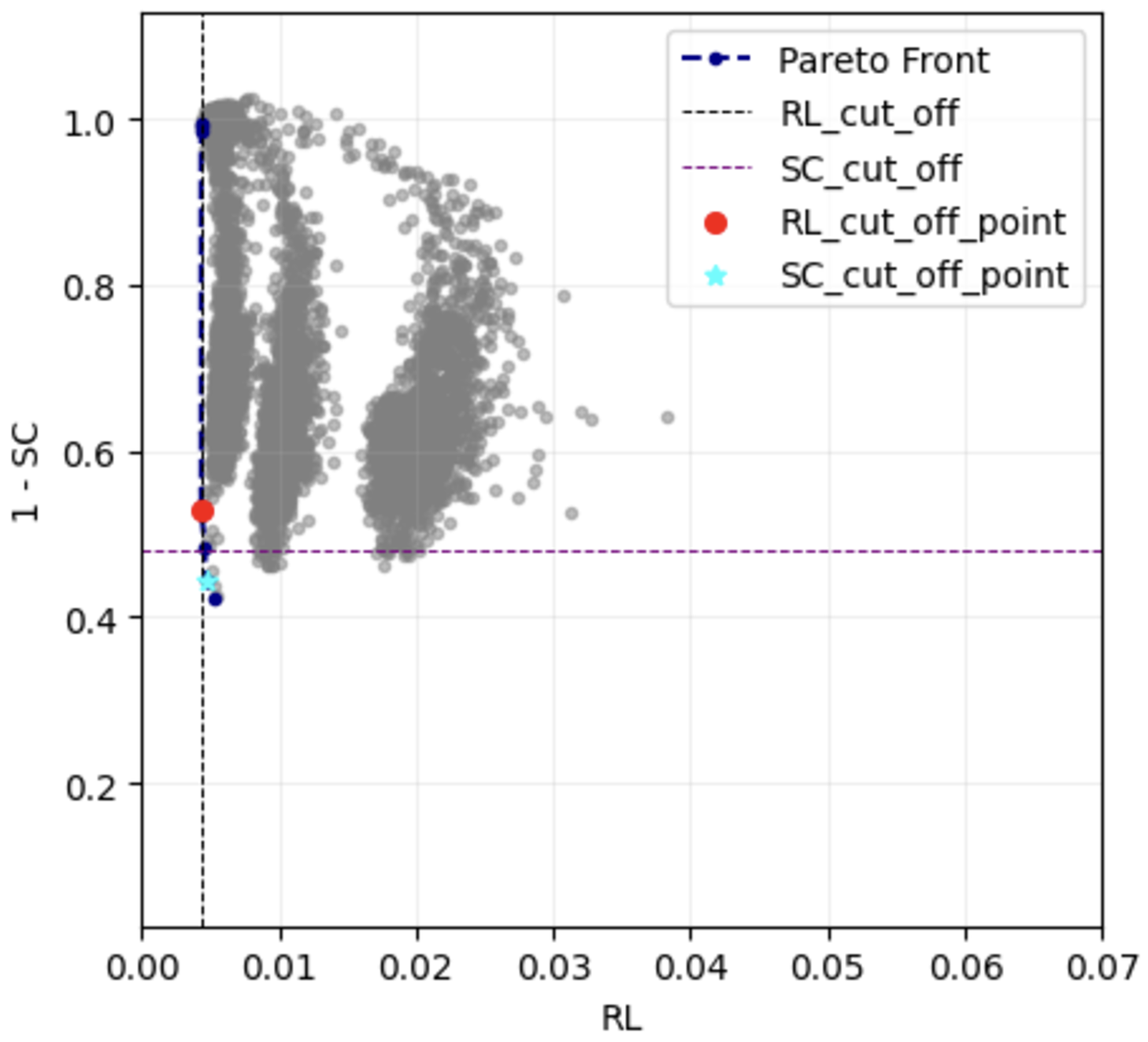}
  \caption{RAE: DCEC}
  \label{fig:hp:c100-rae-dcec}
\end{subfigure}
~
\begin{subfigure}{0.32\columnwidth}
\centering
    \includegraphics[width=0.85\textwidth]{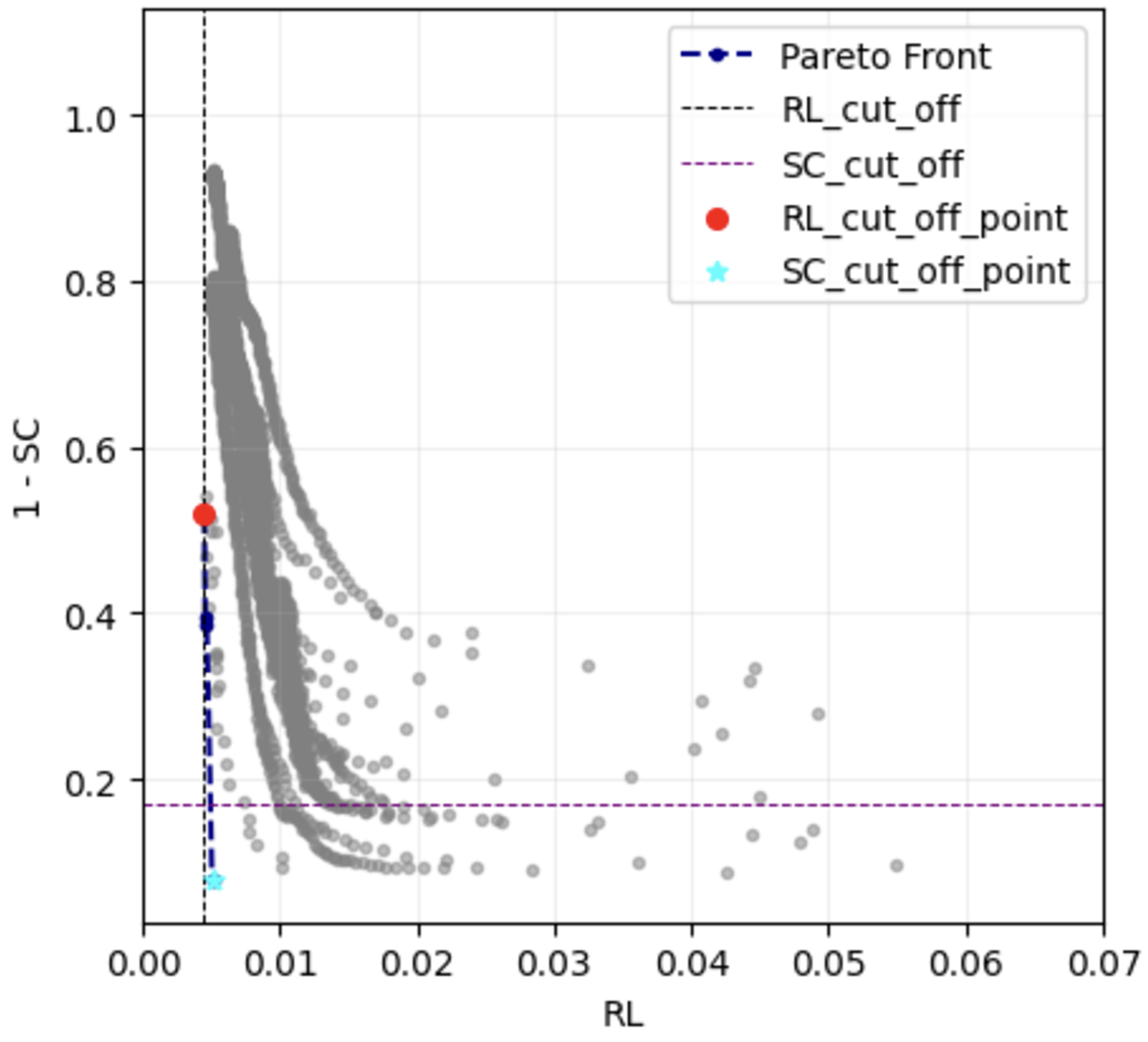}
  \caption{RAE: \dclam}
  \label{fig:hp:c100-rae-dclam}
\end{subfigure}
~
\begin{subfigure}{0.32\columnwidth}
\centering
  \includegraphics[width=0.85\textwidth]{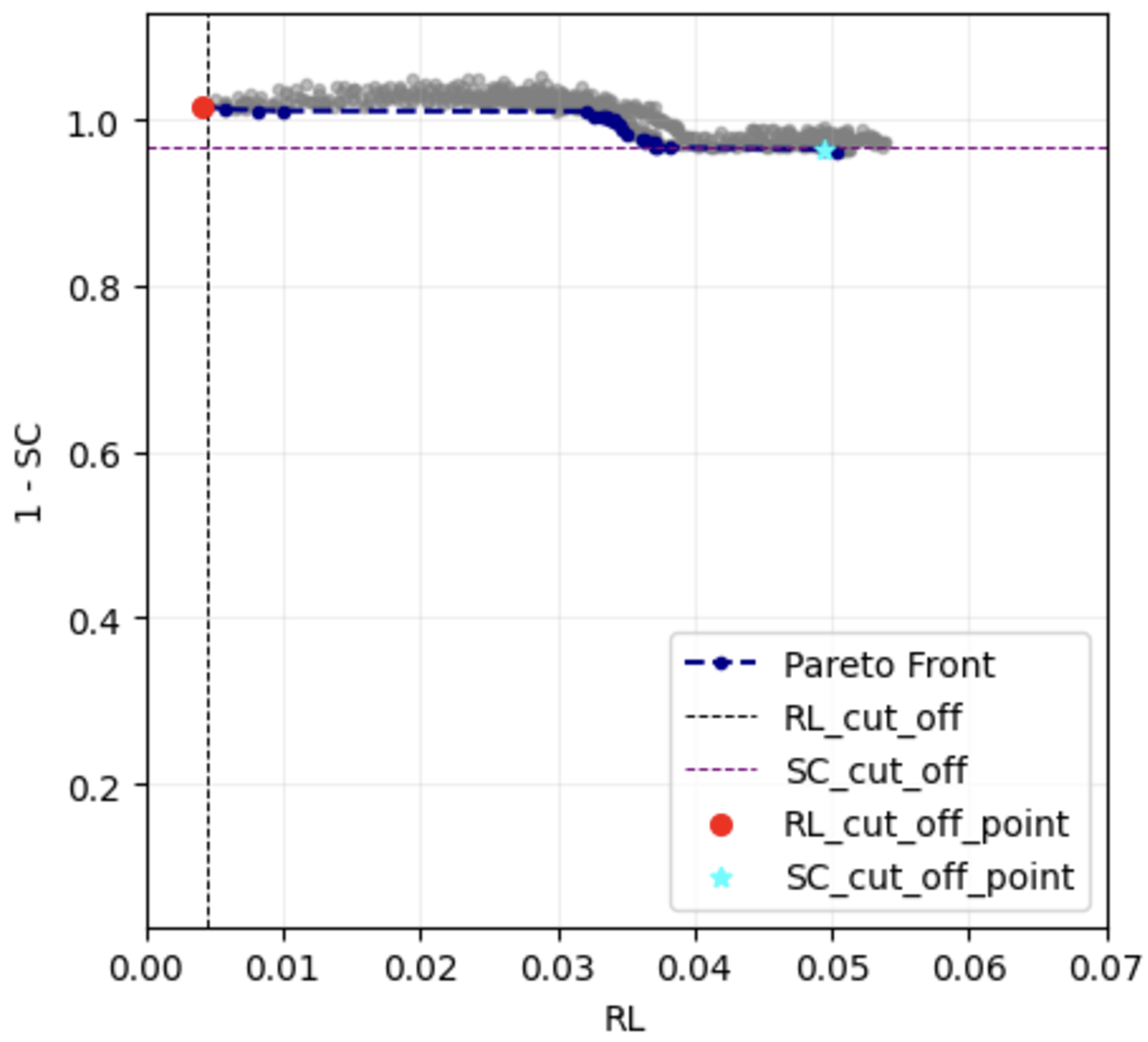}
  \caption{RAE: DEKM}
  \label{fig:hp:c100-rae-dekm}
\end{subfigure}
\end{center}
\caption{\color{black}\textbf{CIFAR100:} Reconstruction loss and clustering quality (1-SC) for all hyperparameter configurations for  DCEC, \dclam and DEKM with CAE and RAE architectures. {\em Lower is better for both axes}.
} 
\label{fig:c100-hp-plots}
\end{figure}

\begin{figure}[!ht]
\begin{center}
\begin{subfigure}{0.32\columnwidth}
\centering
    \includegraphics[width=0.85\textwidth]{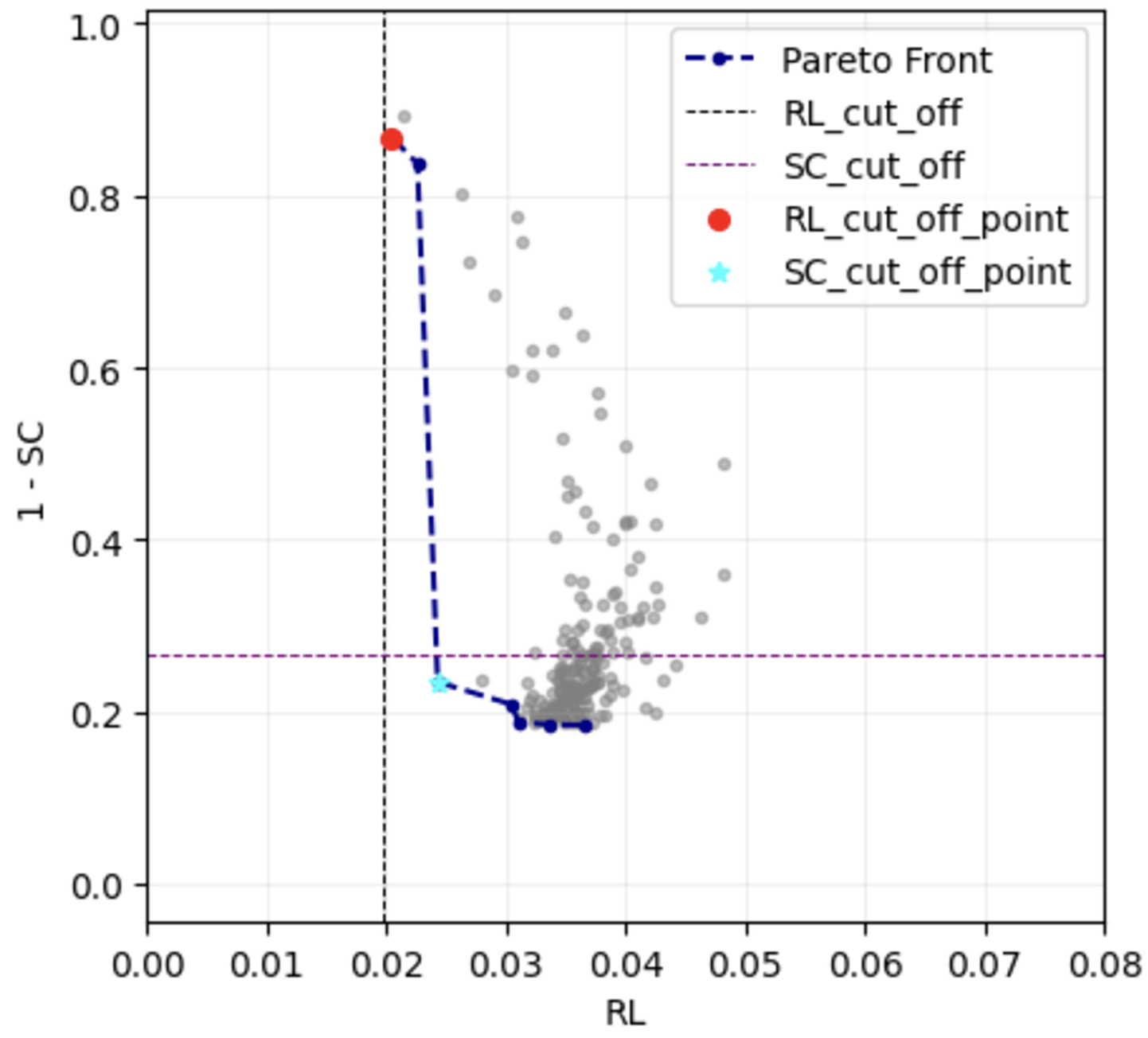}
  \caption{CAE: DCEC}
  \label{fig:hp:stl10-cae-dcec}
\end{subfigure}
~
\begin{subfigure}{0.32\columnwidth}
\centering
  \includegraphics[width=0.85\textwidth]{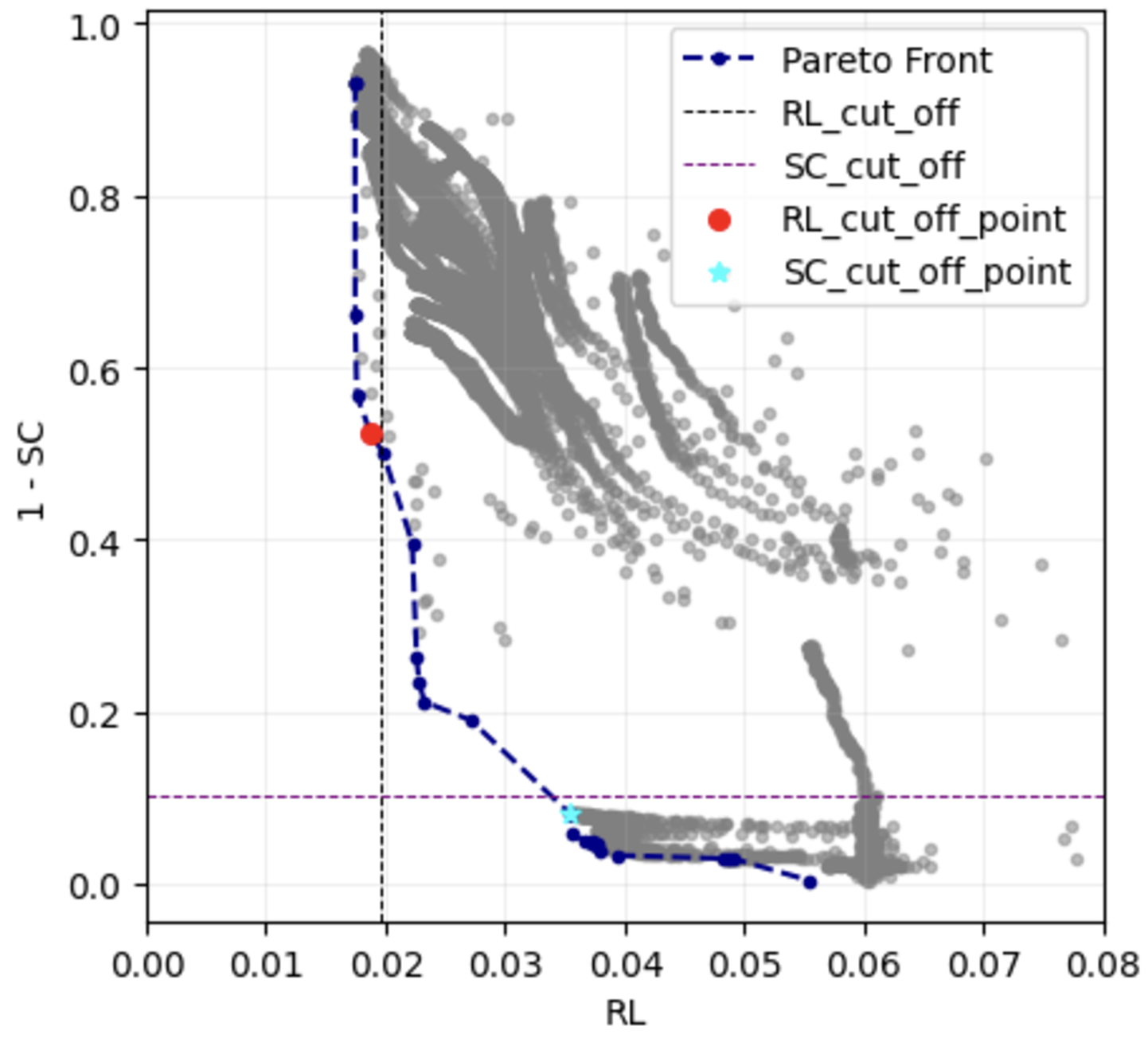}
  \caption{CAE: \dclam}
  \label{fig:hp:stl10-cae-dclam}
\end{subfigure}
~
\begin{subfigure}{0.32\columnwidth}
\centering
    \includegraphics[width=0.85\textwidth]{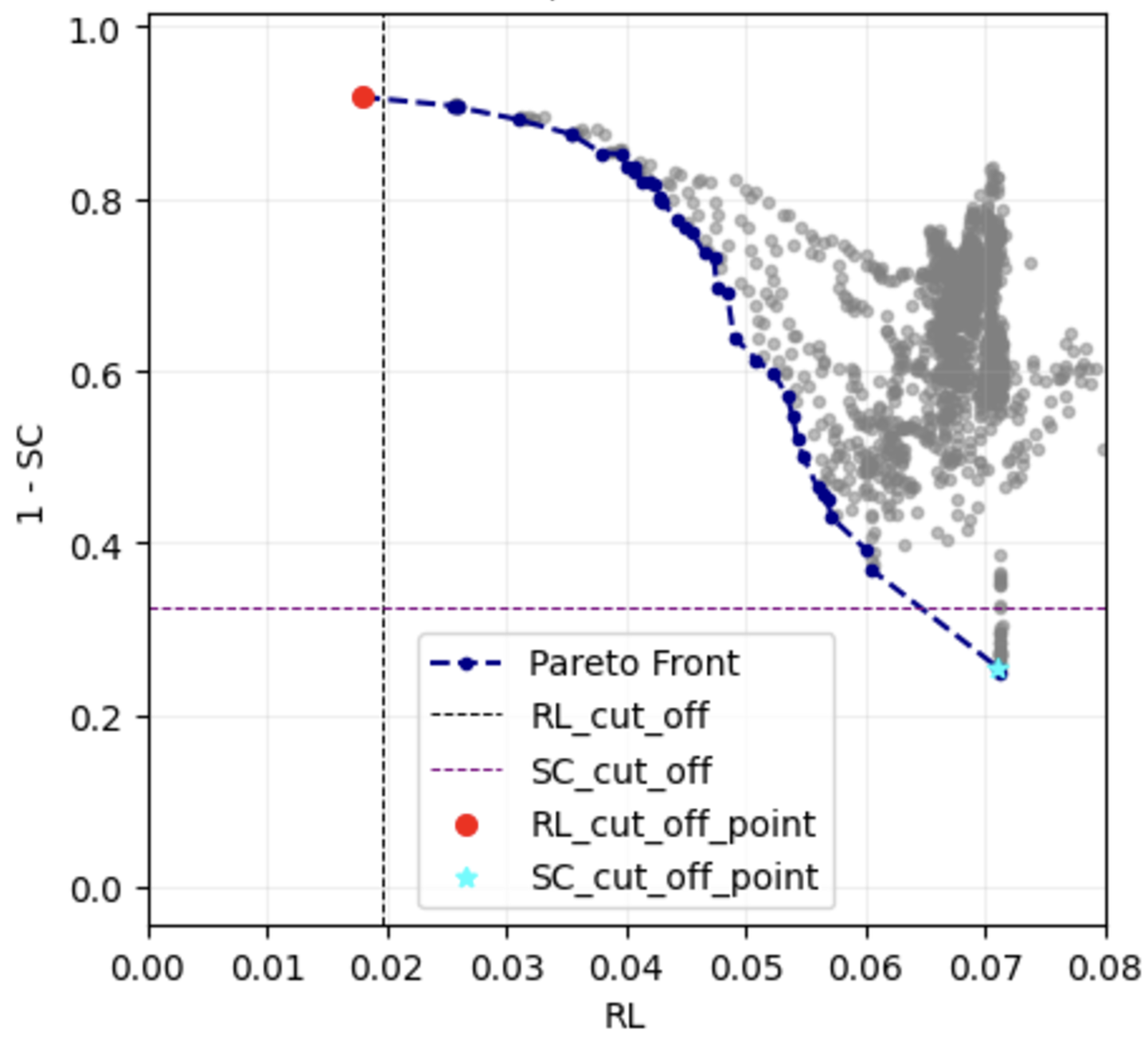}
  \caption{CAE: DEKM}
  \label{fig:hp:stl10-cae-dekm}
\end{subfigure}
~
\begin{subfigure}{0.32\columnwidth}
\centering
  \includegraphics[width=0.85\textwidth]{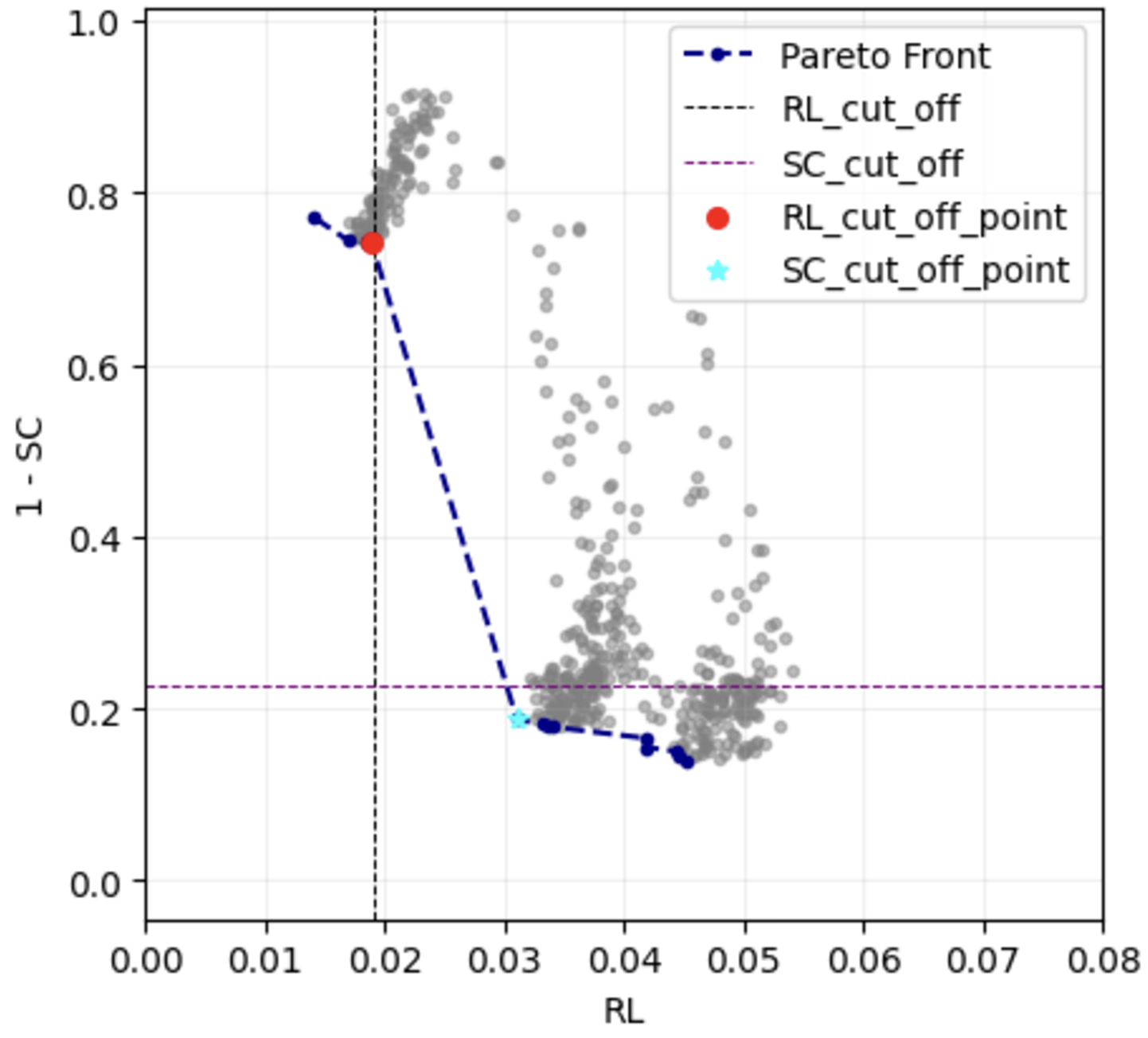}
  \caption{RAE: DCEC}
  \label{fig:hp:stl10-rae-dcec}
\end{subfigure}
~
\begin{subfigure}{0.32\columnwidth}
\centering
    \includegraphics[width=0.85\textwidth]{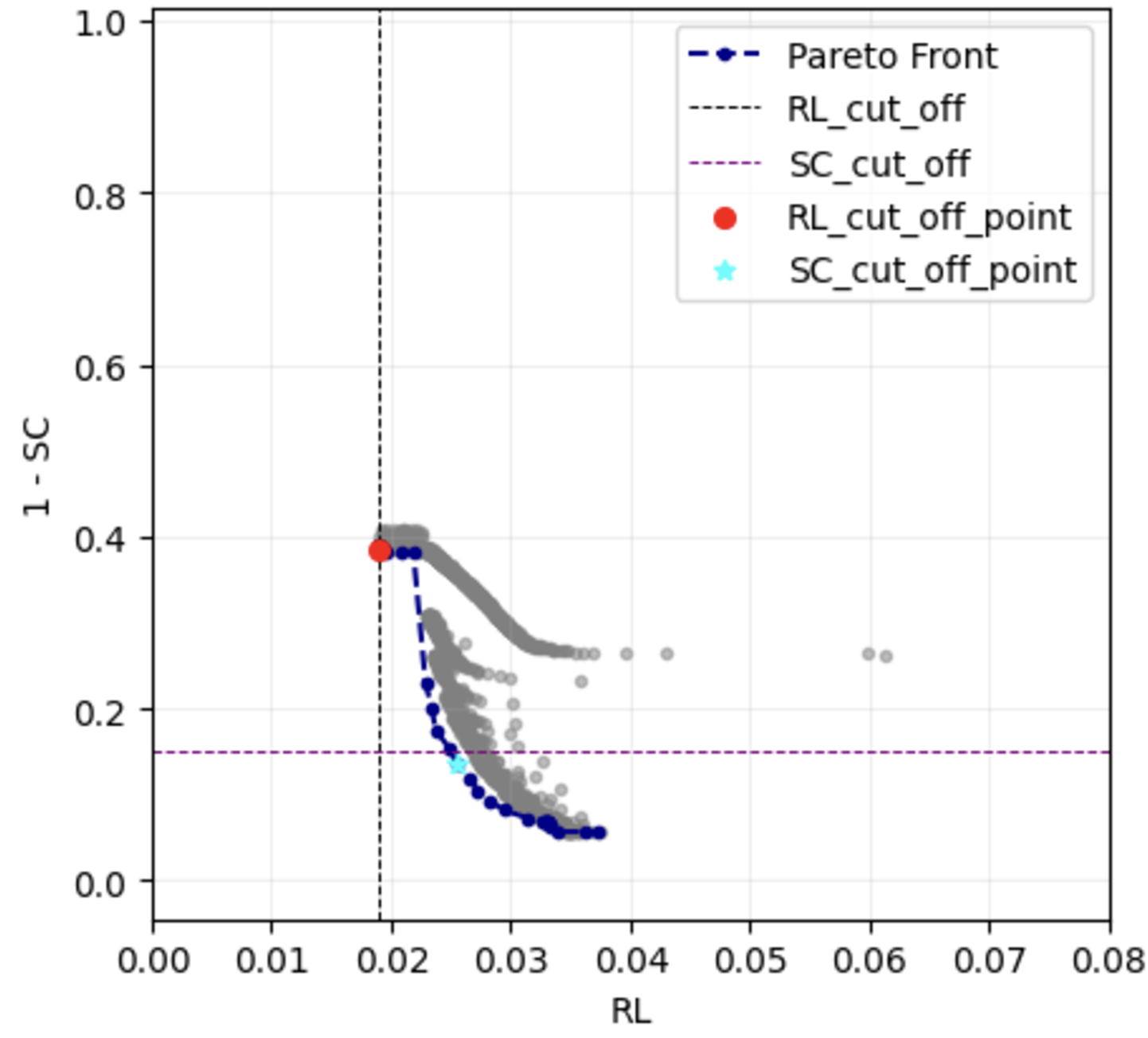}
  \caption{RAE: \dclam}
  \label{fig:hp:stl10-rae-dclam}
\end{subfigure}
~
\begin{subfigure}{0.32\columnwidth}
\centering
  \includegraphics[width=0.85\textwidth]{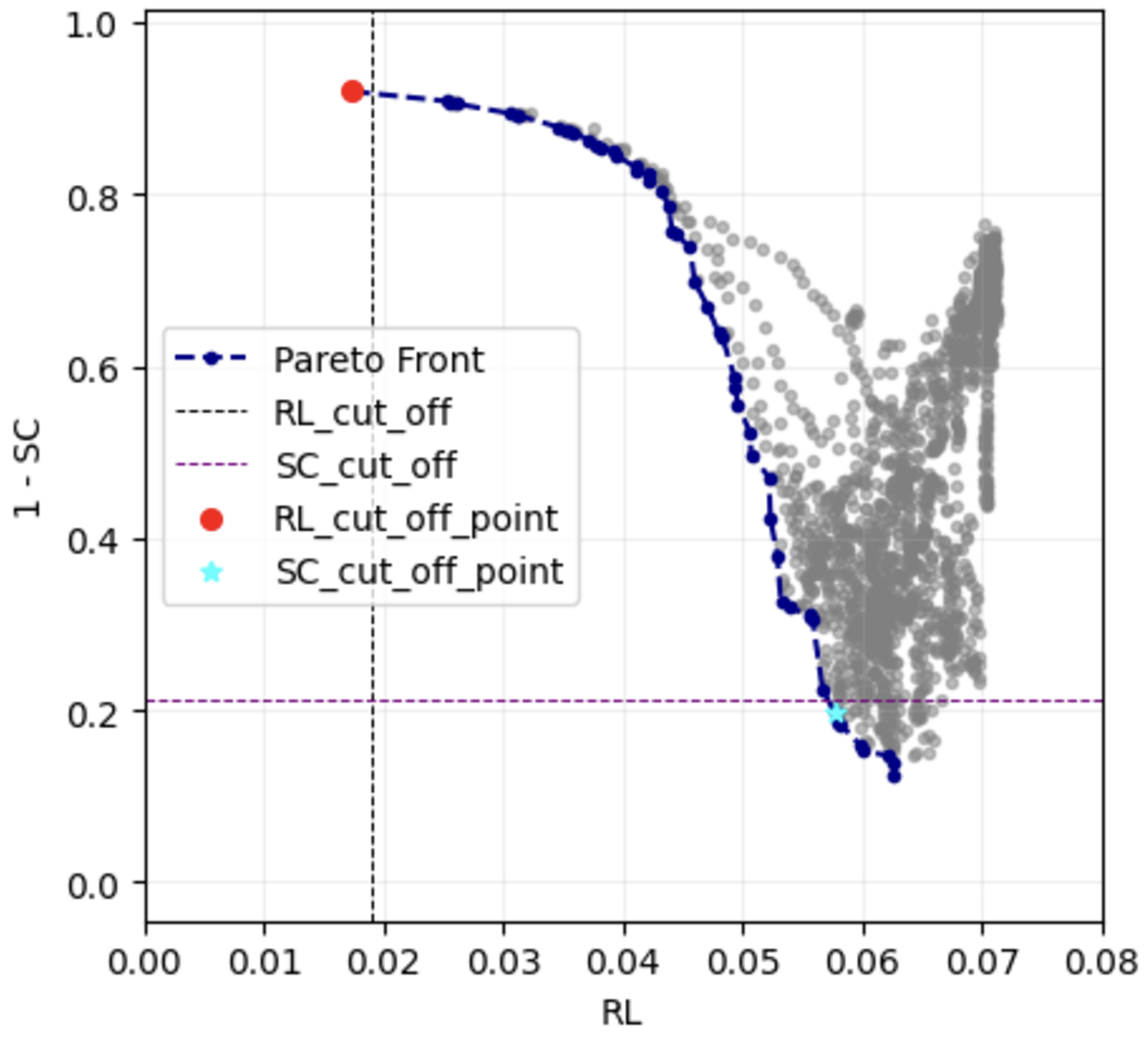}
  \caption{RAE: DEKM}
  \label{fig:hp:stl10-rae-dekm}
\end{subfigure}
\end{center}
\caption{\color{black}\textbf{STL10:} Reconstruction loss and clustering quality (1-SC) for all hyperparameter configurations for  DCEC, \dclam and DEKM with CAE and RAE architectures. {\em Lower is better for both axes}.
} 
\label{fig:stl10-hp-plots}
\end{figure}

In \crefrange{fig:fm-hp-plots}{fig:stl10-hp-plots}, we plot the reconstruction loss (RL) and the silhouette coefficient (SC) for each hyperparameter configuration considered for \dclam and the baselines DCEC and DEKM for the different vision datasets (reported in Tables~\ref{tab:q1-sc-rebuttal}, \ref{tab:q1-rl-rebuttal}, \ref{tab:q1-q2-images-rebuttal}, \ref{tab:scgt-rebuttal}, and \ref{tab:rlgt-rebuttal}). We also highlight the {\em Pareto front} for each of the dataset/method pairs, and the dotted vertical and horizontal lines denote the RL and (1-SC) values corresponding to the 10\% margin from the best RL and (1-SC). Furthermore, the red and cyan highlighted points show the best hyperparameter configuration corresponding to the metric reported in \cref{tab:scgt-rebuttal} and \ref{tab:rlgt-rebuttal}. These results clearly highlight how we thoroughly optimize the hyperparameters, and how we select the final Pareto optimal performance values from the Pareto front to be consistent and fair across all methods. These results clearly show that \dclam offers the best clustering performance in terms of SC, as well as having low reconstruction loss. It also performs very well on the supervised NMI metric. In fact, for NMI, 
 \dclam has the best value in 5 out of the 8 datasets (see \cref{tab:nmigt}).

%
%
%
%
%
%
%

\end{document}